\documentclass[]{bytedance}
\usepackage[toc,page,header]{appendix}

\usepackage{minitoc}
\usepackage{amsfonts}
\usepackage{amssymb}
\usepackage{tabularx}
\usepackage{listings}
\usepackage{xcolor}
\usepackage{cancel}

\usepackage{tabulary,multirow,xspace}
\usepackage{fixmath,mathtools,nicefrac,mmstyle}
\usepackage{subcaption}
\usepackage{caption}
\usepackage{wrapfig} 
\usepackage{colortbl}

\usepackage{wrapfig}
\usepackage{multicol}
\usepackage[most]{tcolorbox}
\usepackage{pifont}

\definecolor{codegreen}{rgb}{0,0.6,0}
\definecolor{codegray}{rgb}{0.5,0.5,0.5}
\definecolor{codepurple}{rgb}{0.58,0,0.82}
\definecolor{backcolour}{rgb}{0.95,0.95,0.92}
\definecolor{boxblue}{RGB}{57,89,163}
\definecolor{boxbluebg}{RGB}{230,237,250} 

\definecolor{lightgraybox}{RGB}{238,238,238}

\newcommand{\pmark}{\ensuremath{\triangle}}
\definecolor{rowblue}{RGB}{230,238,255}

\newtcolorbox{promptfigbox}[1]{
  enhanced,
  colback=lightgraybox,
  colframe=black,
  boxrule=0.6pt,
  arc=2mm,
  left=5mm,
  right=5mm,
  top=3mm, 
  bottom=2mm, 
  title={#1},
  colbacktitle=black,
  coltitle=white,
  fonttitle=\bfseries\small,
  boxed title style={
    colframe=black,
    colback=black,
    boxrule=0pt,
    arc=1mm,
    left=4mm,
    right=4mm,
    top=1.5mm,
    bottom=1.5mm
  },
  attach boxed title to top left={xshift=4mm,yshift=-2mm}
}

\usepackage{booktabs}
\usepackage{multirow}
\usepackage{graphicx}
\usepackage{array}
\usepackage[table]{xcolor}

\definecolor{rowblue}{RGB}{221,233,252}

\definecolor{mygray1}{gray}{.95}
\definecolor{mygray2}{gray}{.9}
\definecolor{mygray3}{gray}{.95}
\usepackage{pifont}

\newlength\savewidth
\newcolumntype{x}[1]{>{\centering\arraybackslash}p{#1pt}}

\newcommand{\app}{\raise.17ex\hbox{$\scriptstyle\sim$}}

\usepackage{xcolor}
\usepackage{graphicx}
\usepackage{amssymb}
\usepackage{pifont}
\usepackage{floatrow}
\usepackage{amsmath} 
\usepackage{float}
\usepackage{wrapfig}
\usepackage{multirow}
\usepackage{tcolorbox}
\tcbuselibrary{breakable, skins, raster}
\usepackage{listings}
\definecolor{ForestGreen}{RGB}{34,139,34}
\definecolor{Emerald}{RGB}{0,128,96}
\definecolor{TealGreen}{RGB}{0,128,128}
\definecolor{DeepGreen}{RGB}{0,100,60}
\definecolor{OliveGreen}{RGB}{85,107,47}

\newcommand{\cmark}{{\color{black}\checkmark}}

\newcommand{\cmarkgreen}{{\color{red}\textbf{\checkmark}}}

\usepackage{listings}

\usepackage{booktabs}
\usepackage{multirow}
\usepackage{makecell}
\usepackage{array}
\usepackage[table]{xcolor}
\usepackage{amssymb}
\usepackage{wasysym}

\usepackage{xcolor}
\definecolor{titleblue}{RGB}{45,88,165}

\usepackage{eso-pic}
\usepackage{caption}

\definecolor{refgreen}{RGB}{30,140,90}
\definecolor{oursrow}{RGB}{232,240,236}

\usepackage{graphicx}
\usepackage{wrapfig}
\usepackage{booktabs}
\usepackage{multirow}
\usepackage{arydshln}
\usepackage{algorithm}
\usepackage{algorithmic}
\definecolor{myblue}{RGB}{210, 225, 255}
\definecolor{mytextblue}{RGB}{51, 161, 201}
\definecolor{mypurple}{RGB}{218, 112, 214}

\definecolor{commentgreen}{rgb}{0.1, 0.4, 0.1}
\definecolor{keywordblue}{rgb}{0.1, 0.1, 0.7}
\definecolor{stringred}{rgb}{0.7, 0.1, 0.1}

\lstdefinestyle{mystyle}{
    commentstyle=\color{commentgreen},
    keywordstyle=\color{keywordblue},   
    stringstyle=\color{stringred},
    basicstyle=\ttfamily\scriptsize, 
    breaklines=true,
    keepspaces=true,
    showstringspaces=false,
    frame=none,                     
    language=Python, 
}

\title{Lance: Unified Multimodal Modeling by \\ Multi-Task Synergy}

\author{
\vspace{-0.6em}
\begin{tabular}{c}
Fengyi Fu$^{1*\ddagger}$ \;
Mengqi Huang$^{1*\dagger\ddagger}$ \;
Shaojin Wu$^{1*}$ \;
Yunsheng Jiang$^{1*}$ \;
Yufei Huo$^{1\ddagger}$ \\
Hao Li$^{1}$ \;
Yinghang Song$^{1}$ \;
Fei Ding$^{1}$ \;
Jianzhu Guo$^{1\dagger\S}$ \;
Qian He$^{1}$ \;
Zheren Fu \\
Zhendong Mao \;
Yongdong Zhang
\end{tabular}
\vspace{-1em}
}

\affiliation{%
$^1$Intelligent Creation Lab, ByteDance
}
\vspace{-3.0em}

\contribution[*]{\centering\small Equal contribution}
\contribution[\dagger]{\centering\small Corresponding Author}
\contribution[\S]{\centering\small Project lead}
\contribution[\ddagger]{\centering\small Work was done during their internship.}
\vspace{-2em}

\abstract{

We present \textbf{Lance}, a lightweight native unified model supporting multimodal understanding, generation, and editing for both images and videos. Rather than relying on model capacity scaling or text-image-dominant designs, Lance explores a practical paradigm for unified multimodal modeling via collaborative multi-task training. It is grounded in two core principles: unified context modeling and decoupled capability pathways. Specifically, Lance is trained from scratch and employs a dual-stream mixture-of-experts architecture on shared interleaved multimodal sequences, enabling joint context learning while decoupling the pathways for understanding and generation. We further introduce modality-aware rotary positional encoding to mitigate interference among heterogeneous visual tokens and boost cross-task alignment. During training, Lance adopts a staged multi-task training paradigm with capability-oriented objectives and adaptive data scheduling to strengthen both semantic comprehension and visual generation performance. Experimental results demonstrate that Lance substantially outperforms existing open-source unified models in image and video generation, while retaining strong multimodal understanding capabilities.

}

\date{\today}
\correspondence{\email{huangmengqi.98@bytedance.com}, \email{ xiaoyu.e@bytedance.com}}
\checkdata[Project Page]{\url{https://lance-project.github.io}}

\begin{document}
\maketitle

\setcounter{figure}{0}
\refstepcounter{figure}
\label{fig:benchmark_comparison}

\edef\benchmarkFigNumber{\thefigure}

\AddToShipoutPictureFG*{%
  \AtPageLowerLeft{%
    \raisebox{1.6cm}{%
      \makebox[\paperwidth][c]{%
        \begin{minipage}{0.92\textwidth}
          \centering
          \color{black}
          \small
          \vspace{-1em}
          \textbf{Figure~\benchmarkFigNumber \quad   Comparison of Lance against representative baselines on multimodal benchmarks.}
        \end{minipage}
      }%
    }%
  }%
}

\vspace*{-2.5em}
{
\centering
\includegraphics[width=0.9\textwidth]{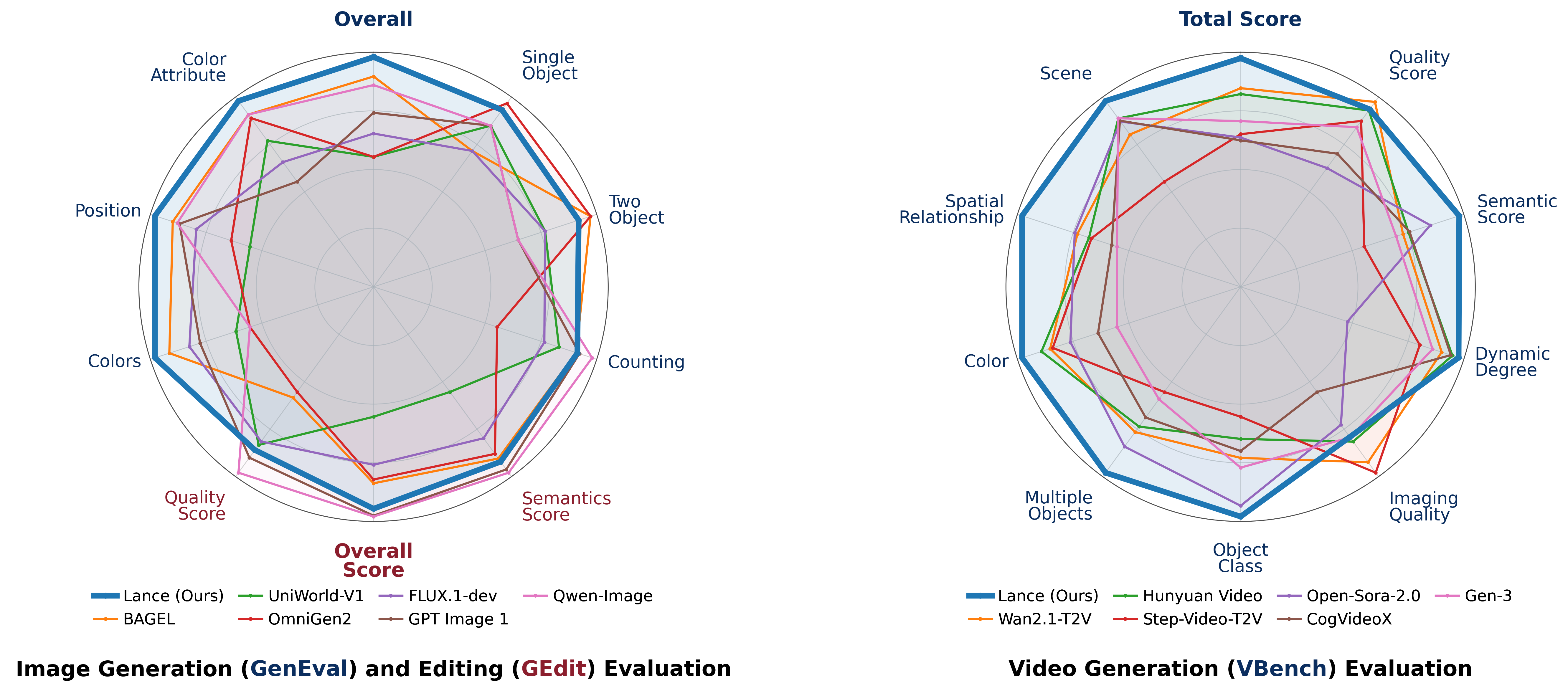}
\par
}
\vspace{-1em}




\section{Introduction}
\label{sec:intro}

Multimodal artificial intelligence is increasingly moving toward a native unified paradigm,  where understanding, reasoning, and generation are integrated within a unified framework. Recently, large language models \cite{alayrac2022flamingo,liu2023visual,li2024llava,Qwen2.5-VL,Qwen3-VL,chen2024internvl} have driven rapid advances in image and video understanding, while diffusion- and flow-based models \cite{esser2024scaling,lipman2024flow,blackforestlabs_flux,labs2025flux,seedream2025seedream,hong2022cogvideo,yang2024cogvideox,seedance2026seedance} have advanced high-fidelity image and video generation. However, most existing systems still evolve along two separate paths: understanding models emphasize semantic reasoning and instruction following, while generative models focus on visual synthesis and spatiotemporal dynamics. Unifying these capabilities in a single unified model remains a central challenge in developing multimodal foundation models with greater generality and stronger practical utility.

\begin{figure}[p]
    \centering
    \vspace*{-0.12\textwidth}
    \includegraphics[width=0.95\textwidth]{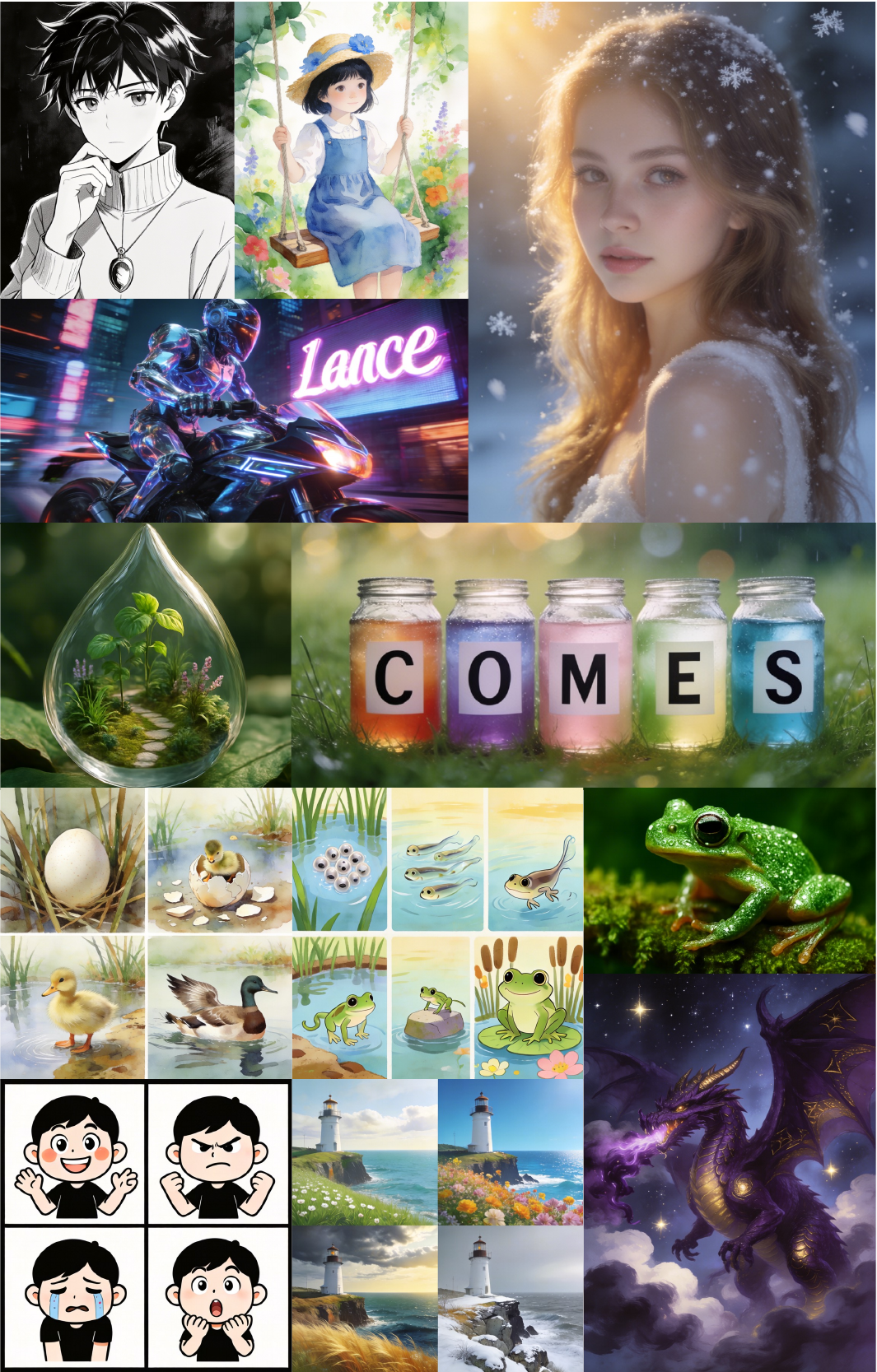}
    \vspace*{-0.01\textwidth}
    \caption{\textbf{Text-to-image generation (T2I) with Lance.}}
    \label{fig:T2I}
\end{figure}

\begin{table*}[t]
\centering
\scriptsize
\setlength{\tabcolsep}{3.3pt}
\renewcommand{\arraystretch}{1.16}

\resizebox{\textwidth}{!}{
\begin{tabular}{
l l
| c c c
| c c c
| c c c
| c c c c
| c
}
\toprule
\multirow{2}{*}{\textbf{Paradigm}}
& \multirow{2}{*}{\textbf{Method}}
& \multicolumn{3}{c|}{\textbf{UND. (Image to Text)}}
& \multicolumn{3}{c|}{\textbf{UND. (Video to Text)}}
& \multicolumn{3}{c|}{\textbf{GEN. (Image)}}
& \multicolumn{4}{c|}{\textbf{GEN. (Video)}}
& \multirow{2}{*}{\makecell[c]{\textbf{Emergent}\\\textbf{Generalization}}} \\
\cmidrule(lr){3-5}
\cmidrule(lr){6-8}
\cmidrule(lr){9-11}
\cmidrule(lr){12-15}
& 
& \textbf{Cap.} & \textbf{Per.} & \textbf{Rea.}
& \textbf{Cap.} & \textbf{Per.} & \textbf{Rea.}
& \textbf{T2I} & \textbf{Edit} & \textbf{S2I}
& \textbf{T2V} & \textbf{I2V} & \textbf{Edit} & \textbf{S2V}
& \\
\midrule

\multirow{6}{*}{\makecell[c]{\textbf{Non-native}\\\textbf{Unified}}}
& MetaQuery-XL \cite{pan2025transfer}
& \cmark & \cmark & \cmark
&  &  &
& \cmark &  &\cmark
&  &  &  &
&  \\

& SEED-X \cite{ge2024seed}
& \cmark & \cmark & \cmark
&  &  &
& \cmark & \cmark &
&  &  &  &
&  \\

& TokenFlow-XL \cite{qu2025tokenflow}
& \cmark & \cmark & \cmark
&  &  &
& \cmark &  &
&  &  &  &
&  \\

& ILLUME \cite{wang2025illume}
& \cmark & \cmark & \cmark
&  &  &
& \cmark & \cmark &
&  &  &  &
&  \\

& InternVL-U \cite{tian2026internvlu}
& \cmark & \cmark & \cmark
&  &  &
& \cmark & \cmark &
&  &  &  &
&  \\

& UniVideo \cite{wei2025univideo}
& \cmark & \cmark & \cmark
& \cmark & \cmark & \cmark
& \cmark & \cmark & \cmark
& \cmark & \cmark & \cmark & \cmark
& \cmarkgreen \\

\midrule

\multirow{16}{*}{\makecell[c]{\textbf{Native}\\\textbf{Unified}}}
& Chameleon \cite{team2024chameleon}
& \cmark & \cmark & \cmark
&  &  &
& \cmark &  &
&  &  &  &
&  \\

& LWM \cite{liu2024world}
& \cmark & \cmark & \cmark
& \cmark & \cmark & \cmark
& \cmark &  &
& \cmark &  &  &
&  \\

& Janus \cite{wu2025janus}
& \cmark & \cmark & \cmark
&  &  &
& \cmark &  &
&  &  &  &
&  \\

& Janus-Pro \cite{chen2025janus}
& \cmark & \cmark & \cmark
&  &  &
& \cmark &  &
&  &  &  &
&  \\

& Transfusion \cite{zhou2024transfusion}
& \cmark & \cmark & \cmark
&  &  &
& \cmark &  &
&  &  &  &
&  \\

& Emu3 \cite{wang2024emu3}
& \cmark & \cmark & \cmark
& \pmark & \pmark & \pmark
& \cmark &  &
& \cmark &  &  &
&  \\

& Show-o \cite{xie2024show}
& \cmark & \cmark & \cmark
&  &  &
& \cmark & \cmark &
&  &  &  &
&  \\

& Show-o2 \cite{xie2025show}
& \cmark & \cmark & \cmark
& \cmark & \cmark & \cmark
& \cmark & \cmark &
& \pmark &  &  &
&  \\

& Bagel \cite{deng2025emerging}
& \cmark & \cmark & \cmark
&  &  & 
& \cmark & \cmark & \cmark
& & &  &
& \cmarkgreen \\

& Mogao \cite{liao2025mogao}
& \cmark & \cmark & \cmark
&  &  &
& \cmark & \pmark & \pmark
&  & &  &
& \\

& HaploOmni \cite{xiao2025haploomni}
& \cmark & \cmark & \cmark
& \cmark & \cmark & \cmark
& \cmark &  &
& \cmark &  &  &
&  \\

& VILA-U \cite{wu2024vila}
& \cmark & \cmark & \cmark
& \cmark & \cmark & \cmark
& \cmark &  &
& \cmark &  &  &
&  \\

& HunyuanImage 3.0 \cite{cao2025hunyuanimage}
& \pmark & \pmark & \pmark
&  &  &
& \cmark & \cmark &
&  &  &  &
&  \\

& Emu3.5 \cite{cui2025emu3}
& \cmark & \cmark & \cmark
& \pmark & \pmark & \pmark
& \cmark & \cmark & \pmark
& \pmark & \pmark &  &
& \cmarkgreen \\

& TUNA \cite{liu2025tuna}
& \cmark & \cmark & \cmark
& \cmark & \cmark & \cmark
& \cmark & \cmark &
& \cmark &  &  &
&  \\

& TUNA-2 \cite{tuna2}
& \cmark & \cmark & \cmark
&  &  & 
& \cmark & \cmark &
&  &  &  &
&  \\

\rowcolor{rowblue}
& \textbf{Lance (Ours)}
& \cmark & \cmark & \cmark
& \cmark & \cmark & \cmark
& \cmark & \cmark & \cmark
& \cmark & \cmark & \cmark & \cmark
& \cmarkgreen \\

\bottomrule
\end{tabular}
}
\caption{
\textbf{Comparison of multimodal unified models by supported task categories.}  $\cmark$ indicates explicit support; $\triangle$ indicates description-only support without official code; blank cells indicate no explicit report.
{Cap.}, {Per.}, {Rea.} indicate understanding ability on captioning, perception, and reasoning. 
 The last column denotes whether the model exhibits emergent generalization on unseen tasks.
Models are categorized as native or non-native unified models based on whether they are jointly pre-trained as a unified architecture or assembled from separately pre-trained components.
}
\label{tab:task_support_matrix}
\end{table*}

\begin{figure*}[!t]
    \centering
    \includegraphics[width=1\textwidth]{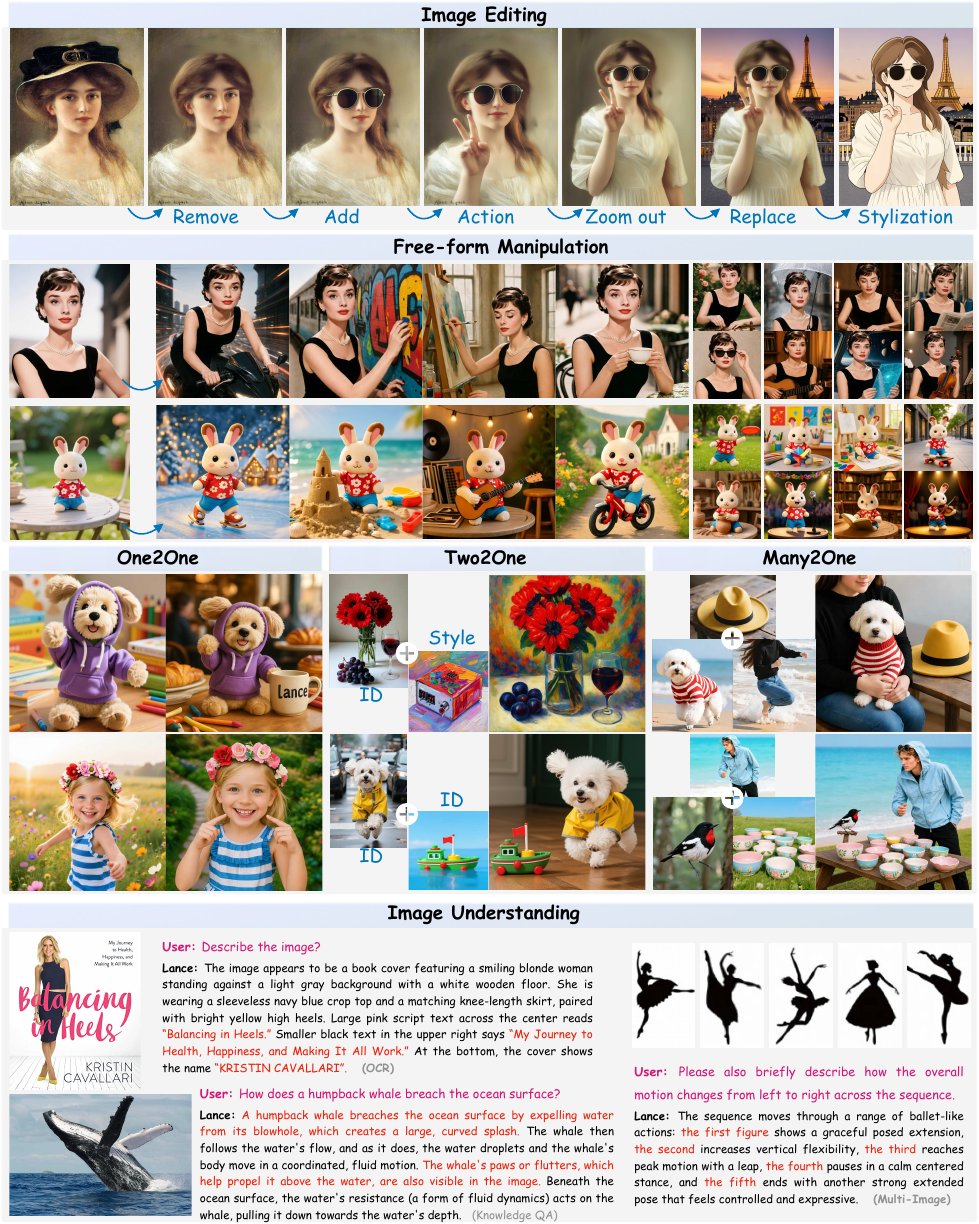}
    \caption{\textbf{Any-to-image generation (X2I) and image understanding (I2T) with Lance.}}
    \label{fig:X2I}
\end{figure*}

\begin{figure*}[!t]
    \centering
    \includegraphics[width=1\textwidth]{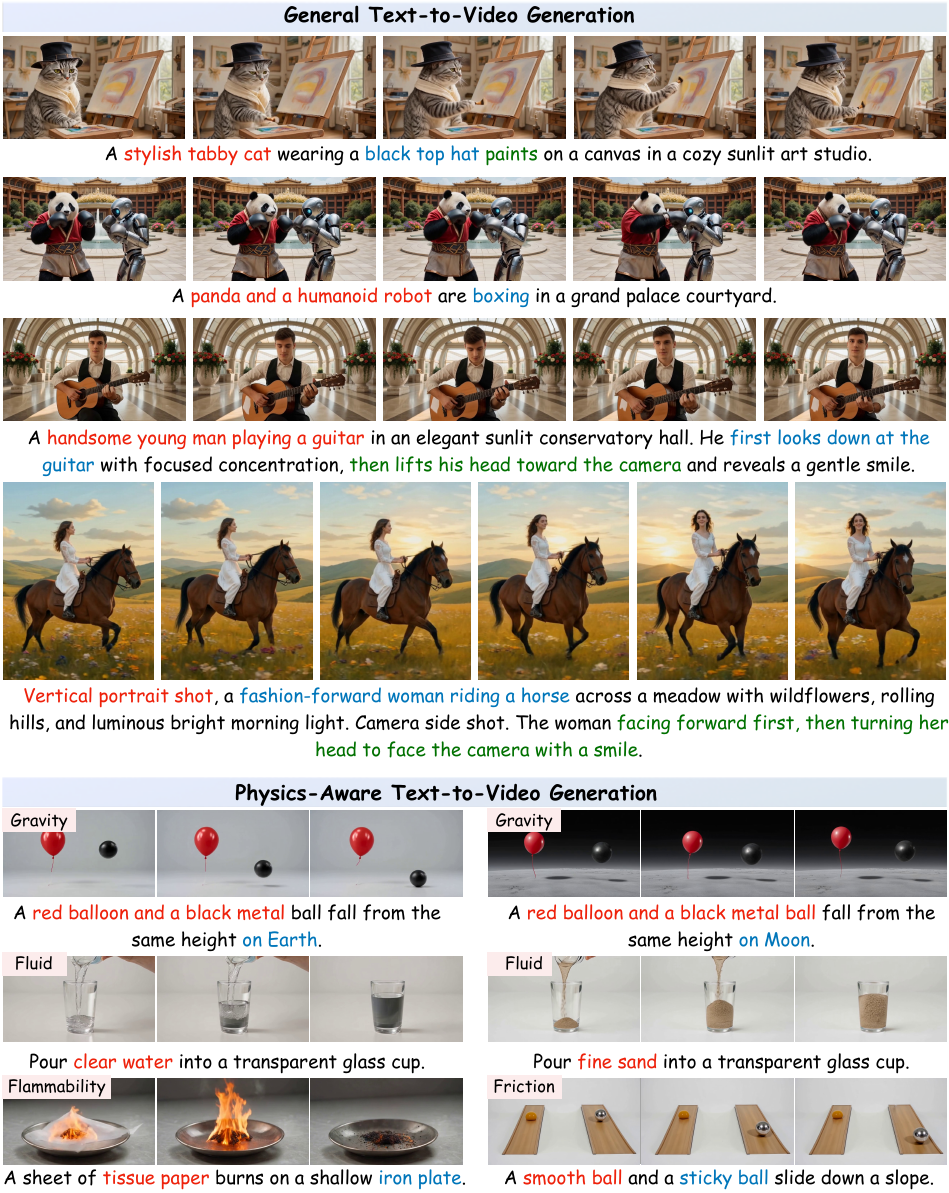}
    \caption{\textbf{Text-to-video generation (T2V) with Lance.}}
    \label{fig:T2V}
\end{figure*}

\begin{figure*}[!t]
    \centering
    \includegraphics[width=1\textwidth]{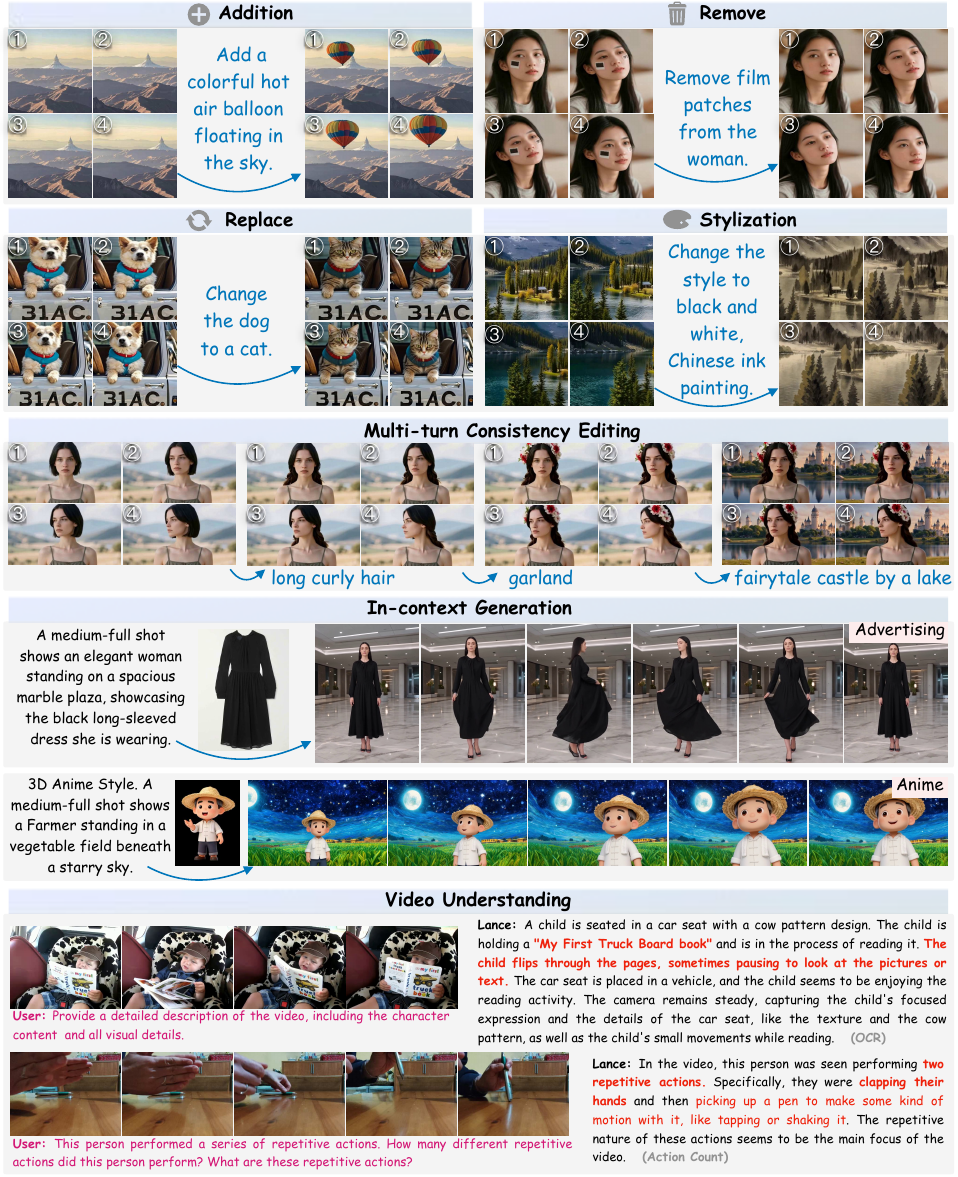}
    \caption{\textbf{Any-to-video generation (X2V) and video understanding (V2T) with Lance.}}
    \label{fig:X2V}
\end{figure*}

Recent unified multimodal models \cite{team2024chameleon,cui2025emu3,deng2025emerging,xie2025show,liao2025mogao,liu2025tuna} have made encouraging progress, yet two fundamental limitations remain. First, the visual-representation requirements of understanding and generation are inherently misaligned: the former benefits from high-level semantic features aligned with language, whereas the latter requires low-level continuous representations that preserve texture, geometry, and temporal dynamics.
Existing approaches therefore typically follow one of two directions. One line of work \cite{xie2024show,team2024chameleon, wang2024emu3, cui2025emu3, liu2025tuna} attempts to support both tasks with a unified visual representation, yielding a simpler modeling formulation but often struggling to balance semantic reasoning and generation quality. Another line~\cite{deng2025emerging,liao2025mogao,xie2025show} adopts decoupled semantic and generative representations, alleviating representational mismatch at the cost of increased architectural and optimization complexity.

Second, and more importantly, existing unified models remain limited in task coverage and training formulation. As summarized in \Cref{tab:task_support_matrix}, most prior methods \cite{team2024chameleon,liu2024world,ge2024seed,qu2025tokenflow,wu2025janus} are still largely confined to text-image domains or partial task combinations, leaving the full image-video understanding and generation space insufficiently explored.
Although recent unified models \cite{deng2025emerging,xie2025show,liu2025tuna} have progressively extended to the video domain, they typically cover only limited subsets of the full image-video task space, while diverse generation-oriented tasks such as editing and subject-driven generation are often introduced as downstream fine-tuning skills rather than being systematically optimized within a unified multi-task training process.
Meanwhile, the comparison in \Cref{tab:task_support_matrix} further suggests that models with broader task coverage are more likely to exhibit emergent generalization on unseen tasks. 
This motivates us to view multi-task learning not simply as capability aggregation, but as a way to promote transfer across modalities and task formulations.

Based on this observation, we present \textbf{Lance}, a lightweight native unified multimodal model that systematically integrates joint learning across X2T, X2I, and X2V tasks, covering image and video understanding, generation, and editing within a single framework. 
By unifying these task families in a single native model, Lance aims to better harness cross-task synergy and further advance the potential of unified multimodal modeling.
Lance is designed to balance \textit{unified context modeling} with \textit{decoupled capability pathways} from both the architectural and training perspectives. 
Architecturally, it adopts a shared interleaved multimodal sequence representation to enable unified context learning, while employing a dual-stream mixture-of-experts framework to allocate dedicated capacity to semantic reasoning and visual synthesis.
 To better coordinate heterogeneous visual tokens within the unified context sequence, we further introduce modality-aware rotary positional encoding, MaPE, which mitigates positional interference and improves cross-task contextual alignment.
In terms of training, Lance follows a staged multi-task training paradigm that casts diverse understanding, generation, and editing tasks into a unified task formulation, and combines capability-oriented objectives with adaptive data scheduling to progressively strengthen semantic understanding and visual synthesis.

Extensive experiments show that Lance achieves strong performance across multimodal understanding and generation benchmarks, with qualitative examples shown in \Cref{fig:T2I,fig:X2I,fig:T2V,fig:X2V}. With only $3$B activated parameters, Lance substantially outperforms existing open-source unified models on image and video generation tasks as shown in \Cref{fig:benchmark_comparison}, while maintaining advanced multimodal understanding ability. Notably, all these gains are achieved within a $128$-GPU training budget, highlighting the feasibility of resource-efficient unified multimodal modeling.

Our main contributions are summarized as follows:

(1) \textbf{Concepts:} We present Lance, a lightweight native unified multimodal model that explicitly supports the full spectrum of image/video understanding and generation tasks within a single model, extending unified modeling beyond text-image domains and partial task coverage. Lance emphasizes multi-task synergy not as simple capability aggregation, but as a mechanism for promoting transfer across modality-task boundaries.

(2) \textbf{Technique:}
We develop a dual-stream mixture-of-experts architecture that preserves a shared interleaved multimodal sequence representation while allocating dedicated visual representations and model capacity to understanding and generation. We further introduce a modality-aware positional encoding scheme and a staged multi-task training paradigm to improve heterogeneous visual token coordination and cross-task context modeling.

(3) \textbf{Performance:} Extensive experiments demonstrate that Lance achieves competitive performance across multimodal understanding and generation benchmarks with only $3$B activated parameters. 

\section{Related Work}
\label{sec:Related Work}

\subsection{Multimodal Large Language Models}
Multimodal large language models (MLLMs) have become the dominant paradigm for image and video understanding by aligning pretrained visual encoders with powerful language backbones.
Representative early systems include Flamingo \cite{alayrac2022flamingo}, IDEFICS \cite{laurenccon2023obelics}, and InstructBLIP \cite{dai2023instructblip}, while later open-source families such as LLaVA \cite{liu2023visual,liu2024improved,liu2024llavanext,li2024llava}, Qwen-VL \cite{Qwen-VL,Qwen2-VL,Qwen2.5-VL,Qwen3-VL}, and InternVL \cite{chen2024internvl,gao2024mini,chen2024far,wang2025internvl3_5} further improve instruction following, high-resolution perception, and long-context multimodal reasoning. 
This line of work mainly follows the LLaVA paradigm \cite{liu2023visual}, in which visual inputs are first encoded by a vision encoder \cite{radford2021learning,tschannen2025siglip} and then concatenated with text tokens for joint modeling by a language model decoder.
Some proprietary models such as GPT \cite{achiam2023gpt} and Gemini \cite{team2024gemini,team2023gemini} also demonstrate strong multimodal reasoning ability. 
Recent progress further extends these models to interleaved image-text modeling \cite{yang2024vision,cui2025emu3,deng2025emerging} and video understanding \cite{li2025videochat,lin2024video,yang2025cambrian}.
Despite their strong semantic abstraction and cross-modal alignment capabilities, these models are primarily optimized for understanding and text generation, rather than native visual synthesis.

\subsection{Visual Generative Models}
Visual generation has been dominated by diffusion- and flow-based frameworks \cite{ho2020denoising,esser2024scaling,lipman2024flow,wu2024vmix,huang2024realcustom, mao2024realcustom++,fu2025feededit,fu2026layeredit,mou2025dreamo,blackforestlabs_flux,labs2025flux}, which serve as mainstream paradigms for high-fidelity image and video synthesis.
As for image generation, representative large-scale systems include Stable Diffusion \cite{rombach2022high,podell2024sdxl,wu2024taiyidiffusionxl,esser2024scaling}, FLUX \cite{blackforestlabs_flux,labs2025flux}, Qwen-Image \cite{wu2025qwen}, and HunyuanImage 3.0 \cite{cao2025hunyuanimage}, while multimodal image generation models such as RealCustom++ \cite{huang2024realcustom, mao2025realcustom++} and UNO series \cite{wu2025less,cheng2025umo,wu2025uso} further advance these frameworks by supporting diverse multimodal conditional inputs. As for video generation, recent systems such as Wan \cite{wan2025wan}, HunyuanVideo \cite{wu2025hunyuanvideo} and CogVideo \cite{hong2022cogvideo,yang2024cogvideox} demonstrate the effectiveness of continuous latent modeling with dedicated temporal VAEs.
In contrast to continuous latent generators, autoregressive visual token models \cite{ramesh2021zero,chang2022maskgit,esser2021taming,peebles2023scalable,kondratyuk2023videopoet,tian2024visual,huang2023towards,mao2026toward} formulate image generation as next-token prediction, providing a simpler unified token interface, but often face trade-offs in visual fidelity and generation efficiency.
Recently, several studies \cite{liu2024mardini,li2024autoregressive,fan2025unified} have explored hybrid frameworks that combine diffusion modeling with autoregressive modeling, aiming to leverage the advantages of both in generation quality and modeling flexibility, thereby further advancing visual generation capabilities.

\subsection{Unified Multimodal Models}

Recent unified multimodal models (UMMs) attempt to bridge multimodal understanding and visual generation within a single framework. One line follows a fully {autoregressive formulation}, represented by Chameleon \cite{team2024chameleon}, Emu3/Emu3.5 \cite{wang2024emu3,cui2025emu3}, and more recent systems such as TokenFlow \cite{qu2025tokenflow}, HunyuanImage 3.0 \cite{cao2025hunyuanimage}. These models cast both understanding and generation into next-token prediction under a shared token space. These models offer a clean unified interface and naturally support mixed-modality sequence modeling, but they may still face nontrivial trade-offs among reasoning ability, visual fidelity, and generation efficiency.
Another line adopts {autoregressive–diffusion hybrid formulations}, combining language modeling for text with diffusion- or flow-based modeling for visual generation.
Representative works include Transfusion \cite{zhou2024transfusion}, Show-o/Show-o2 \cite{xie2024show,xie2025show}, BLIP3-o \cite{chen2025blip3}, BAGEL \cite{deng2025emerging}, and others \cite{zhao2025unified,liu2025tuna,wang2025ovis,he2025emma,li2025onecat,tian2025unigen,ma2025janusflow,dai2026chatumm,feng2026dreamlite}.
Within this family, recent work further explores decoupling in representation design, module architecture, and optimization. For instance, Janus-series models \cite{zhao2025unified,ma2025janusflow} decouple visual encoding for understanding and generation; RealGeneral \cite{lin2025realgeneral} tames a pretrained video foundation model for unified image generation and editing; Show-o2 \cite{xie2025show} integrates autoregressive language modeling with flow matching, extending native unification to both image and video modalities; BAGEL \cite{deng2025emerging} studies expert specialization under a shared decoder-only backbone; TUNA \cite{liu2025tuna} emphasizes unified continuous visual representations; and InternVL-U \cite{tian2026internvlu} couples a strong open MLLM with a specialized generation head. 
In addition to native unified models, modular bridging systems such as OmniBridge \cite{xiao2025omnibridge} connect pretrained understanding and generation models through latent-space alignment, offering a more lightweight but less fully native alternative.

Although unified multimodal modeling has advanced rapidly, much of the literature remains image-centric. Extending unified modeling to the video domain is substantially more challenging because it requires not only semantic understanding but also temporal reasoning, motion modeling, long-context generation, and consistent editing.
Early general any-to-any or modular systems such as NEXT-GPT \cite{wu2024next} and GPT4Video \cite{wang2024gpt4video} extend MLLMs with external generative backends to support multimodal understanding and video generation, but their video synthesis capability is still largely mediated through additional generators rather than native joint video modeling. More recent video-focused frameworks, including Omni-Video \cite{tan2025omni}, UniVideo \cite{wei2025univideo}, and TV2TV \cite{han2025tv2tv}, move closer to genuinely unified video models by jointly addressing video understanding, generation, editing, or interleaved language-video modeling under a more integrated architecture. Meanwhile, several task-unified video editing frameworks, such as AnyV2V \cite{ku2024anyv2v}, VACE \cite{jiang2025vace}, UNIC \cite{ye2025unic}, EditVerse \cite{ju2025editverse}, and FullDiT \cite{ju2025fulldit}, expand the controllability of video generation, but typically do not aim for full understanding-generation unification within a single multimodal model. 
Overall, multi-task synergy for image-video unified multimodal modeling remains to be further explored.

\section{Methodology}
\label{sec:Methodology}

\begin{figure*}[t]
    \centering
    \includegraphics[width=0.99\textwidth]{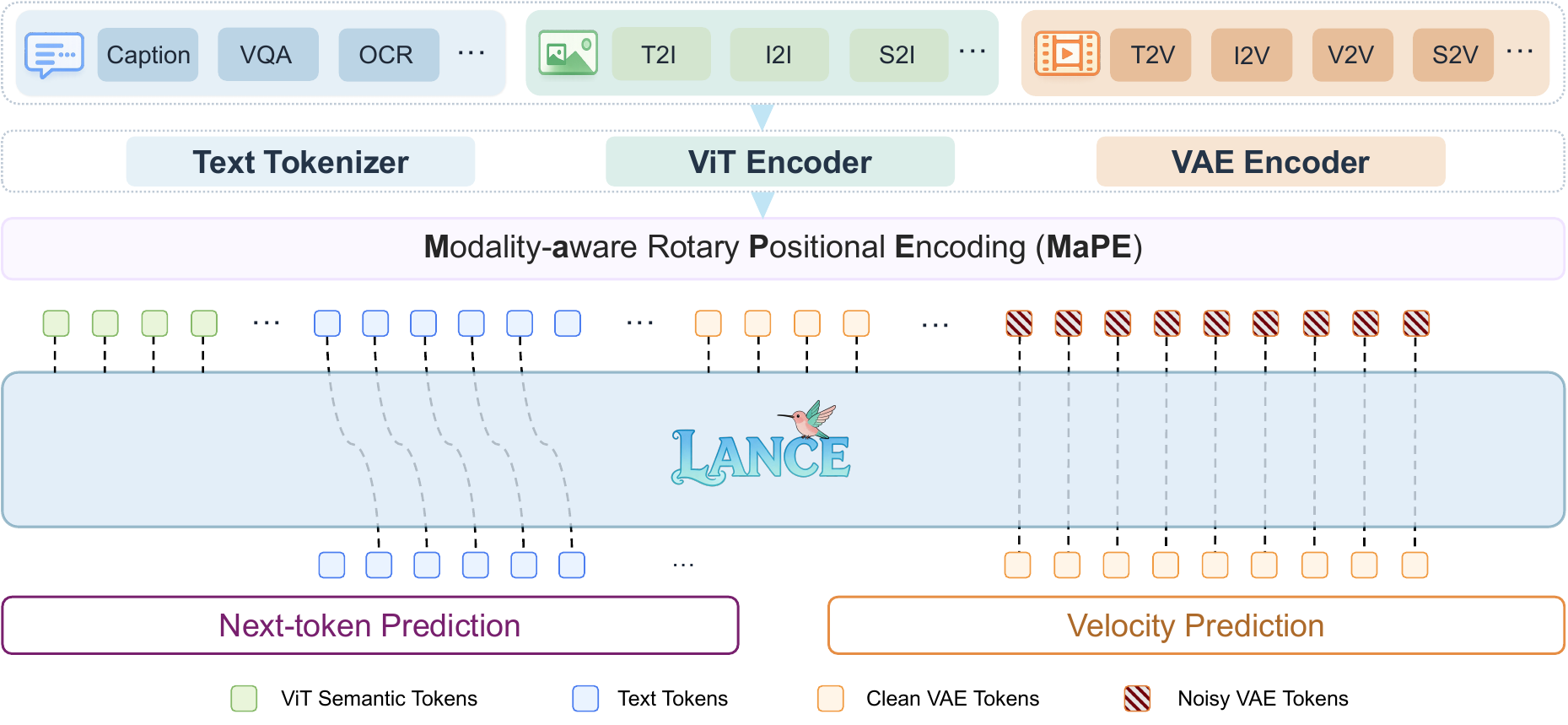}
    \caption{
    \textbf{Overview of Lance.}
     Given multi-task inputs spanning X2T, X2I, and X2V, Lance encodes all input tokens into a unified MaPE-enhanced multimodal context sequence. The dual-expert backbone performs generalized 3D causal attention over the shared context and produces task-specific hidden states, which are further decoded by an LM head for autoregressive next-token prediction and by a flow head for velocity prediction in the visual latent space.}
    \label{fig:model_framework}
\end{figure*}

The core idea of Lance is that broad multi-task learning can further unlock the potential of unified multimodal models. However, different task families, such as multimodal understanding, generation, and editing, impose substantially different requirements on modeling objectives, visual representations, and optimization dynamics. An effective unified model should therefore enable different tasks to interact within \textit{unified context learning}, while mitigating interference among heterogeneous objectives through \textit{decoupled capability pathways}.

\subsection{Design Motivation and Principles}


Lance is built upon two principles: \textit{unified context learning} and \textit{decoupled capability pathways}. 
Unified context learning is enabled by interleaved multimodal sequence modeling and multi-task collaborative optimization, while decoupled capability pathways are motivated by the following observations.

\textbf{Autoregressive vs. Diffusion.}
Autoregressive next-token prediction remains the dominant paradigm for language modeling \cite{touvron2023llama,achiam2023gpt,liu2024deepseek} and multimodal understanding \cite{Qwen3-VL,xu2024pllava,li2025videochat}. 
In contrast, high-quality image and video synthesis is more effectively modeled in continuous latent spaces with diffusion or flow-matching objectives \cite{ding2021cogview,li2023blip,cai2024diffusion_selfdistill,labs2025flux,wu2025qwen}. 
 Some unified models \cite{team2024chameleon,wu2024vila,wang2024emu3,qu2025tokenflow} also explore fully autoregressive formulations for joint understanding and generation, which may suffer from sequential decoding and limited generation efficiency. 
We therefore adopt autoregressive language modeling for understanding and flow matching for generation.

\textbf{Unified Visual Representations vs. Decoupled Visual Representations.}
Understanding and generation rely on different forms of visual information. Understanding mainly benefits from high-level semantic visual features that are well aligned with language (\eg, SigLIP 2 \cite{tschannen2025siglip} or Qwen2.5-VL \cite{Qwen2.5-VL}), whereas generation relies on low-level latent representations that preserve appearance and spatiotemporal structure~\cite{wan2025wan}. 
Some existing works \cite{liu2025tuna} have explored shared visual representations, but a single representation may be insufficient to simultaneously satisfy semantic reasoning and high-fidelity synthesis.
Meanwhile, recent studies~\cite{yu2024representation,zheng2025diffusion} suggest that semantic features can also benefit generation modeling. Lance therefore keeps semantic visual tokens and generative latent tokens decoupled, while organizing them within a shared interleaved multimodal sequence for unified context learning.

\textbf{Shared Backbone vs. Specialized Expert Capacity.}
A fully shared backbone that uses single stream to process various modalities \cite{huang2022dse,xie2025show,liu2025tuna} offers a clean unified architecture, but it forces understanding and generation to compete for the same parameters under substantially different objectives. Recent evidence from Bagel \cite{deng2025emerging} and HunyuanImage 3.0 \cite{cao2025hunyuanimage} further suggests that decoupling generation-oriented parameters and understanding-oriented parameters yields clear advantages over dense shared backbones.
These observations motivate Lance to preserve a unified multimodal token interface for bottleneck-free context fusion, while allocating specialized expert capacity to understanding and generation pathways.

\subsection{Overall Architecture}

\textbf{Overall Framework.}
An overview of our framework is shown in \Cref{fig:model_framework}. 
Given interleaved inputs of text, images, and videos, Lance first converts each modality into task-appropriate token representations. These heterogeneous tokens are then organized into a shared interleaved multimodal sequence with modality-aware rotary positional encoding, supporting unified context modeling across diverse task formats.
To reconcile unified context learning with task-specific capability specialization, Lance adopts a dual-expert architecture initialized from Qwen2.5-VL~\cite{Qwen2.5-VL}. The understanding expert, denoted as $\mathrm{LLM}_{\mathrm{UND}}$, processes text and semantic visual tokens for multimodal reasoning and text generation, while the generation expert, denoted as $\mathrm{LLM}_{\mathrm{GEN}}$, processes VAE latent tokens for visual synthesis and editing. The two experts operate over the same interleaved multimodal context, preserving cross-task interaction while avoiding direct competition between heterogeneous objectives. 
Task-specific heads are further used for autoregressive language modeling and flow-based visual generation, respectively.

\textbf{Unified Context Learning.}
Lance first converts heterogeneous inputs into a shared interleaved multimodal sequence. (1) Text instructions are embedded using the language embedding layer of Qwen2.5-VL~\cite{Qwen2.5-VL}. (2) For understanding-oriented visual inputs, Lance employs the Qwen2.5-VL ViT encoder~\cite{Qwen2.5-VL}, which uses $14\times$ spatial and $2\times$ temporal patching followed by a $2\times2$ spatial merge to produce compact semantic visual tokens. These tokens provide language-aligned visual semantics for multimodal understanding and reasoning. (3) For generation-oriented visual inputs, we encode images or videos into continuous latent representations using the Wan2.2 3D causal VAE encoder \cite{wan2025wan}. This encoder jointly supports image and video modalities through a unified latent space with $16\times$ spatial downsampling and $4\times$ temporal downsampling for videos.
 The resulting latent features preserve the low-level appearance and temporal structure required for high-fidelity visual generation, and are projected into the hidden space of the generation backbone through a lightweight MLP connector.

As a result, Lance represents each sample as a unified interleaved multimodal sequence of text tokens, ViT semantic tokens, clean VAE latent tokens, and noisy VAE latent tokens:
\begin{equation}
\mathcal{S}
=
\cdots \oplus
\mathcal{B}_{\mathrm{text}}(\mathbf{T})
\oplus
\mathcal{B}_{\mathrm{vis}}(\mathbf{V}_{\mathrm{vit}})
\oplus
\mathcal{B}_{\mathrm{vis}}(\mathbf{V}_{\mathrm{vae}}^{\mathrm{clean}})
\oplus
\mathcal{B}_{\mathrm{vis}}(\mathbf{V}_{\mathrm{vae}}^{\mathrm{noisy}})
\oplus
\mathcal{B}_{\mathrm{text}}(\mathbf{T}')
\oplus \cdots ,
\end{equation}
\begin{equation}
\mathcal{B}_{\mathrm{text}}(\mathbf{T})
=
[\texttt{BOT}, \mathbf{T}, \texttt{EOT}],
\quad
\mathcal{B}_{\mathrm{vis}}(\mathbf{V})
=
[\texttt{BOV}, \mathbf{V}, \texttt{EOV}].
\end{equation}
This formulation supports understanding, generation, and mixed interleaved multimodal samples within a single context modeling framework.

To handle such heterogeneous sequences, Lance adopts \textit{generalized 3D causal attention}. The sequence is partitioned into modality-specific segments, where each segment attends to preceding clean segments to preserve causal dependencies. Within each segment, text tokens use causal attention, while visual tokens use bidirectional attention to capture spatial and spatiotemporal structure. This provides a unified attention mechanism for multimodal understanding, generation, and conditional editing.

\textbf{Decoupled Capability Pathways.}
Although Lance organizes all modalities within a shared sequence, it processes understanding and generation through specialized expert pathways. The understanding expert $\mathrm{LLM}_{\mathrm{UND}}$ primarily operates on text tokens and semantic visual tokens, and autoregressively predicts target text tokens for multimodal understanding. Its hidden states are mapped by a language modeling head and optimized with the standard next-token prediction loss:
\begin{equation}
\mathcal{L}_{\mathrm{UND}}
= - \sum_i \log p_{\theta_{\mathrm{UND}}}(y_i \mid y_{<i}, \mathcal{S}).
\end{equation}
The generation expert $\mathrm{LLM}_{\mathrm{GEN}}$ operates on VAE latent tokens and predicts generation-side hidden states conditioned on the interleaved multimodal context. These hidden states are projected through an LLM-to-VAE connector into the latent space and passed to a flow prediction head. Let $x_1$ denote the clean VAE latent and $x_0 \sim \mathcal{N}(0, I)$ denote Gaussian noise. We construct the interpolated latent as $x_t = t x_1 + (1-t)x_0$, where $t \sim \mathcal{U}(0,1)$, and optimize the generation expert with:
\begin{equation}
\mathcal{L}_{\mathrm{GEN}} =
\mathbb{E}_{x_0, x_1, t}
\left[
\left\|
v_{\theta_{\mathrm{GEN}}}(x_t, \mathcal{S}, t) - (x_1 - x_0)
\right\|_2^2
\right].
\end{equation}
Here, $\theta_{\mathrm{UND}}$ and $\theta_{\mathrm{GEN}}$ denote the pathway-specific parameters for understanding and generation, respectively, including their Transformer-decoder expert backbones and corresponding prediction heads.

The overall objective is:
\begin{equation}
\mathcal{L}
= \lambda_u \mathcal{L}_{\mathrm{UND}}
+ \lambda_g \mathcal{L}_{\mathrm{GEN}}.
\end{equation}

This design enables Lance to preserve unified context interaction while allowing semantic understanding and visual synthesis to specialize in their own representations, parameters, and objectives.


\subsection{Modality-Aware Rotary Positional Encoding}

\noindent
\begin{minipage}[t]{0.5\columnwidth} 
\setlength{\parindent}{1em}
\noindent Unified multimodal training places heterogeneous visual token groups within the same interleaved sequence, including ViT semantic tokens, clean VAE condition tokens, and noisy VAE target tokens. These tokens differ not only in their source encoders, but also in their functional roles: semantic tokens provide language-aligned visual cues for understanding, clean VAE latents serve as visual conditions, and noisy VAE latents are optimized as generation targets. Standard 3D-RoPE can encode spatiotemporal layouts, but it does not explicitly distinguish these heterogeneous token groups, which may lead to positional ambiguity and weaken cross-task alignment.

\noindent In the original 3D-RoPE formulation of Qwen2.5-VL~\cite{Qwen2.5-VL}, visual tokens follow a unified spatiotemporal positional assignment. Let $D$ denote the starting position index of a visual token group. For visual tokens with temporal length $T$, height $H$, and width $W$, a token at location $(t,h,w)$ is assigned a 3D position as:
\begin{equation}
\hat{\mathbf{p}}^{\mathrm{vis}}_{t,h,w}
= D + [t,\ h,\ w]
= [D+t,\; D+h,\; D+w],
\end{equation}
where $t \in [0,T-1]$, $h \in [0,H-1]$, and $w \in [0,W-1]$.
\end{minipage}
\hfill
\begin{minipage}[t]{0.44\columnwidth} 
    \centering
    \vspace{-0.5em}
    \includegraphics[width=0.95\linewidth]{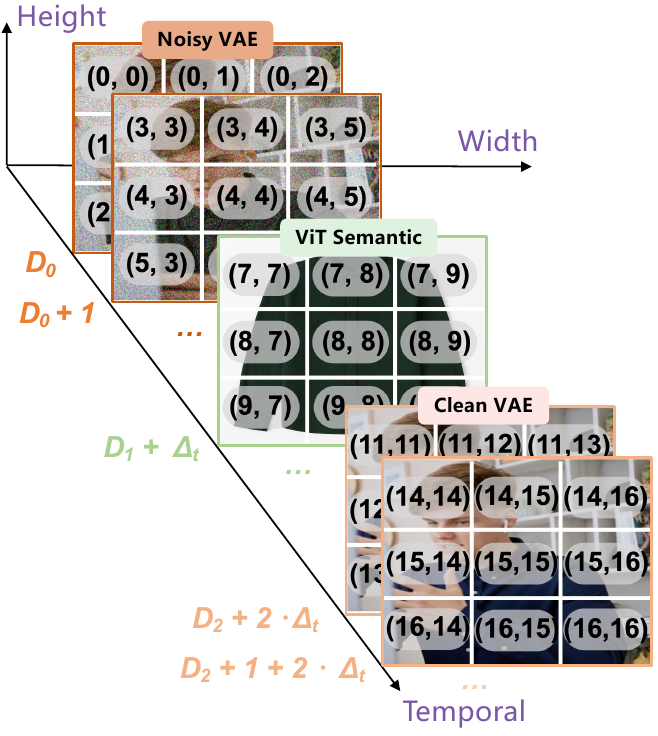}
    \captionof{figure}{\textbf{Illustration of modality-aware rotary positional encoding (MaPE).}
    where $D_i$ denotes the starting position index of the $i$-th modality token group along the temporal dimension.
    }
    \label{fig:mape}
\end{minipage}

\medskip
This design is effective for standard image/video-language modeling. However, in unified multimodal training, a single sequence may contain multiple visual token groups from different modalities $\mathcal{M}=\{\mathbf{V}_{\mathrm{vae}}^{\mathrm{noisy}} , \mathbf{V}_{\mathrm{vit}}, \mathbf{V}_{\mathrm{vae}}^{\mathrm{clean}} \}$.
 Assigning them only according to their spatiotemporal layouts may make their functional boundaries ambiguous in the positional space.

To address this issue, we introduce \textbf{Modality-Aware Rotary Positional Encoding} (MaPE), which injects token-group awareness into the positional indices. As shown in \Cref{fig:mape}, for $i$-th modality group $m_i \in \mathcal{M}$, we first define its base 3D-RoPE as
$\hat{\mathbf{p}}^{(m_i)}_{t,h,w}
=
[\hat{t}^{(m_i)}_{t,h,w},\;
 \hat{h}^{(m_i)}_{t,h,w},\;
 \hat{w}^{(m_i)}_{t,h,w}],$
where the base coordinates follow the standard spatiotemporal assignment. MaPE then applies a fixed offset step $\Delta_t$ along the temporal dimension according to the index $i$ of each modality group:
\begin{equation}
\mathbf{p}^{(m_i)}_{t,h,w}
=
\hat{\mathbf{p}}^{(m_i)}_{t,h,w}
+
[i \cdot \Delta_t, 0, 0]
=
[
\hat{t}^{(m_i)}_{t,h,w}+i \cdot \Delta_t,\;
\hat{h}^{(m_i)}_{t,h,w},\;
\hat{w}^{(m_i)}_{t,h,w}
],
\quad i \in \{0,1,2\}.
\end{equation}

Applying modality offsets only to the temporal dimension provides two advantages. First, it explicitly separates different visual token groups in the global positional space, enabling the model to better distinguish the roles of semantic ViT features, clean VAE conditions, and noisy VAE targets. Second, since the spatial coordinates remain unchanged, the intrinsic spatial layouts within images and videos are preserved. 
Moreover, introducing modality offsets $\Delta_t$ along the $t$-dimension does not disrupt the temporal structure within a video. Since the offset is a shared constant shift for all tokens within the same modality group, the temporal order and relative distances of video latents are fully preserved.
As a result, the model can better discriminate heterogeneous visual tokens while maintaining spatial consistency and temporal coherence.



\begin{table}[t]
\centering
\small
\setlength{\tabcolsep}{7pt}
\begin{tabular}{p{6.8cm}!{\vrule width 0.6pt}c c c c}
\toprule
& \textbf{PT} & \textbf{CT} & \textbf{SFT} & \textbf{RL} \\
\midrule

\textbf{Hyperparameters} & & & & \\
Learning rate & $1.0\times10^{-4}$ & $1.0\times10^{-4}$ & $2.5\times10^{-5}$ & $2.0\times10^{-6}$ \\
LR scheduler & Constant & Constant & Cosine & Constant \\
Weight decay & 0.0 & 0.0 & 0.0 & 0.0 \\
Gradient norm clip & 1.0 & 1.0 & 1.0 & 1.0 \\
Optimizer & \multicolumn{4}{c}{AdamW ($\beta_1=0.9,\ \beta_2=0.95,\ \epsilon=1.0\times10^{-15}$)} \\
Loss weight (CE : MSE) & 0.25 : 1 & 0.5 : 1 & 0.25 : 1  & $-$  \\
Warm-up steps & 2500 & 2500 & 500 & 50 \\
Training steps & 350k & 80k & 15k & 800 \\
Sequence length per rank (min, max) & (44K, 50K) & (74K, 80K) & (74K, 80K) & (74K, 80K) \\
\# Seen training tokens & 1.5T & 300B & 72B & 0.5B \\
Max context window & 40k & 70k & 70k & 70k \\
Gen resolution (min short side, max long side) & (192, 848) & (480, 848) & (480, 848) & (480, 848) \\
Und resolution (min short side, max long side) & (168, 826) & (462, 826) & (462, 826) & (462, 826) \\
Diffusion timestep shift & 1.0 & 4.0 & 4.0 & 4.0 \\
\bottomrule
\end{tabular}
\caption{\textbf{Training hyperparameters of Lance.} 
}
\label{Tab_train}
\end{table}

\begin{table}[t]
\centering
\caption{\textbf{Training data mixture schedule of Lance.}
Img., Vid., Gen., and Und. denote image, video, generation, and understanding, respectively.
CT is divided into three stages that progressively increase the proportion of challenging generation tasks.}
\label{tab:data_mixture_schedule}
\setlength{\tabcolsep}{7pt}
\renewcommand{\arraystretch}{1.15}
\resizebox{1\linewidth}{!}{
\begin{tabular}{ll!{\vrule width 0.6pt}ccccc}
\toprule
\textbf{Mixture} & \textbf{Ratio Type} 
& \textbf{PT} & \textbf{CT-I} & \textbf{CT-II} & \textbf{CT-III} & \textbf{SFT} \\
\midrule

\multirow{1}{*}{\textbf{Global}}


& \textbf{Vid.-Gen. : Vid.-Und. : Img.-Gen. : Img.-Und.} 
& $64:16:16:4$
& $64:16:16:4$
& $64:16:16:4$
& $64:16:16:4$
& $64:16:16:4$ \\

\midrule

\multirow{2}{*}{\textbf{Generation}}
& \textbf{T2I : I-Edit : S2I} 
& $100:0:0$
& $70 : 15:15$
& $60:20:20$
& $50:25:25$
& $60:20:20$ \\

& \textbf{T2V : I2V : V-Edit : S2V} 
& $100:0:0:0$
& $60 : 10 : 15:15$
& $40:20:20:20$
& $25:25:25:25$
& $60 : 10 : 15:15$ \\

\bottomrule
\end{tabular}
}
\end{table}


\begin{table*}[t]
\centering
\small
\setlength{\tabcolsep}{5pt}
\renewcommand{\arraystretch}{1.15}
\begin{tabular}{lcp{7.5cm}cc}
\toprule
\textbf{Output Type} & \textbf{Notation} & \textbf{Task} & \textbf{\# Samples} & \textbf{Phases} \\
\midrule

\multirow{3}{*}{\textbf{Text}}
& I2T  & General image captioning                     & 1B    & PT, CT \\
& V2T  & General video captioning                     & 140M  & PT, CT \\
\rowcolor{gray!20}
& I2T  & High-quality image captioning                & 190K  & SFT \\
\rowcolor{gray!20}
& V2T  & High-quality video captioning                & 5K    & SFT \\
\rowcolor{gray!20}
& X2T  & Interleaved multimodal understanding         & 2.73M & CT, SFT \\
\midrule

\multirow{5}{*}{\textbf{Image}}
& T2I  & General image generation                     & 1B    & PT, CT \\
& X2I & General image editing             & 2.8M  & CT \\
& X2I  & General subject-driven image generation          & 3.6M  & CT \\
\rowcolor{gray!20}
& T2I  & High-quality image generation                & 190K  & SFT \\
\rowcolor{gray!20}
& X2I & High-quality image editing                   & 84K   & SFT \\
\midrule

\multirow{5}{*}{\textbf{Video}}
& T2V/I2V  & General video generation                     & 140M  & PT, CT \\
& X2V  & General video editing            & 2.6M  & CT \\
& X2V  & General subject-driven video generation         & 1M    & CT \\
\rowcolor{gray!20}
& T2V/I2V  & High-quality video generation                & 5K    & SFT \\
\rowcolor{gray!20}
& X2V  & High-quality video editing                   & 9K    & SFT \\
\rowcolor{gray!20}
& X2V  & High-quality subject-driven video generation & 5.5K &  SFT \\
\bottomrule
\end{tabular}
\caption{\textbf{Summary of task categories and sample statistics for Lance.}
Within each output type, high-quality data are listed separately and highlighted in gray. ``Phases'' indicates the training phase(s) where each data type is applied. 
}
\label{tab:task_data_summary}
\end{table*}

\section{Training and Data}
\label{sec:training_data}

Lance adopts a staged multi-task training strategy to progressively develop and balance multimodal understanding and generation within a unified task formulation. As shown in \Cref{Tab_train}, the pipeline consists of four stages: PT establishes basic image/video understanding and generation from large-scale paired data; CT expands the task space with interleaved multi-task data and promotes cross-task transfer; SFT refines instruction following, visual fidelity, editing accuracy, and identity consistency with curated supervision; and RL further optimizes image generation with task-specific rewards. The data mixture schedule and task statistics are summarized in \Cref{tab:data_mixture_schedule,tab:task_data_summary}.

\subsection{Pre-Training Stage (PT)}

\textbf{Training Objectives.}
The pre-training stage establishes preliminary multimodal alignment and basic visual generation capabilities. To this end, we freeze the VAE and ViT encoders and optimize the remaining components, including the multimodal backbone, QK-Norm modules, and MLP connectors.

\textbf{Pre-Training Data.}
The PT stage is trained on large-scale image-text and video-text pairs, organized around paired captioning and conditional generation tasks. The image-text subset comprises approximately $1$B samples spanning diverse visual domains, including natural scenes, human-centric, object-centric, knowledge-oriented, and stylized content. The video-text subset comprises approximately $140$M samples and covers diverse dynamic scenarios, including actions, events, scene transitions, and long-range temporal processes. 
To improve scalability, we adopt a progressive resolution curriculum of $192$p \(\rightarrow\) $360$p \(\rightarrow\) $480$p, with dynamic resolution enabled at each stage. In addition, we use an image:video sampling ratio of approximately $1:4$ to account for the greater difficulty of video modeling and to strengthen temporal reasoning and generation.

\vspace{0.5em}

\begin{minipage}{\textwidth}
\centering
\small
\setlength{\parskip}{0pt}

\begin{promptfigbox}{System Prompt for I2T/V2T captioning tasks}
{\ttfamily
\textless|im\_start|\textgreater system\\[1pt]
Generate a detailed and accurate description of the \{{\color{blue}image}/{\color{orange}{video}}\}, including all the visual details {\color{orange}{\{and key moments\}}}.\textless|im\_end|\textgreater\\[1pt]
\textless|im\_start|\textgreater user\\[1pt]
\textless|vision\_start|\textgreater {{\color{red}\textless|user\_vision|\textgreater}}\textless|vision\_end|\textgreater\textless|im\_end|\textgreater\\[1pt]
\textless|im\_start|\textgreater assistant
}
\end{promptfigbox}

\vspace{-0.3em}

\begin{promptfigbox}{System Prompt for other I2T/V2T tasks}
{\ttfamily
\textless|im\_start|\textgreater system\\[1pt]
View the \{{\color{blue}{image}}/{\color{orange}{video}}\} attentively and provide a suitable answer to the posed question.\textless|im\_end|\textgreater\\[1pt]
\textless|im\_start|\textgreater user\\[1pt]
\textless|vision\_start|\textgreater {{\color{red}\textless|user\_vision|\textgreater}}\textless|vision\_end|\textgreater
{{\color{red}\textless|user\_text|\textgreater}}\textless|im\_end|\textgreater\\[1pt]
\textless|im\_start|\textgreater assistant
}
\end{promptfigbox}

\vspace{-0.5em}

\captionof{figure}{\textbf{System prompts for understanding tasks.} {\color{red}{Red}} placeholders denote user-provided text and visual inputs.}
\label{fig:prompt-und}

\end{minipage}

\begin{figure*}[t]
\centering
\begin{minipage}{\textwidth}
\centering
\small
\setlength{\parskip}{0pt}

\begin{promptfigbox}{System Prompt for T2I/T2V tasks}
{\ttfamily
\textless|im\_start|\textgreater system\\[1pt]
Describe the \{{\color{blue}image}/{\color{orange}{video}}\} by detailing the color, quantity, text, shape, size, texture, spatial relationships {\color{orange}{\{and motion/camera movements\}}} of the objects and background:\textless|im\_end|\textgreater\\[1pt]
\textless|im\_start|\textgreater user\\[1pt]
{\color{red}\bfseries \textless|user\_text|\textgreater}\textless|im\_end|\textgreater\\[1pt]
\textless|im\_start|\textgreater assistant
}
\end{promptfigbox}

\vspace{-0.3em}

\begin{promptfigbox}{System Prompt for other X2I/X2V tasks}
{\ttfamily
\textless|im\_start|\textgreater system\\[1pt]
Describe the key features of the input \{{\color{blue}image}/{\color{orange}{video}}\} (color, shape, size, texture, objects, background), then explain how the user’s text instruction should alter or modify the \{{\color{blue}image}/{\color{orange}{video}}\}. Generate a new \{{\color{blue}image}/{\color{orange}{video}}\} that meets the user’s requirements while maintaining consistency with the original input where appropriate.\textless|im\_end|\textgreater\\[1pt]
\textless|im\_start|\textgreater user\\[1pt]
\textless|vision\_start|\textgreater {{\color{red}\textless|user\_vision|\textgreater}}\textless|vision\_end|\textgreater
{{\color{red}\textless|user\_text|\textgreater}}\textless|im\_end|\textgreater\\[1pt]
\textless|im\_start|\textgreater assistant
}
\end{promptfigbox}

\vspace{-0.5em}

\end{minipage}
\caption{\textbf{System prompts for generation tasks.} {\color{red}{Red}} placeholders denote user-provided text and visual inputs.}
\label{fig:prompt-gen}
\end{figure*}

\subsection{Continual Training Stage (CT)}

\textbf{Training Objectives.}
The continual training stage extends the PT model from basic paired supervision to unified multi-task multimodal learning. By introducing richer interleaved multimodal data and more diverse input-output mappings, CT expands the task space and improves task-aware multimodal generalization.

\textbf{Continual Training Data.}
During CT, we progressively introduce a broader set of tasks for both understanding and generation. For understanding, we incorporate $2.73$M interleaved multimodal understanding samples, covering pure text understanding (T2T, $41$K), captioning ($443$K), classification ($142$K), conversation ($72$K), grounding ($200$K), reasoning ($194$K), VQA ($600$K), and OCR ($120$K). For generation, we incorporate large-scale any-to-image/video data, including $2.8$M image editing samples and $2.6$M video editing samples, together with $3.6$M subject-driven image generation samples and $1$M subject-driven video generation samples. To accommodate the increased task diversity, we adopt a progressive data-mixture strategy that gradually increases the sampling ratio of more challenging tasks, such as editing and subject-driven generation, while correspondingly reducing the proportion of simpler caption-style supervision (detailed in \Cref{tab:data_mixture_schedule}). In total, the CT stage consumes approximately $300$B training tokens.

\textbf{Task-specific System Prompts.}
To better distinguish heterogeneous tasks within a unified multimodal context, we further introduce task-specific \textit{system prompts} for understanding and generation tasks, as illustrated in \Cref{fig:prompt-und} and \Cref{fig:prompt-gen}. These prompts provide explicit task priors and guide task-specific input-output formats while preserving unified sequence modeling.

\subsection{Supervised Fine-Tuning Stage (SFT)}

\textbf{Training Objectives.}
The supervised fine-tuning stage refines the model with high-quality, task-aligned supervision under a reduced learning rate. Unlike PT and CT, which focus on capability acquisition and task expansion, SFT emphasizes instruction fidelity, visual consistency, editing accuracy, and identity preservation, improving controllability and downstream task performance.

\textbf{Supervised Fine-Tuning Data.}
The SFT stage uses curated high-quality data spanning both understanding and generation tasks. For understanding, we use $190$K high-quality image captioning samples, $5$K high-quality video captioning samples, together with $2.73$M interleaved multimodal understanding samples for continued instruction refinement. For image generation, we include $190$K high-quality image generation samples and $84$K high-quality image editing samples. For video generation, we further incorporate $5$K high-quality video generation samples, $9$K high-quality video editing samples, and $5.5$K high-quality subject-driven video generation samples. Compared with the large-scale corpora used in PT and CT, these curated data provide stronger task alignment and higher annotation quality, and thus offer more precise supervision for improving instruction following and generation fidelity.

\subsection{Reinforcement Learning Stage}

\textbf{Training Objectives.}
The reinforcement learning stage further refines the model's image generation capability by directly optimizing generation behavior with task-specific rewards. Unlike SFT, which learns from static supervised targets through maximum likelihood, RL uses Group Relative Policy Optimization (GRPO) to encourage outputs that better satisfy fine-grained textual constraints. In particular, this stage focuses on improving text rendering accuracy, image-text correspondence, and prompt compositional adherence.

\textbf{Reinforcement Learning Data.}
The RL stage uses $20$K image generation prompts that emphasize fine-grained text-related requirements. During optimization, PaddleOCR \cite{cui2025paddleocr} serves as the reward model to evaluate the consistency between the generated image and the textual constraints specified in the prompt. This reward provides direct feedback on text rendering quality and text-image alignment, helping improve aspects that are difficult to fully capture with supervised fine-tuning alone.

\section{Experiments}
\label{sec:Experiments}
\subsection{Experimental Setup}

\begin{table*}[!t]
\centering
\small
\setlength{\tabcolsep}{2.25pt}
\renewcommand{\arraystretch}{1.08}

\newcommand{\metricssep}{\rule[-0.35em]{0.35pt}{1.55em}}

\resizebox{\textwidth}{!}{
\begin{tabular}{@{}l c c c c c c c c c c c c c c c c@{}}
\toprule
\multirow{2}{*}{\textbf{Models}}
& \multirow{2}{*}{\textbf{Params.}}
& \metricssep
& \multicolumn{6}{c}{\textbf{DPG-Bench}}
& \metricssep
& \multicolumn{7}{c}{\textbf{GenEval}} \\
\cmidrule(lr){4-9} \cmidrule(lr){11-17}
&
& \metricssep
& \textbf{Global} & \textbf{Entity} & \textbf{Attribute} & \textbf{Relation} & \textbf{Other} & \textbf{Overall}
& \metricssep
& \textbf{1-Obj.} & \textbf{2-Obj.} & \textbf{Count} & \textbf{Colors} & \textbf{Position} & \textbf{Attr.} & \textbf{Overall} \\
\midrule

\multicolumn{17}{@{}>{\columncolor{gray!12}}c@{}}{\textit{Generation-only Models}} \\

PixArt-$\alpha$ \cite{chen2024pixart}
& 0.6B
& \metricssep
& 74.97 & 79.32 & 78.60 & 82.57 & 76.96 & 71.11
& \metricssep
& 0.98 & 0.50 & 0.44 & 0.80 & 0.08 & 0.07 & 0.48 \\

SDXL \cite{podell2024sdxl}
& 3.5B
& \metricssep
& 83.27 & 82.43 & 80.91 & 86.76 & 80.41 & 74.65
& \metricssep
& 0.98 & 0.74 & 0.39 & 0.85 & 0.15 & 0.23 & 0.55 \\

Hunyuan-DiT \cite{li2024hunyuan}
& 1.5B
& \metricssep
& 84.59 & 80.59 & 88.01 & 74.36 & 86.41 & 78.87
& \metricssep
& -- & -- & -- & -- & -- & -- & -- \\


DALL-E 3 \cite{betker2023improving}
& --
& \metricssep
& 90.97 & 89.61 & 88.39 & 90.58 & 89.83 & 83.50
& \metricssep
& 0.96 & 0.87 & 0.47 & 0.83 & 0.43 & 0.45 & 0.67 \\

SD3-Medium \cite{esser2024scaling}
& 2B
& \metricssep
& 87.90 & 91.01 & 88.83 & 80.70 & 88.68 & 84.08
& \metricssep
& 0.99 & 0.94 & 0.72 & 0.89 & 0.33 & 0.60 & 0.74 \\

Emu3-Gen \cite{wang2024emu3}
& 8B
& \metricssep
& 85.21 & 86.68 & 86.84 & 90.22 & 83.15 & 80.60
& \metricssep
& 0.98 & 0.71 & 0.34 & 0.81 & 0.17 & 0.21 & 0.54 \\

FLUX.1-dev$^\dagger$ \cite{blackforestlabs_flux}
& 12B
& \metricssep
& 74.35 & 90.00 & 88.96 & 90.87 & 88.33 & 83.84
& \metricssep
& 0.98 & 0.93 & 0.75 & 0.93 & 0.68 & 0.65 & 0.82 \\

GPT Image 1 \cite{openai2025gptimage1}
& --
& \metricssep
& -- & -- & -- & -- & -- & --
& \metricssep
& 0.99 & 0.92 & 0.85 & 0.92 & 0.75 & 0.61 & 0.84 \\

Qwen-Image \cite{wu2025qwen}
& 20B
& \metricssep
& 91.32 & 91.56 & 92.02 & 94.31 & 92.73 & 88.32
& \metricssep
& 0.99 & 0.92 & 0.89 & 0.88 & 0.76 & 0.77 & 0.87 \\

\midrule
\multicolumn{17}{@{}>{\columncolor{gray!12}}c@{}}{\textit{Unified Models}} \\


SEED-X \cite{ge2024seed}
& --
& \metricssep
& -- & -- & -- & -- & -- & --
& \metricssep
& 0.97 & 0.58 & 0.26 & 0.80 & 0.19 & 0.14 & 0.49 \\

TokenFlow-XL \cite{qu2025tokenflow}
& --
& \metricssep
& -- & -- & -- & -- & -- & --
& \metricssep
& 0.95 & 0.60 & 0.41 & 0.81 & 0.16 & 0.24 & 0.55 \\


Janus \cite{wu2025janus}
& --
& \metricssep
& 82.33 & 87.38 & 87.70 & 85.46 & 86.41 & 79.68
& \metricssep
& 0.97 & 0.68 & 0.30 & 0.84 & 0.46 & 0.42 & 0.61 \\


Emu3-Gen$^\dagger$ \cite{wang2024emu3}
& 8B
& \metricssep
& -- & -- & -- & -- & -- & 81.60
& \metricssep
& \underline{0.99} & 0.81 & 0.42 & 0.80 & 0.49 & 0.45 & 0.66 \\

Show-o \cite{xie2024show}
& --
& \metricssep
& -- & -- & -- & -- & -- & --
& \metricssep
& 0.98 & 0.80 & 0.66 & 0.84 & 0.31 & 0.50 & 0.68 \\

Janus-Pro-7B \cite{chen2025janus}
& 7B
& \metricssep
& 86.90 & 88.90 & 89.40 & 89.32 & 89.48 & 84.19
& \metricssep
& \underline{0.99} & 0.89 & 0.59 & 0.90 & 0.79 & 0.66 & 0.80 \\


Ovis-U1 \cite{wang2025ovis}
& 1.2B
& \metricssep
& 82.37 & 90.08 & 88.68 & \underline{93.35} & 85.20 & 83.72
& \metricssep
& -- & -- & -- & -- & -- & -- & -- \\

OmniGen2 \cite{wu2025omnigen2}
& 4B
& \metricssep
& 88.81 & 88.83 & 90.18 & 89.37 & 90.27 & 83.57
& \metricssep
& \textbf{1.00} & 0.95 & 0.64 & 0.88 & 0.55 & 0.76 & 0.80 \\

Show-o2 \cite{xie2025show}
& 7B
& \metricssep
& 89.00 & \textbf{91.78} & 89.96 & 91.81 & \textbf{91.64} & 86.14
& \metricssep
& \textbf{1.00} & 0.87 & 0.58 & 0.92 & 0.52 & 0.62 & 0.76 \\

UniWorld-V1 \cite{lin2025uniworld}
& 13B
& \metricssep
& 83.64 & 88.39 & 88.44 & 89.27 & 87.22 & 81.38
& \metricssep
& \underline{0.99} & 0.93 & 0.79 & 0.89 & 0.49 & 0.70 & 0.80 \\

BAGEL$^\dagger$ \cite{deng2025emerging}
& 7B
& \metricssep
& 88.94 & 90.37 & \underline{91.29} & 90.82 & 88.67 & 85.07
& \metricssep
& 0.98 & 0.95 & \textbf{0.84} & \underline{0.95} & 0.78 & 0.77 & 0.88 \\

Mogao \cite{liao2025mogao}
& 7B
& \metricssep
& 82.37 & 90.03 & 88.26 & 93.18 & 85.40 & 84.33
& \metricssep
& \textbf{1.00} & \textbf{0.97} & \underline{0.83} & 0.93 & 0.84 & 0.80 & \underline{0.89} \\

InternVL-U \cite{tian2026internvlu}
& 1.7B
& \metricssep
& \underline{90.39} & 90.78 & 90.68 & 90.29 & 88.77 & 85.18
& \metricssep
& \underline{0.99} & 0.94 & 0.74 & 0.91 & 0.77 & 0.74 & 0.85 \\

TUNA \cite{liu2025tuna}
& 7B
& \metricssep
& \textbf{90.42} & \underline{91.68} & 90.94 & 91.87 & \underline{90.73} & \textbf{86.76}
& \metricssep
& \textbf{1.00} & \textbf{0.97} & 0.81 & 0.91 & \textbf{0.88} & \textbf{0.83} & \textbf{0.90} \\

TUNA-2 \cite{tuna2}
& 7B
& \metricssep
& 89.50 & 91.40 & \textbf{92.07} & 91.91 & 88.81 & \underline{86.54}
& \metricssep
& \underline{0.99} & \underline{0.96} & 0.80 & 0.91 & 0.84 & 0.76 & 0.87 \\

\rowcolor{rowblue}
\textbf{Lance (Ours)}
& \textbf{3B}
& \metricssep
& 83.89 & 91.07 & 89.36 & \textbf{93.38} & 80.80 & 84.67
& \metricssep
& \textbf{1.00} & 0.94 & \textbf{0.84} & \textbf{0.97} & \underline{0.87} & \underline{0.81} & \textbf{0.90} \\

\bottomrule
\end{tabular}
}
\caption{\textbf{Image generation results on DPG-Bench and GenEval.}
$^\dagger$ refers to methods using LLM rewriters in GenEval.
\textbf{Bold}: best results among unified models.
\underline{Underline}: second-best among unified models.}
\label{tab:image_generation_combined}
\end{table*}

\begin{figure*}[!t]
    \centering
    \includegraphics[width=1\textwidth]{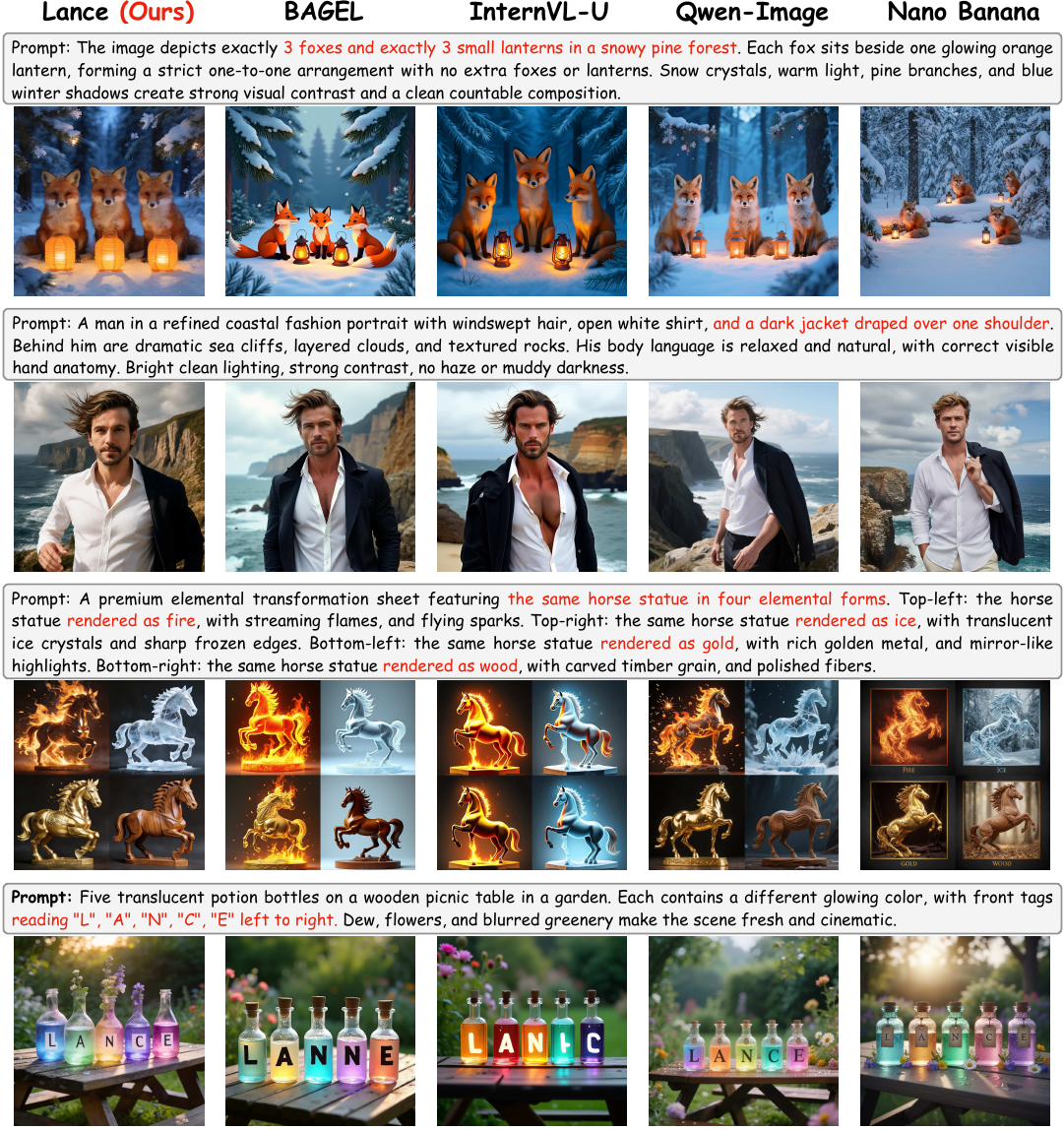} 
    \caption{\textbf{T2I qualitative comparison.} 
    Instructions that are correctly reflected in our results but missed or incorrectly rendered by some baseline models are highlighted in {\color{red}{\textbf{red}}}.
    }
    \label{fig:T2I-baseline}
\end{figure*}


Lance is implemented upon Qwen2.5-VL $3$B \cite{Qwen2.5-VL}, using its weights to initialize the visual understanding encoder and the multimodal context backbones $\mathrm{LLM}_{\mathrm{UND}}$ and $\mathrm{LLM}_{\mathrm{GEN}}$. 
To stabilize large-scale unified multi-task training, each attention block in $\mathrm{LLM}_{\mathrm{UND}}$ and $\mathrm{LLM}_{\mathrm{GEN}}$ is equipped with QK-Norm \cite{dehghani2023scaling}. 
Since QK-Norm changes the original query-key activation distribution of the initialized backbone, Lance is not a direct reuse of the pretrained Qwen2.5-VL model, but is substantially re-adapted through the proposed staged training recipe.
For the visual generation encoder, we adopt the 3D causal VAE encoder from Wan2.2 \cite{wan2025wan}, to support a unified processing of image and video modalities.
Following prior work \cite{ho2022classifier}, we also adopt classifier-free guidance (CFG) for visual and text conditions. During the PT stage, for text-to-image generation data, the text condition is dropped with a probability of $10\%$.
The offset step $\Delta_t$ used in MaPE is set to $1000$.
During the CT and SFT stages, for multimodal conditions, the full condition is dropped with a probability of $5\%$, while the text-only condition is additionally dropped with a probability of $5\%$ and the visual condition is retained.
During inference, the CFG scale for text conditions in generation tasks is set to $4$. Unless otherwise specified, the image input resolution is set to $768 \times 768$, while videos are sampled at $480p$ resolution with a frame rate of $12$ fps.

\subsection{Main Results}

\subsubsection{Image Generation}

\textbf{Quantitative Results.}
We evaluate the image generation capability of Lance on GenEval~\cite{ghosh2023geneval} and DPG-Bench~\cite{hu2024ella}. As shown in \Cref{tab:image_generation_combined}, Lance achieves top-tier performance among unified models on GenEval, matching the best overall score ($\textbf{0.90}$) while showing strong compositional ability on counting, colors, and spatial position. On DPG-Bench, Lance obtains competitive overall performance and performs particularly well on relation modeling, indicating its ability to preserve fine-grained semantic consistency under complex prompts. These results suggest that Lance can effectively support high-quality image synthesis within a unified multimodal framework, despite using only $3$B activated parameters.

\textbf{Qualitative Results.}
We conduct a qualitative comparison of Lance with $7$B Bagel \cite{deng2025emerging}, $1.7$B InternVL-U \cite{tian2026internvlu}, $20$B Qwen-Image  ~\cite{wu2025qwen} and Nano Banana \cite{Gemini3pro}. 
As shown in \Cref{fig:T2I-baseline}, compared with open-source unified multimodal baselines such as Bagel \cite{deng2025emerging} and InternVL-U \cite{tian2026internvlu}, Lance demonstrates stronger visual aesthetics and image-text alignment (\eg, lantern count in $1$-st case,
jacket draped over one shoulder in $2$-nd case).
Overall, Lance generates significantly higher-quality images than Bagel \cite{deng2025emerging} and InternVL-U \cite{tian2026internvlu}, and achieves comparable performance with the $20$B large-scale model Qwen-Image  ~\cite{wu2025qwen} and the commercial closed-source model Nano Banana \cite{Gemini3pro}.

\begin{table*}[!t]
\centering
\small
\setlength{\tabcolsep}{3.0pt}
\renewcommand{\arraystretch}{1.10}

\newcommand{\vbenchsep}{\rule[-0.35em]{0.35pt}{1.55em}}

\resizebox{0.9\textwidth}{!}{
\begin{tabular}{@{}l c c *{10}{c}@{}}
\toprule
\multicolumn{13}{c}{\textbf{(a) VBench Metrics Part I}} \\
\midrule
\multirow{1}{*}{\textbf{Models}}
& \multirow{1}{*}{\textbf{Params.}}
& \vbenchsep
& \makecell[c]{\textbf{Quality}\\\textbf{Score}}
& \makecell[c]{\textbf{Semantic}\\\textbf{Score}}
& \makecell[c]{\textbf{Subj.}\\\textbf{Consist.}}
& \makecell[c]{\textbf{Bkg.}\\\textbf{Consist.}}
& \makecell[c]{\textbf{Temp.}\\\textbf{Flicker.}}
& \makecell[c]{\textbf{Motion}\\\textbf{Smooth.}}
& \makecell[c]{\textbf{Dynamic}\\\textbf{Degree}}
& \makecell[c]{\textbf{Aesthetic}\\\textbf{Quality}}
& \makecell[c]{\textbf{Imaging}\\\textbf{Quality}}
& \makecell[c]{\textbf{Object}\\\textbf{Class}} \\
\midrule

\multicolumn{13}{@{}>{\columncolor{gray!12}}c@{}}{\textit{Generation-only Models}} \\

ModelScope \cite{wang2023modelscope}
& 1.7B
& \vbenchsep
& 78.05 & 66.54 & 89.87 & 95.29 & 98.28 & 95.79 & 66.39 & 52.06 & 58.57 & 82.25 \\

LaVie \cite{wang2025lavie}
& 3B
& \vbenchsep
& 78.78 & 70.31 & 91.41 & 97.47 & 98.30 & 96.38 & 49.72 & 54.94 & 61.90 & 91.82 \\

Show-1 \cite{zhang2025show}
& 6B
& \vbenchsep
& 80.42 & 72.98 & 95.53 & 98.02 & 99.12 & 98.24 & 44.44 & 57.35 & 58.66 & 93.07 \\

AnimateDiff-V2 \cite{guo2023animatediff}
& --
& \vbenchsep
& 82.90 & 69.75 & 95.30 & 97.68 & 98.75 & 97.76 & 40.83 & 67.16 & 70.10 & 90.90 \\

VideoCrafter-2.0 \cite{chen2024videocrafter2}
& --
& \vbenchsep
& 82.20 & 73.42 & 96.85 & 98.22 & 98.41 & 97.73 & 42.50 & 63.13 & 67.22 & 92.55 \\

CogVideoX \cite{yang2024cogvideox}
& 5B
& \vbenchsep
& 82.75 & 77.04 & 96.23 & 96.52 & 98.66 & 96.92 & 70.97 & 61.98 & 62.90 & 85.23 \\

Kling \cite{Kling2024}
& --
& \vbenchsep
& 83.39 & 75.68 & 98.33 & 97.60 & 99.30 & 99.40 & 46.94 & 61.21 & 65.62 & 87.24 \\

Open-Sora-2.0 \cite{opensora2}
& --
& \vbenchsep
& 82.10 & 80.14 & 98.75 & 98.00 & 99.40 & 99.49 & 20.74 & 64.33 & 65.62 & 94.50 \\

Gen-3 \cite{RunwayGen32024}
& --
& \vbenchsep
& 84.11 & 75.17 & 97.10 & 96.62 & 98.61 & 99.23 & 60.14 & 63.34 & 66.82 & 87.81 \\

Step-Video-T2V \cite{ma2025step}
& 30B
& \vbenchsep
& 84.46 & 71.28 & 98.05 & 97.67 & 99.40 & 99.08 & 53.06 & 61.23 & 70.63 & 80.56 \\

HunyuanVideo \cite{wu2025hunyuanvideo}
& --
& \vbenchsep
& 85.07 & 76.88 & 97.22 & 97.60 & 99.39 & 99.05 & 71.94 & 60.28 & 67.24 & 83.48 \\

Wan2.1-T2V \cite{wan2025wan}
& 14B
& \vbenchsep
& 85.59 & 76.11 & 97.52 & 98.09 & 99.46 & 98.30 & 65.46 & 66.07 & 69.43 & 86.28 \\

\midrule
\multicolumn{13}{@{}>{\columncolor{gray!12}}c@{}}{\textit{Unified Models}} \\

HaploOmni \cite{xiao2025haploomni}
& 7B
& \vbenchsep
& -- & -- & \underline{96.40} & \underline{97.60} & -- & 96.80 & 65.30 & -- & -- & -- \\

Emu3 \cite{wang2024emu3}
& 8B
& \vbenchsep
& -- & -- & 95.32 & \textbf{97.69} & -- & \textbf{98.93} & \textbf{79.27} & 59.64 & -- & 86.17 \\

VILA-U \cite{wu2024vila}
& 7B
& \vbenchsep
& 76.26 & 65.04 & -- & -- & -- & -- & -- & -- & -- & -- \\

Show-o2 \cite{xie2025show}
& 2B
& \vbenchsep
& 82.10 & 78.31 & \textbf{97.28} & 96.78 & 97.68
& 98.25 & 40.83 & \underline{65.15} & \textbf{67.06} & 94.81 \\

TUNA \cite{liu2025tuna}
& 1.5B
& \vbenchsep
& \underline{84.32} & \underline{83.04} & 95.99 & 96.72 & \underline{98.02}
& \underline{98.33} & 69.39 & \textbf{65.88} & \underline{66.83} & \underline{95.41} \\

\rowcolor{rowblue}
\textbf{Lance (Ours)}
& \textbf{3B}
& \vbenchsep
& \textbf{85.14} & \textbf{84.96} & 94.52 & 94.28 & \textbf{99.66}
& 95.93 & \underline{75.83} & 64.33 & 66.78 & \textbf{96.58} \\

\bottomrule
\end{tabular}}

\vspace{0.6em}

\resizebox{0.9\textwidth}{!}{
\begin{tabular}{@{}l c c *{9}{c}@{}}
\toprule
\multicolumn{12}{c}{\textbf{(b) VBench Metrics Part II}} \\
\midrule
\multirow{1}{*}{\textbf{Models}}
& \multirow{1}{*}{\textbf{Params.}}
& \vbenchsep
& \makecell[c]{\textbf{Multi.}\\\textbf{Objects}}
& \makecell[c]{\textbf{Human}\\\textbf{Action}}
& \makecell[c]{\textbf{Color}}
& \makecell[c]{\textbf{Spatial}\\\textbf{Relation}}
& \makecell[c]{\textbf{Scene}}
& \makecell[c]{\textbf{Appear.}\\\textbf{Style}}
& \makecell[c]{\textbf{Temp.}\\\textbf{Style}}
& \makecell[c]{\textbf{Overall}\\\textbf{Consist.}}
& \makecell[c]{\textbf{Total}\\\textbf{Score}$\uparrow$} \\
\midrule

\multicolumn{12}{@{}>{\columncolor{gray!12}}c@{}}{\textit{Generation-only Models}} \\

ModelScope \cite{wang2023modelscope}
& 1.7B
& \vbenchsep
& 38.98 & 92.40 & 81.72 & 33.68 & 39.26 & 23.39 & 25.37 & 25.67 & 75.75 \\

LaVie \cite{wang2025lavie}
& 3B
& \vbenchsep
& 33.32 & 96.80 & 86.39 & 34.09 & 52.69 & 23.56 & 25.93 & 26.41 & 77.08 \\

Show-1 \cite{zhang2025show}
& 6B
& \vbenchsep
& 45.47 & 95.60 & 86.35 & 53.50 & 47.03 & 23.06 & 25.28 & 27.46 & 78.93 \\

AnimateDiff-V2 \cite{guo2023animatediff}
& --
& \vbenchsep
& 36.88 & 92.60 & 87.47 & 34.60 & 50.19 & 22.42 & 26.03 & 27.04 & 80.27 \\

VideoCrafter-2.0 \cite{chen2024videocrafter2}
& --
& \vbenchsep
& 40.66 & 95.00 & 92.92 & 35.86 & 55.29 & 25.13 & 25.84 & 28.23 & 80.44 \\

CogVideoX \cite{yang2024cogvideox}
& 5B
& \vbenchsep
& 62.11 & 99.40 & 82.81 & 66.35 & 53.20 & 24.91 & 25.38 & 27.59 & 81.61 \\

Kling \cite{Kling2024}
& --
& \vbenchsep
& 68.05 & 93.40 & 89.90 & 73.03 & 50.86 & 19.62 & 24.17 & 26.42 & 81.85 \\

Open-Sora-2.0 \cite{opensora2}
& --
& \vbenchsep
& 77.72 & 95.40 & 85.98 & 76.18 & 52.71 & 22.98 & 25.91 & 27.57 & 81.71 \\

Gen-3 \cite{RunwayGen32024}
& --
& \vbenchsep
& 53.64 & 96.40 & 80.90 & 65.09 & 54.57 & 24.31 & 24.71 & 26.69 & 82.32 \\

Step-Video-T2V \cite{ma2025step}
& 30B
& \vbenchsep
& 50.55 & 94.00 & 88.25 & 71.47 & 24.38 & 23.17 & 26.01 & 27.12 & 81.83 \\

HunyuanVideo \cite{wu2025hunyuanvideo}
& --
& \vbenchsep
& 66.71 & 94.40 & 89.79 & 72.13 & 54.46 & 22.21 & 24.52 & 26.95 & 83.43 \\

Wan2.1-T2V \cite{wan2025wan}
& 14B
& \vbenchsep
& 69.58 & 95.40 & 88.59 & 75.39 & 45.75 & 22.64 & 23.19 & 25.91 & 83.69 \\

\midrule
\multicolumn{12}{@{}>{\columncolor{gray!12}}c@{}}{\textit{Unified Models}} \\

HaploOmni \cite{xiao2025haploomni}
& 7B
& \vbenchsep
& -- & -- & -- & -- & 34.60 & -- & -- & -- & 78.10 \\

Emu3 \cite{wang2024emu3}
& 8B
& \vbenchsep
& 44.64 & 77.71 & -- & 68.73 & 37.11 & 20.92 & -- & -- & 80.96 \\

VILA-U \cite{wu2024vila}
& 7B
& \vbenchsep
& -- & -- & -- & -- & -- & -- & -- & -- & 74.01 \\

Show-o2 \cite{xie2025show}
& 2B
& \vbenchsep
& 76.01 & 95.20 & 80.89 & 62.61 & 57.67 & \textbf{23.29} & \underline{25.27} & 27.00 & 81.34 \\

TUNA \cite{liu2025tuna}
& 1.5B
& \vbenchsep
& \underline{92.31} & \underline{97.50} & \underline{87.67} & \underline{78.12} & \underline{58.59}
& \underline{23.18} & 24.68 & \textbf{27.71} & \underline{84.06} \\

\rowcolor{rowblue}
\textbf{Lance (Ours)}$^\dagger$
& \textbf{3B}
& \vbenchsep
& \textbf{93.86} & \textbf{97.80} & \textbf{92.61} & \textbf{93.61} & \textbf{64.75}
& 23.14 & \textbf{25.53} & \underline{27.04} & \textbf{85.11} \\
\bottomrule
\end{tabular}}
\caption{\textbf{Video generation results on VBench.}
$^\dagger$ refers to methods using LLM rewriters.
\textbf{Bold}: best results among unified models.
\underline{Underline}: second-best among unified models.}
\label{tab:vbench_full}
\end{table*}

\begin{figure*}[!t]
    \centering
    \includegraphics[width=0.96\textwidth]{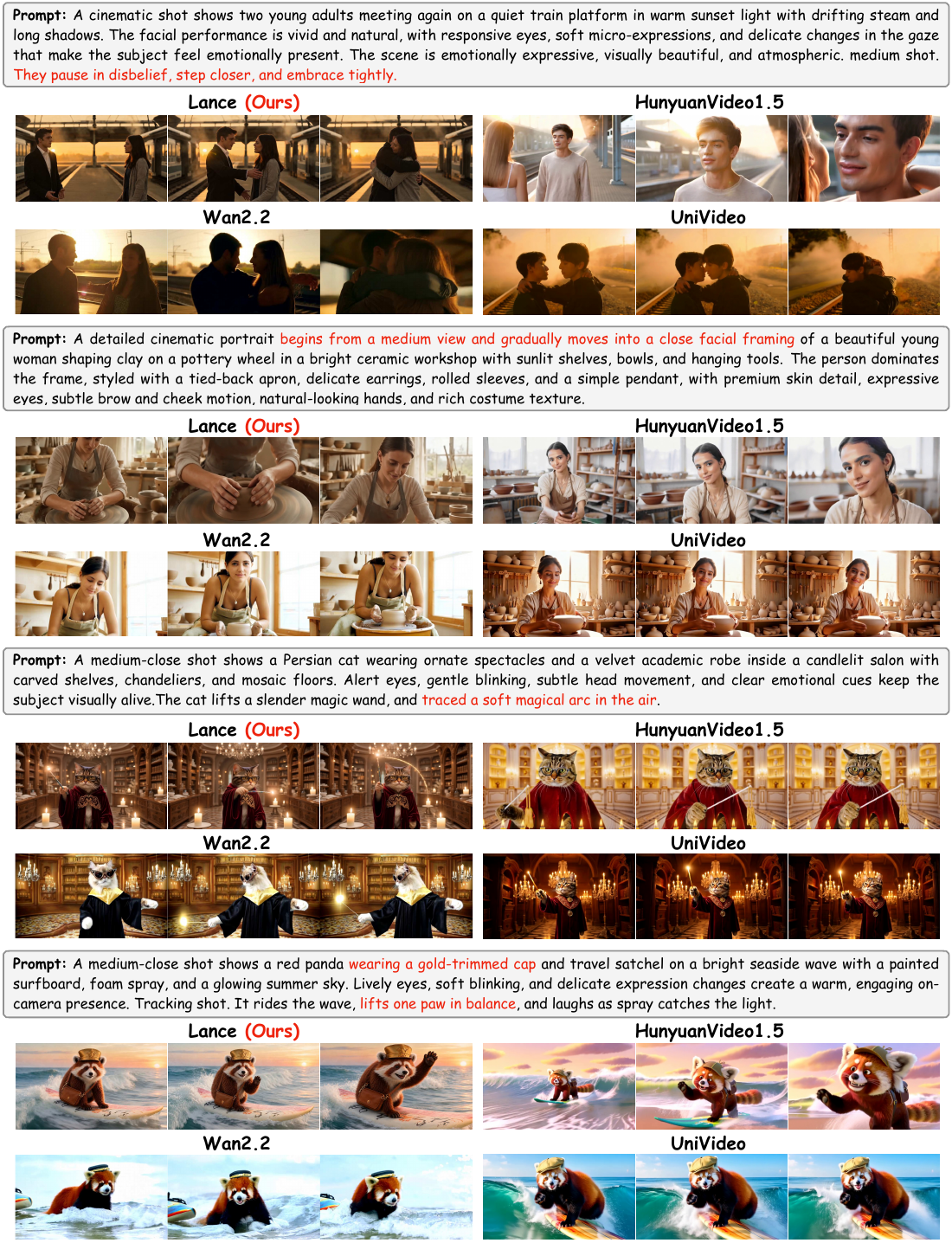}
    \caption{\textbf{T2V qualitative comparison.}
    Instructions that are correctly reflected in our results but missed or incorrectly rendered by some baseline models are highlighted in {\color{red}{\textbf{red}}}.
    }
    \label{fig:T2V-baseline}
\end{figure*}

\subsubsection{Video Generation}

\textbf{Quantitative Results.}
We evaluate the text-to-video generation capability of Lance on VBench~\cite{huang2024vbench}. As shown in \Cref{tab:vbench_full}, Lance achieves the best Total Score ($\textbf{85.11}$) among unified models with only $3$B activated parameters. Beyond the overall score, Lance also shows strong performance across both quality-oriented and semantic-oriented dimensions, including visual quality, object grounding, color consistency, spatial relationships, scene understanding, and temporal style. These results indicate that the proposed unified framework effectively supports compositional video generation and text-video alignment, while scaling naturally from image generation to more challenging spatiotemporal generation tasks.

\textbf{Qualitative Results.}
We conduct a qualitative comparison between Lance and $8.3$B HunyuanVideo1.5 \cite{wu2025hunyuanvideo}, $5$B Wan2.2-TI2V \cite{wan2025wan}, and $7$B UniVideo \cite{wei2025univideo}.
As shown in \Cref{fig:T2V-baseline}, the generated videos exhibit strong semantic fidelity, coherent motion, and appealing visual quality. In challenging cases involving complex human interactions (\eg, $1$-st case, ``two adults hugging"), or explicit camera transitions (\eg, $2$-nd case, from a ``medium view" to ``close facial framing"), our model follows the prompt accurately and produces videos with stable visual texture and consistent temporal evolution. These examples further demonstrate the effectiveness of the unified architecture for high-quality text-to-video generation.

\begin{table*}[!t]
\centering
\small
\setlength{\tabcolsep}{3.0pt}
\renewcommand{\arraystretch}{1.10}

\newcommand{\geditsep}{\rule[-0.35em]{0.35pt}{1.55em}}

\resizebox{\textwidth}{!}{
\begin{tabular}{@{}l c c *{12}{c}@{}}
\toprule

\multirow{2}{*}{\textbf{Models}}
& \multirow{2}{*}{\textbf{Params.}}
& \geditsep
& \multicolumn{12}{c}{\textbf{GEdit-Bench}} \\
\cmidrule(lr){4-15}
&
& \geditsep
& \textbf{BC} & \textbf{CA} & \textbf{MM} & \textbf{MC} & \textbf{PB} & \textbf{ST}
& \textbf{SA} & \textbf{SR} & \textbf{SRp} & \textbf{TM} & \textbf{TT} & \textbf{Avg/G\_O} \\
\midrule

\multicolumn{15}{@{}>{\columncolor{gray!12}}c@{}}{\textit{Generation-only Models}} \\

Gemini 2.0 \cite{team2024gemini}
& --
& \geditsep
& -- & -- & -- & -- & -- & -- & -- & -- & -- & -- & -- & 6.32 \\

GPT Image 1 \cite{openai2025gptimage1}
& --
& \geditsep
& 6.96 & 6.85 & 7.10 & 5.41 & 6.74 & 7.44 & 7.51 & 8.73 & 8.55 & 8.45 & 8.69 & 7.49 \\

Qwen-Image-Edit \cite{wu2025qwen}
& 20B
& \geditsep
& 8.23 & 8.30 & 7.33 & 8.05 & 7.49 & 6.74 & 8.57 & 8.09 & 8.29 & 8.48 & 8.50 & 8.01 \\

\midrule
\multicolumn{15}{@{}>{\columncolor{gray!12}}c@{}}{\textit{Unified Models}} \\

Lumina-DiMOO \cite{xin2025lumina}
& 8B
& \geditsep
& 3.43 & 4.27 & 3.08 & 2.77 & 4.74 & 5.19 & 4.44 & 3.80 & 4.38 & 2.68 & 4.20 & 3.91 \\

Ovis-U1 \cite{wang2025ovis}
& 1.2B
& \geditsep
& \underline{7.49} & 6.88 & 6.21 & 4.79 & 5.98 & \underline{6.46}
& 7.49 & \underline{7.25} & \underline{7.27} & 4.48 & 6.31 & 6.42 \\

BAGEL \cite{deng2025emerging}
& 7B
& \geditsep
& 7.32 & 6.91 & 6.38 & 4.75 & 4.57 & 6.15
& \textbf{7.90} & 7.16 & 7.02 & \underline{7.32} & 6.22 & 6.52 \\

InternVL-U \cite{tian2026internvlu}
& 1.7B
& \geditsep
& 7.08 & 7.05 & 6.38 & \underline{7.02} & \underline{6.03} & 6.27
& 7.13 & 6.55 & 6.33 & 6.59 & \underline{6.85} & 6.66 \\

InternVL-U (w/ CoT) \cite{tian2026internvlu}
& 1.7B
& \geditsep
& 7.05 & \textbf{7.87} & \underline{6.50} & 6.99 & 5.77 & 6.10
& 7.33 & 7.16 & 7.12 & \textbf{7.36} & 6.46 & \underline{6.88} \\

\rowcolor{rowblue}
\textbf{Lance (Ours)}
& \textbf{3B}
& \geditsep
& \textbf{7.73} & \underline{7.74} & \textbf{7.28} & \textbf{7.83} & \textbf{7.50} & \textbf{7.03}
& \underline{7.64} & \textbf{7.85} & \textbf{7.71} & 4.46 & \textbf{7.57} & \textbf{7.30} \\

\bottomrule
\end{tabular}
}
\caption{\textbf{
Image editing results on GEdit-Bench.}
\textbf{Bold}: best results among unified models.
\underline{Underline}: second-best among unified models.}
\label{tab:gedit_bench}
\end{table*}

\begin{figure}[!t]
    \centering
    \includegraphics[width=\linewidth]{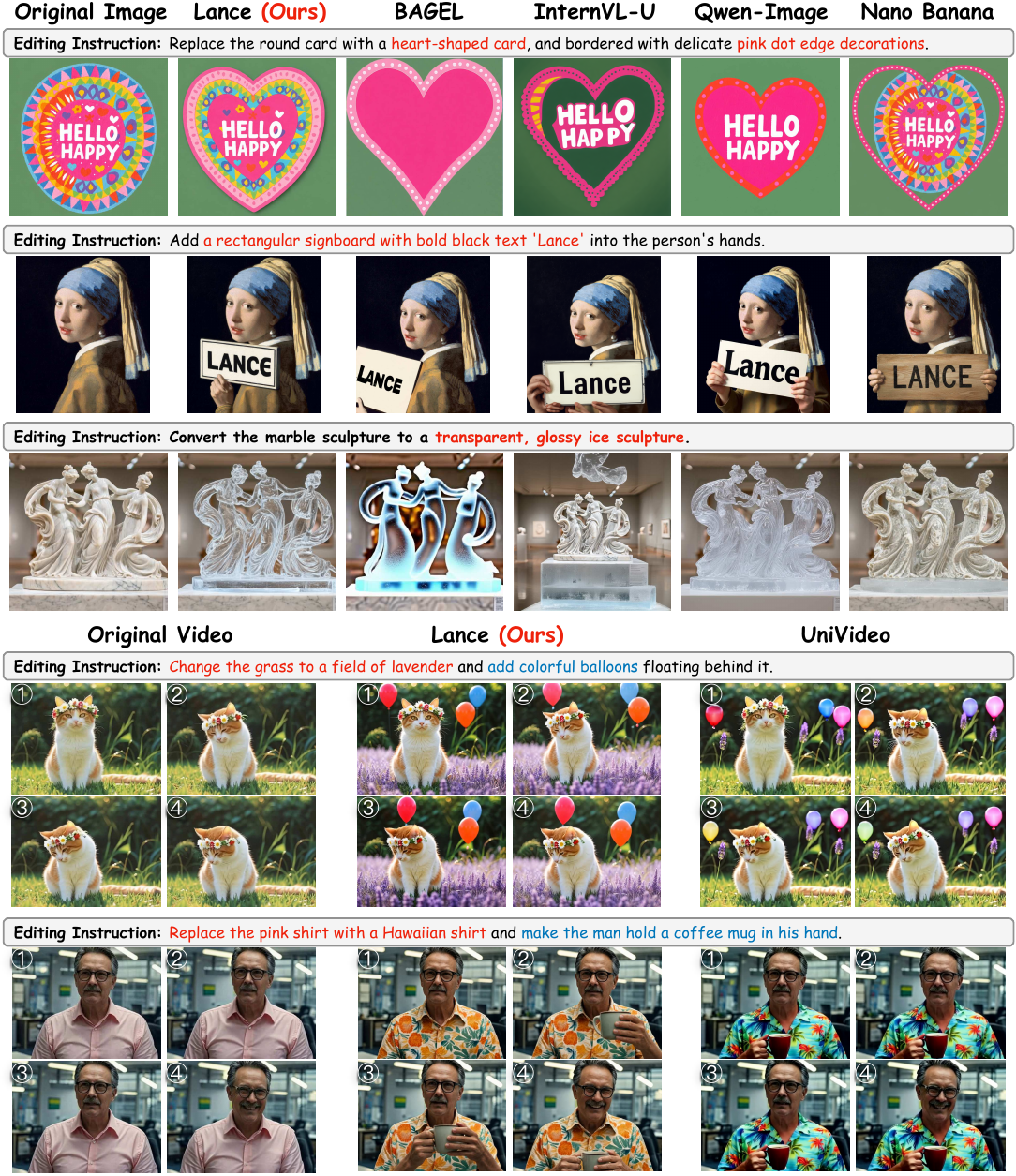}
    \caption{\textbf{Multimodal editing qualitative comparison.}    
    Lance performs precise image editing with realistic texture and structure preservation, and supports temporally coherent video editing with natural motion dynamics.}
    \label{fig:multimodal_editing_qual}
\end{figure}

\subsubsection{Multimodal Editing}

\textbf{Quantitative Results.}
We evaluate the image editing capability of our model on GEdit-Bench \cite{liu2025step1x}. As shown in \Cref{tab:gedit_bench}, our model achieves the best Avg/G$\_$O score (7.30) among unified models, demonstrating strong overall editing performance under a compact parameter budget. In particular, our model obtains the best results in several key editing categories, including background change, material modification, motion change, portrait beautification, subject removal, replacement, and tone transfer.  These results suggest that the proposed unified framework can effectively support a broad range of image editing operations. We also observe that Lance is relatively weaker on text modification, indicating that text-specific editing remains an important direction for future improvement.

\textbf{Qualitative Results.}
We further provide qualitative results for both image and video editing in \Cref{fig:multimodal_editing_qual}.
For image editing, Lance achieves visually coherent image editing with well-preserved structures and realistic textures, \eg, the plausible hand geometry and fine details in the $2$-nd case. For video editing, Lance performs accurate multi-attribute modifications while maintaining natural motion dynamics, such as the temporally consistent hand movement of the person holding a cup in the last case. Overall, these results demonstrate Lance's high-fidelity editing ability in both spatial realism and temporal coherence, highlighting the potential of unified models for multimodal editing.

\begin{table*}[!t]
\centering
\small
\setlength{\tabcolsep}{2.4pt}
\renewcommand{\arraystretch}{1.10}

\providecommand{\mvbenchsep}{\rule[-0.35em]{0.35pt}{1.55em}}

\resizebox{\textwidth}{!}{
\begin{tabular}{@{}l c c *{20}{c}@{}}
\toprule

\multirow{2}{*}{\textbf{Models}}
& \multirow{2}{*}{\textbf{Params.}}
& \mvbenchsep
& \multicolumn{20}{c}{\textbf{MVBench}} \\
\cmidrule(lr){4-23}
&
& \mvbenchsep
& \textbf{AS} & \textbf{AP} & \textbf{AA} & \textbf{FA} & \textbf{UA}
& \textbf{OE} & \textbf{OI} & \textbf{OS} & \textbf{MD} & \textbf{AL}
& \textbf{ST} & \textbf{AC} & \textbf{MC} & \textbf{MA} & \textbf{SC}
& \textbf{CO} & \textbf{EN} & \textbf{ER} & \textbf{CI} & \textbf{Avg.$\uparrow$} \\
\midrule

\multicolumn{23}{@{}>{\columncolor{gray!12}}c@{}}{\textit{Understanding-only Models}} \\

Video-LLaMA \cite{zhang-etal-2023-video}
& 7B
& \mvbenchsep
& 27.5 & 25.5 & 51.0 & 29.0 & 39.0
& 48.0 & 40.5 & 38.0 & 22.5 & 22.5
& 43.0 & 34.0 & 22.5 & 32.5 & 45.5
& 40.0 & 30.0 & 21.0 & 37.0 & 34.1 \\

LLaMA-Adapter \cite{zhang2023llamaadapter}
& 7B
& \mvbenchsep
& 23.0 & 28.0 & 51.0 & 30.0 & 33.0
& 53.5 & 32.5 & 33.5 & 25.5 & 21.5
& 30.5 & 29.0 & 22.5 & 41.5 & 39.5
& 31.5 & 22.5 & 28.0 & 32.0 & 31.7 \\

Video-ChatGPT \cite{Maaz2023VideoChatGPT}
& 7B
& \mvbenchsep
& 23.5 & 26.0 & 62.0 & 22.5 & 26.5
& 54.0 & 28.0 & 40.0 & 23.0 & 20.0
& 31.0 & 30.5 & 25.5 & 39.5 & 48.5
& 33.0 & 29.5 & 26.0 & 35.5 & 32.7 \\

VideoChat \cite{li2025videochat}
& 7B
& \mvbenchsep
& 33.5 & 26.5 & 56.0 & 33.5 & 40.5
& 53.0 & 40.5 & 30.0 & 25.5 & 27.0
& 48.5 & 35.0 & 20.5 & 42.5 & 46.0
& 41.0 & 23.5 & 23.5 & 36.0 & 35.5 \\

VideoChat2 \cite{li2024mvbench}
& 7B
& \mvbenchsep
& 66.0 & 47.5 & 83.5 & 49.5 & 60.0
& 58.0 & 71.5 & 42.5 & 23.0 & 23.0
& 88.5 & 39.0 & 42.0 & 58.5 & 44.0
& 36.5 & 35.0 & 40.5 & 65.5 & 51.1 \\

ST-LLM \cite{liu2024st}
& 7B
& \mvbenchsep
& 66.0 & 53.5 & 84.0 & 44.0 & 58.5
& 80.5 & 73.5 & 38.5 & 42.5 & 31.0
& 86.5 & 36.5 & 56.5 & 78.5 & 43.0
& 46.5 & 34.5 & 41.5 & 58.5 & 54.9 \\

GPT-4V \cite{openai2023gpt4v}
& --
& \mvbenchsep
& 55.5 & 63.5 & 72.0 & 46.5 & 73.5
& 18.5 & 59.0 & 29.5 & 12.0 & 40.5
& 83.5 & 39.0 & 12.0 & 22.5 & 45.0
& 52.0 & 31.0 & 59.0 & 11.0 & 43.5 \\

PLLaVA \cite{xu2024pllava}
& 34B
& \mvbenchsep
& 67.5 & 53.0 & 82.0 & 47.0 & 79.0
& 68.5 & 67.5 & 36.5 & 37.5 & 49.5
& 91.0 & 40.5 & 43.0 & 70.0 & 51.5
& 66.5 & 39.5 & 63.5 & 59.0 & 58.1 \\

Video-CCAM \cite{fei2024video}
& 9B
& \mvbenchsep
& 83.0 & 67.0 & 89.5 & 49.0 & 72.0
& 86.5 & 81.0 & 45.0 & 28.0 & 29.0
& 90.0 & 59.0 & 67.0 & 85.0 & 63.5
& 77.0 & 34.0 & 73.5 & 59.0 & 64.6 \\

Qwen2.5-VL \cite{Qwen2.5-VL}
& 3B
& \mvbenchsep
& -- & -- & -- & -- & --
& -- & -- & -- & -- & --
& -- & -- & -- & -- & --
& -- & -- & -- & -- & 67.0 \\

TimeMarker \cite{chen2024timemarker}
& 8B
& \mvbenchsep
& 79.0 & 74.5 & 89.0 & 53.5 & 77.0
& 94.0 & 76.0 & 41.5 & 52.5 & 47.0
& 91.5 & 53.0 & 76.5 & 92.5 & 57.0
& 70.5 & 23.5 & 53.5 & 82.5 & 67.4 \\

InternVideo2 \cite{wang2024internvideo2}
& 7B
& \mvbenchsep
& 86.0 & 70.0 & 87.0 & 56.0 & 75.0
& 91.0 & 86.0 & 40.0 & 48.0 & 53.0
& 90.0 & 41.0 & 73.0 & 92.0 & 52.0
& 56.0 & 33.0 & 57.0 & 74.0 & 67.3 \\

\midrule
\multicolumn{23}{@{}>{\columncolor{gray!12}}c@{}}{\textit{Unified Models}} \\

Show-o2 \cite{xie2025show}
& 1.5B
& \mvbenchsep
& \underline{63.8} & 59.5 & 63.5 & 40.0 & \underline{70.5}
& 54.5 & 66.0 & 36.5 & \underline{36.0} & 27.0
& \underline{88.0} & \underline{43.5} & 43.0 & 58.0 & \underline{44.5}
& \underline{54.0} & 28.5 & 39.5 & \underline{45.0} & 50.6 \\

Show-o2 \cite{xie2025show}
& 7B
& \mvbenchsep
& 60.1 & \underline{67.0} & 68.0 & 45.5 & \textbf{78.0}
& 51.0 & \textbf{73.5} & \textbf{44.5} & \underline{36.0} & \textbf{39.0}
& \textbf{92.5} & \textbf{51.5} & 36.0 & 59.5 & \textbf{52.0}
& \textbf{64.0} & \textbf{38.0} & \textbf{60.0} & 43.0 & \underline{55.7} \\

TUNA \cite{liu2025tuna}
& 1.5B
& \mvbenchsep
& -- & -- & -- & -- & --
& -- & -- & -- & -- & --
& -- & -- & -- & -- & --
& -- & -- & -- & -- & 54.4 \\

UniVideo \cite{wei2025univideo}
& 7B
& \mvbenchsep
& 54.3 & 41.5 & \textbf{77.5} & \textbf{50.0} & 62.5
& \underline{68.2} & 50.5 & \underline{37.5} & \underline{36.0} & 29.5
& 35.5 & 28.5 & \underline{52.5} & \underline{70.5} & 33.5
& 40.5 & \underline{37.5} & 36.5 & 38.0 & 46.3 \\

\rowcolor{rowblue}
\textbf{Lance (Ours)}
& \textbf{3B}
& \mvbenchsep
& \textbf{73.9} & \textbf{76.5} & \underline{71.5} & \underline{49.0} & 63.5
& \textbf{96.0} & \underline{72.5} & 33.0 & \textbf{63.5} & \underline{33.0}
& 86.0 & 41.0 & \textbf{82.0} & \textbf{97.5} & 43.0
& 47.5 & 31.5 & \underline{40.0} & \textbf{77.0} & \textbf{62.0} \\

\bottomrule
\end{tabular}
}
\caption{\textbf{Video understanding results on MVBench.}
\textbf{Bold}: best results among unified models.
\underline{Underline}: second-best among unified models.}
\label{tab:mvbench_main}
\end{table*}

\subsubsection{Multimodal Understanding}

\textbf{Quantitative Results.}
We evaluate the video understanding ability of Lance on MVBench \cite{li2024mvbench}, a widely used multi-choice benchmark for assessing temporal perception and video-centric understanding. 
As reported in \Cref{tab:mvbench_main}, Lance achieves the highest overall score (\textbf{62.0}) among existing unified multimodal models, with an approximately \textbf{11.3\%} relative improvement compared to the second-best unified model, Show-o2 7B \cite{xie2025show}. Lance also surpasses most of the specialized understanding models, with only half or even fewer parameters, indicating that unified multi-task training can preserve strong video understanding while enabling generation and editing capabilities.

\textbf{Qualitative Results.}
We present qualitative examples for image and video understanding in \Cref{fig:X2I,fig:X2V}.
Lance handles diverse understanding tasks, including OCR, knowledge-grounded reasoning, multi-image motion analysis, detailed video captioning, and action counting. The examples show that Lance can recognize fine-grained visual details, reason over static images, and capture temporal dynamics in videos. These results indicate that Lance maintains strong multimodal understanding ability while jointly supporting generation and editing within a unified model.

\section{Ablation Study}


\begin{figure*}[!t]
    \centering
    \includegraphics[width=1\textwidth]{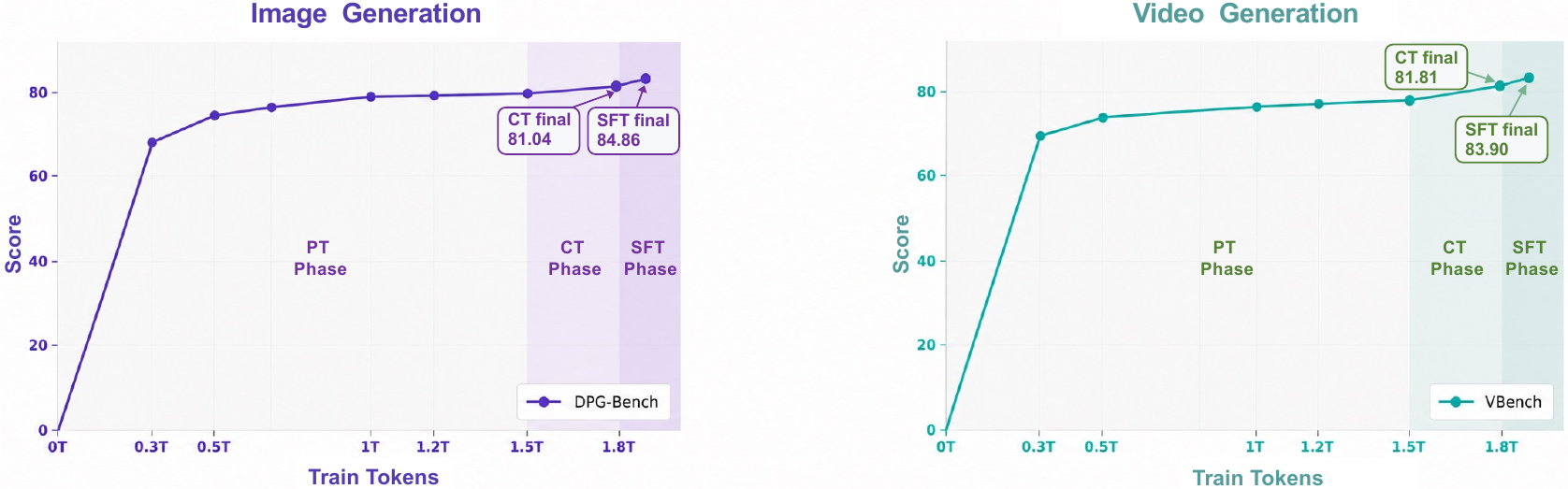}
    \caption{
    \textbf{Scaling behavior of image and video generation performance with increasing training tokens.}
    We report DPG-Bench for image generation and VBench for video generation across different training token budgets. 
    }
    \label{fig:token_scaling_curve}
\end{figure*}

\begin{figure*}[!t]
    \centering
    \includegraphics[width=1\textwidth]{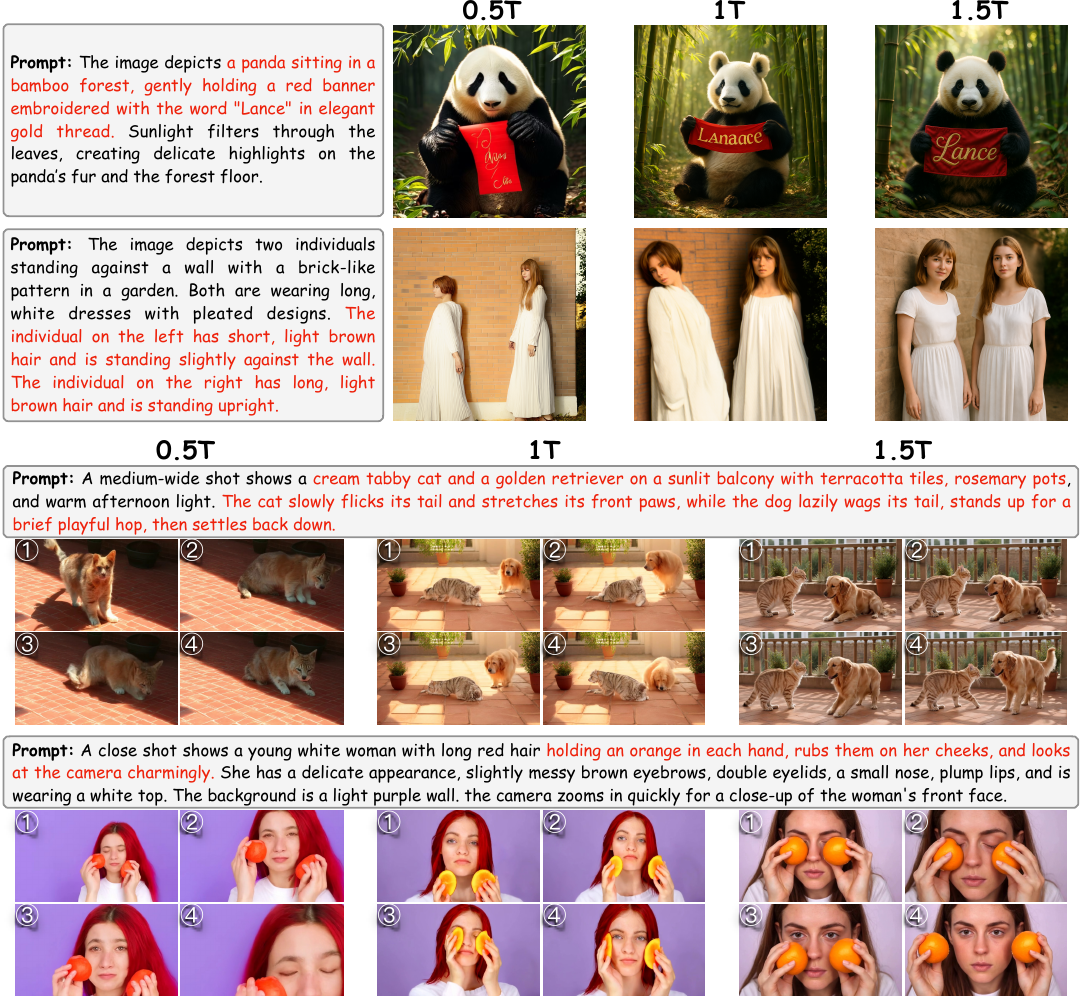}
    \caption{
    \textbf{Comparison of model variants trained with different token budgets.}
    We present qualitative cases of text-to-image and video generation using model variants trained with $0.5$T, $1$T, and $1.5$T tokens. As the training budget increases, the model demonstrates improved prompt alignment, visual fidelity, and temporal consistency.
    }
    \label{fig:token_ablation}
\end{figure*}

\begin{table}[t]
\centering
\caption{\textbf{Ablation on cross-task data.} 
Gen. denotes base generation data, Und. denotes understanding data, and MT-Gen. denotes multi-task generation data, including editing, subject-driven generation, etc.
}
\label{tab:ablation_auxiliary_data_generation}
\setlength{\tabcolsep}{8pt}
\renewcommand{\arraystretch}{1.15}
\resizebox{1\linewidth}{!}{
\begin{tabular}{llccc}
\toprule
\multirow{2}{*}{\textbf{Ablation Type}} 
& \multirow{2}{*}{\textbf{Setting}}
& \multicolumn{1}{c}{\textbf{Image Generation}}
& \multicolumn{1}{c}{\textbf{Video Generation}}
& \multicolumn{1}{c}{\textbf{Video Understanding}} \\
\cmidrule(lr){3-3}
\cmidrule(lr){4-4}
\cmidrule(lr){5-5}
& 
& \textbf{GenEval $\uparrow$}
& \textbf{VBench $\uparrow$}
& \textbf{MVBench $\uparrow$} \\
\midrule

\rowcolor{rowblue}
\textbf{Base} 
& {Gen. only} 
& 80.88 
& 81.25 
& -- \\

\midrule

\multirow{2}{*}{\textbf{+ Understanding data}} 
& Gen.:Und. = 8:2 
& {81.65} 
& \underline{82.91} 
& 58.06 \\

& Gen.:Und. = 9:1 ({MT-Gen. Base})
& 80.93 
& 81.47 
& 57.99 \\

\midrule

\multirow{2}{*}{\textbf{+ Multi-task data}} 
& Gen.:Und. = 9:1, Gen.:MT-Gen. = 8:2  
& \underline{81.89} 
& 82.88 
& \textbf{59.18} \\

& Gen.:Und. = 9:1, Gen.:MT-Gen. = 6:4 
& \textbf{82.06} 
& \textbf{83.05} 
& \underline{58.95} \\

\bottomrule
\end{tabular}
}
\end{table}

\subsection{Training Dynamics Analysis}
To systematically analyze the evolution of model capabilities during training, we further conduct quantitative and qualitative evaluations of model variants under different training-token budgets.

\textbf{Quantitative Analysis.}
As shown in \Cref{fig:token_scaling_curve}, image and video generation exhibit broadly consistent scaling trends as training tokens increase, with rapid gains in the early PT stage followed by a slower-growth regime. This indicates that large-scale paired training first establishes core generation capability, while later tokens mainly refine prompt alignment, visual fidelity, and temporal consistency.
Moreover, the CT stage further improves native generation capability, even though it mainly introduces multi-task data such as editing and instruction-following data rather than additional pure generation data (\Cref{tab:task_data_summary}). These results suggest that multi-task integration not only strengthens editing and instruction-following behaviors, but also brings positive transfer to visual generation, further validating the role of multi-task synergy in enhancing unified multimodal modeling.

\textbf{Qualitative Analysis.}
\Cref{fig:token_ablation} shows visual results consistent with the quantitative trends. As the training budget increases from $0.5$T to $1.5$T, Lance progressively improves prompt alignment, visual fidelity, text rendering, and temporal coherence. Early models capture coarse semantics but still suffer from distorted text, inaccurate attributes, and unstable motion, while the $1.5$T model produces more faithful compositions and more coherent multi-object dynamics.

\subsection{Effect of Cross-Task Data Synergy}

We conduct ablation studies to further analyze how different task mixtures affect the generation and understanding ability of Lance, focusing on the effects of understanding data and multi-task generation data. 
The results are summarized in \Cref{tab:ablation_auxiliary_data_generation}.

\textbf{Effect of Understanding Data.}
Introducing understanding-oriented data brings clear gains when used at an appropriate ratio. In particular, the Gen.:Und. = $8:2$ setting improves both image and video generation, suggesting that understanding data provides useful semantic grounding for visual synthesis. 

\textbf{Effect of Multi-task Data.}
Multi-task generation data enhances the base generation capability via joint training. Both mixture ratios outperform the generation-only baseline, with Gen.:MT-Gen. = $6:4$ achieving the best overall results. 
More unexpectedly, the benefits are not limited to generation: incorporating multi-task generation data also improves video understanding.
These results suggest that multi-task synergy is not merely a simple accumulation of capabilities, but may serve as an important mechanism for unlocking the further potential of unified models through mutual reinforcement across tasks.


\subsection{Effect of Modality-Aware Rotary Positional Encoding}

We further ablate the proposed Modality-Aware Rotary Positional Encoding (MaPE) to verify its effectiveness in unified multimodal modeling. As shown in \Cref{tab:ablation_mape}, removing MaPE consistently degrades performance across generation, editing, and understanding. 
The improvement is especially clear on image editing (from $6.30$ to $6.86$), where the model needs to jointly reason over visual conditions and generation targets. This suggests that MaPE reduces positional ambiguity among heterogeneous visual token groups, leading to better cross-task contextual alignment and more stable visual synthesis.

\begin{table}[t]
\centering
\caption{\textbf{Ablation on Modality-Aware Rotary Positional Encoding (MaPE).} We report GenEval for image generation, GEdit for image editing, VBench for video generation, and MVBench for video understanding.}
\label{tab:ablation_mape}
\setlength{\tabcolsep}{8pt}
\renewcommand{\arraystretch}{1.15}
\resizebox{0.8\linewidth}{!}{
\begin{tabular}{lcccc}
\toprule
\multirow{2}{*}{\textbf{Setting}} 
& \multicolumn{1}{c}{\textbf{Image Generation}} 
& \multicolumn{1}{c}{\textbf{Image Editing}} 
& \multicolumn{1}{c}{\textbf{Video Generation}} & \multicolumn{1}{c}{\textbf{Video Understanding}} \\
\cmidrule(lr){2-2} \cmidrule(lr){3-3} \cmidrule(lr){4-4} \cmidrule(lr){5-5}
& \textbf{GenEval $\uparrow$} & \textbf{GEdit $\uparrow$} & \textbf{VBench $\uparrow$} &\textbf{MVBench $\uparrow$} \\
\midrule
\rowcolor{rowblue}
\textbf{w/ MaPE} & \textbf{80.94} & \textbf{6.86} & \textbf{81.81} &\textbf{59.16} \\
\textbf{w/o MaPE} & 80.56 & 6.30 & 80.95 & 59.02 \\
\bottomrule
\end{tabular}
}
\end{table}

\section{Conclusion, Limitations and Future Work}
\label{sec:Conclusion}

In this work, we present Lance, a lightweight native unified multimodal model for image and video understanding, generation, and editing. Our key finding is that multi-task synergy can effectively advance unified multimodal modeling, enabling diverse tasks to mutually enhance each other within a shared framework. To this end, Lance combines unified interleaved context modeling with decoupled capability pathways, allowing semantic understanding and visual synthesis to interact while preserving task-specific specialization. Extensive experiments demonstrate that Lance achieves strong performance across image generation, video generation, multimodal editing, and video understanding benchmarks. Notably, these results are obtained with only $3$B activated parameters and a maximum $128$-GPU training budget, showing that capable unified multimodal models can be built in a resource-efficient manner.

Lance opens several promising directions for future exploration. 
\begin{itemize}
    \item \textbf{Post-training:} More comprehensive video-aware reward models, together with reward-based optimization methods~\cite{liu2026flow,xue2025dancegrpo,zheng2025diffusionnft}, could provide stronger supervision for temporally coherent, visually appealing, and user-aligned generation.

    \item \textbf{Data Curation:} Although the RL stage partially improves text-related generation, the current pre-training and continued-training stages do not include dedicated text-rendering data, which leaves accurate and layout-consistent text rendering as a remaining limitation. Incorporating such data may further improve this capability.
        
    \item \textbf{Model Scaling:} Scaling model capacity, expert capacity, and context length may further improve Lance's overall capability and cross-task generalization.

    \item \textbf{Broader Modalities:} Incorporating audio, speech, 3D, depth, and embodied sensory signals would be a natural step toward general-purpose any-to-any multimodal intelligence.

    \item \textbf{Streaming Multimodal Interaction:}  Integrating streaming perception and generation mechanisms~\cite{huang2026self,wu2026stream,tu2026stream} could extend Lance toward real-time interaction and closed-loop multimodal agents.
\end{itemize}

We hope Lance can serve as a practical foundation for future research on efficient, scalable, and task-general unified multimodal systems.





\paragraph{Author Contributions.}
Fengyi Fu, Mengqi Huang, Shaojin Wu, Yufei Huo and Jianzhu Guo contributed to code development, algorithm design, model training, and evaluation.
Jianzhu Guo and Mengqi Huang initialized the codebase.
Fengyi Fu, Mengqi Huang, Jianzhu Guo and Shaojin Wu were involved in the pre-training, continued training, and supervised fine-tuning stages.
Yufei Huo was responsible for reinforcement learning training.
Yunsheng Jiang, Hao Li, and Yinghang Song contributed to the data infrastructure.
Jianzhu Guo led the overall project direction and supervision.
The remaining authors contributed through technical discussions and feedback.

\paragraph{Acknowledgments.}
We thank Zhuowei Chen, Gen Li, and other colleagues for their valuable discussions, suggestions, and support on Lance.


\clearpage

\bibliographystyle{plainnat}
\bibliography{main}

@String(CVPR= {IEEE Conf. Comput. Vis. Pattern Recog.})

@String(NIPS= {Adv. Neural Inform. Process. Syst.})

@String(ICLR = {Int. Conf. Learn. Represent.})

@String(AAAI = {AAAI})

@String(CVPR  = {CVPR})

@String(NIPS  = {NeurIPS})

@String(ICLR  = {ICLR})

@article{wu2024vila,
  title={Vila-u: a unified foundation model integrating visual understanding and generation},
  author={Wu, Yecheng and Zhang, Zhuoyang and Chen, Junyu and Tang, Haotian and Li, Dacheng and Fang, Yunhao and Zhu, Ligeng and Xie, Enze and Yin, Hongxu and Yi, Li and others},
  journal={arXiv preprint arXiv:2409.04429},
  year={2024}
}

@article{xiao2025haploomni,
  title={Haploomni: Unified single transformer for multimodal video understanding and generation},
  author={Xiao, Yicheng and Song, Lin and Yang, Rui and Cheng, Cheng and Xu, Zunnan and Zhang, Zhaoyang and Ge, Yixiao and Li, Xiu and Shan, Ying},
  journal={arXiv preprint arXiv:2506.02975},
  year={2025}
}

@article{li2024hunyuan,
  title={Hunyuan-dit: A powerful multi-resolution diffusion transformer with fine-grained chinese understanding},
  author={Li, Zhimin and Zhang, Jianwei and Lin, Qin and Xiong, Jiangfeng and Long, Yanxin and Deng, Xinchi and Zhang, Yingfang and Liu, Xingchao and Huang, Minbin and Xiao, Zedong and others},
  journal={arXiv preprint arXiv:2405.08748},
  year={2024}
}

@inproceedings{chen2024pixart,
  title={Pixart-$\sigma$: Weak-to-strong training of diffusion transformer for 4k text-to-image generation},
  author={Chen, Junsong and Ge, Chongjian and Xie, Enze and Wu, Yue and Yao, Lewei and Ren, Xiaozhe and Wang, Zhongdao and Luo, Ping and Lu, Huchuan and Li, Zhenguo},
  booktitle={European Conference on Computer Vision},
  pages={74--91},
  year={2024},
  organization={Springer}
}

@article{betker2023improving,
  title={Improving image generation with better captions},
  author={Betker, James and Goh, Gabriel and Jing, Li and Brooks, Tim and Wang, Jianfeng and Li, Linjie and Ouyang, Long and Zhuang, Juntang and Lee, Joyce and Guo, Yufei and others},
  journal={Computer Science. https://cdn. openai. com/papers/dall-e-3. pdf},
  volume={2},
  number={3},
  pages={8},
  year={2023}
}

@inproceedings{wang2025illume,
  title={Illume: Illuminating your llms to see, draw, and self-enhance},
  author={Wang, Chunwei and Lu, Guansong and Yang, Junwei and Huang, Runhui and Han, Jianhua and Hou, Lu and Zhang, Wei and Xu, Hang},
  booktitle={Proceedings of the IEEE/CVF International Conference on Computer Vision},
  pages={21612--21622},
  year={2025}
}

@misc{blackforestlabs_flux,
  author       = {Black Forest Labs},
  title        = {FLUX: Official Inference Repository for FLUX.1 Models},
  year         = 2024,
  url          = {https://github.com/black-forest-labs/flux},
  note         = {Accessed: 2025-02-07}
}

@inproceedings{rombach2022high,
  title={High-resolution image synthesis with latent diffusion models},
  author={Rombach, Robin and Blattmann, Andreas and Lorenz, Dominik and Esser, Patrick and Ommer, Bj{\"o}rn},
  booktitle={CVPR},
  pages={10684--10695},
  year={2022}
}

@misc{wu2024taiyidiffusionxl,
      title={Taiyi-Diffusion-XL: Advancing Bilingual Text-to-Image Generation with Large Vision-Language Model Support}, 
      author={Xiaojun Wu and Dixiang Zhang and Ruyi Gan and Junyu Lu and Ziwei Wu and Renliang Sun and Jiaxing Zhang and Pingjian Zhang and Yan Song},
      year={2024},
      eprint={2401.14688},
      archivePrefix={arXiv},
      primaryClass={cs.CL}
}

@inproceedings{podell2024sdxl,
    title={{SDXL}: Improving Latent Diffusion Models for High-Resolution Image Synthesis},
    author={Dustin Podell and Zion English and Kyle Lacey and Andreas Blattmann and Tim Dockhorn and Jonas M{\"u}ller and Joe Penna and Robin Rombach},
    booktitle={ICLR},
    year={2024},
    url={https://openreview.net/forum?id=di52zR8xgf}
}

@inproceedings{esser2024scaling,
  title={Scaling rectified flow transformers for high-resolution image synthesis},
  author={Esser, Patrick and Kulal, Sumith and Blattmann, Andreas and Entezari, Rahim and M{\"u}ller, Jonas and Saini, Harry and Levi, Yam and Lorenz, Dominik and Sauer, Axel and Boesel, Frederic and others},
  booktitle={ICML},
  year={2024}
}

@article{lipman2024flow,
  title={Flow matching guide and code},
  author={Lipman, Yaron and Havasi, Marton and Holderrieth, Peter and Shaul, Neta and Le, Matt and Karrer, Brian and Chen, Ricky TQ and Lopez-Paz, David and Ben-Hamu, Heli and Gat, Itai},
  journal={arXiv preprint arXiv:2412.06264},
  year={2024}
}

@inproceedings{ramesh2021zero,
  title={Zero-shot text-to-image generation},
  author={Ramesh, Aditya and Pavlov, Mikhail and Goh, Gabriel and Gray, Scott and Voss, Chelsea and Radford, Alec and Chen, Mark and Sutskever, Ilya},
  booktitle={ICML},
  pages={8821--8831},
  year={2021},
  organization={Pmlr}
}

@article{ding2021cogview,
  title={Cogview: Mastering text-to-image generation via transformers},
  author={Ding, Ming and Yang, Zhuoyi and Hong, Wenyi and Zheng, Wendi and Zhou, Chang and Yin, Da and Lin, Junyang and Zou, Xu and Shao, Zhou and Yang, Hongxia and others},
  journal={NIPS},
  volume={34},
  pages={19822--19835},
  year={2021}
}

@article{ho2020denoising,
  title={Denoising diffusion probabilistic models},
  author={Ho, Jonathan and Jain, Ajay and Abbeel, Pieter},
  journal={NIPS},
  volume={33},
  pages={6840--6851},
  year={2020}
}

@article{li2023blip,
  title={Blip-diffusion: Pre-trained subject representation for controllable text-to-image generation and editing},
  author={Li, Dongxu and Li, Junnan and Hoi, Steven},
  journal={Advances in Neural Information Processing Systems},
  volume={36},
  pages={30146--30166},
  year={2023}
}

@article{mao2024realcustom++,
  title={Realcustom++: Representing images as real-word for real-time customization},
  author={Mao, Zhendong and Huang, Mengqi and Ding, Fei and Liu, Mingcong and He, Qian and Zhang, Yongdong},
  journal={arXiv preprint arXiv:2408.09744},
  year={2024}
}

@article{cai2024diffusion_selfdistill,
  title={Diffusion self-distillation for zero-shot customized image generation},
  author={Cai, Shengqu and Chan, Eric and Zhang, Yunzhi and Guibas, Leonidas and Wu, Jiajun and Wetzstein, Gordon},
  journal={arXiv preprint arXiv:2411.18616},
  year={2024}
}

@inproceedings{peebles2023scalable,
  title={Scalable diffusion models with transformers},
  author={Peebles, William and Xie, Saining},
  booktitle={Proceedings of the IEEE/CVF international conference on computer vision},
  pages={4195--4205},
  year={2023}
}

@article{tian2024visual,
  title={Visual autoregressive modeling: Scalable image generation via next-scale prediction},
  author={Tian, Keyu and Jiang, Yi and Yuan, Zehuan and Peng, Bingyue and Wang, Liwei},
  journal={Advances in neural information processing systems},
  volume={37},
  pages={84839--84865},
  year={2024}
}

@article{kondratyuk2023videopoet,
  title={Videopoet: A large language model for zero-shot video generation},
  author={Kondratyuk, Dan and Yu, Lijun and Gu, Xiuye and Lezama, Jos{\'e} and Huang, Jonathan and Schindler, Grant and Hornung, Rachel and Birodkar, Vighnesh and Yan, Jimmy and Chiu, Ming-Chang and others},
  journal={arXiv preprint arXiv:2312.14125},
  year={2023}
}

@inproceedings{chang2022maskgit,
  title={Maskgit: Masked generative image transformer},
  author={Chang, Huiwen and Zhang, Han and Jiang, Lu and Liu, Ce and Freeman, William T},
  booktitle={Proceedings of the IEEE/CVF conference on computer vision and pattern recognition},
  pages={11315--11325},
  year={2022}
}

@inproceedings{esser2021taming,
  title={Taming transformers for high-resolution image synthesis},
  author={Esser, Patrick and Rombach, Robin and Ommer, Bjorn},
  booktitle={Proceedings of the IEEE/CVF conference on computer vision and pattern recognition},
  pages={12873--12883},
  year={2021}
}

@article{wu2025hunyuanvideo,
  title={Hunyuanvideo 1.5 technical report},
  author={Wu, Bing and Zou, Chang and Li, Changlin and Huang, Duojun and Yang, Fang and Tan, Hao and Peng, Jack and Wu, Jianbing and Xiong, Jiangfeng and Jiang, Jie and others},
  journal={arXiv preprint arXiv:2511.18870},
  year={2025}
}

@inproceedings{fu2026layeredit,
  title={LayerEdit: Disentangled Multi-Object Editing via Conflict-Aware Multi-Layer Learning},
  author={Fu, Fengyi and Huang, Mengqi and Zhang, Lei and Mao, Zhendong},
  booktitle={Proceedings of the AAAI Conference on Artificial Intelligence},
  volume={40},
  number={5},
  pages={4003--4011},
  year={2026}
}

@inproceedings{fu2025feededit,
  title={Feededit: Text-based image editing with dynamic feedback regulation},
  author={Fu, Fengyi and Zhang, Lei and Huang, Mengqi and Mao, Zhendong},
  booktitle={Proceedings of the Computer Vision and Pattern Recognition Conference},
  pages={2661--2670},
  year={2025}
}

@article{hong2022cogvideo,
  title={Cogvideo: Large-scale pretraining for text-to-video generation via transformers},
  author={Hong, Wenyi and Ding, Ming and Zheng, Wendi and Liu, Xinghan and Tang, Jie},
  journal={arXiv preprint arXiv:2205.15868},
  year={2022}
}

@article{yang2024cogvideox,
  title={Cogvideox: Text-to-video diffusion models with an expert transformer},
  author={Yang, Zhuoyi and Teng, Jiayan and Zheng, Wendi and Ding, Ming and Huang, Shiyu and Xu, Jiazheng and Yang, Yuanming and Hong, Wenyi and Zhang, Xiaohan and Feng, Guanyu and others},
  journal={arXiv preprint arXiv:2408.06072},
  year={2024}
}

@article{ma2025step,
  title={Step-video-t2v technical report: The practice, challenges, and future of video foundation model},
  author={Ma, Guoqing and Huang, Haoyang and Yan, Kun and Chen, Liangyu and Duan, Nan and Yin, Shengming and Wan, Changyi and Ming, Ranchen and Song, Xiaoniu and Chen, Xing and others},
  journal={arXiv preprint arXiv:2502.10248},
  year={2025}
}

@misc{Kling2024,
  author       = {{Kling AI}},
  title        = {Kling AI},
  year         = {2024},
  howpublished = {\url{https://klingai.kuaishou.com/}},
  note         = {Accessed: 2024-06-06}
}

@misc{Gemini3pro,
  author       = {{Google DeepMind}},
  title        = {{Gemini 3 Pro Image Model Card}},
  year         = {2025},
  month        = nov,
  howpublished = {\url{https://storage.googleapis.com/deepmind-media/Model-Cards/Gemini-3-Pro-Image-Model-Card.pdf}},
  note         = {Model card published: November 2025}
}

@article{liu2025step1x,
  title={Step1x-edit: A practical framework for general image editing},
  author={Liu, Shiyu and Han, Yucheng and Xing, Peng and Yin, Fukun and Wang, Rui and Cheng, Wei and Liao, Jiaqi and Wang, Yingming and Fu, Honghao and Han, Chunrui and others},
  journal={arXiv preprint arXiv:2504.17761},
  year={2025}
}

@article{opensora2,
    title={Open-Sora 2.0: Training a Commercial-Level Video Generation Model in 200k}, 
    author={Xiangyu Peng and Zangwei Zheng and Chenhui Shen and Tom Young and Xinying Guo and Binluo Wang and Hang Xu and Hongxin Liu and Mingyan Jiang and Wenjun Li and Yuhui Wang and Anbang Ye and Gang Ren and Qianran Ma and Wanying Liang and Xiang Lian and Xiwen Wu and Yuting Zhong and Zhuangyan Li and Chaoyu Gong and Guojun Lei and Leijun Cheng and Limin Zhang and Minghao Li and Ruijie Zhang and Silan Hu and Shijie Huang and Xiaokang Wang and Yuanheng Zhao and Yuqi Wang and Ziang Wei and Yang You},
    year={2025},
    journal={arXiv preprint arXiv:2503.09642},
}

@misc{RunwayGen32024,
  author       = {{Runway}},
  title        = {Introducing Gen-3 Alpha: A New Frontier for Video Generation},
  year         = {2024},
  month        = jun,
  howpublished = {\url{https://runwayml.com/research/introducing-gen-3-alpha}},
  note         = {Accessed: 2024-06-17}
}

@article{cao2025hunyuanimage,
  title={Hunyuanimage 3.0 technical report},
  author={Cao, Siyu and Chen, Hangting and Chen, Peng and Cheng, Yiji and Cui, Yutao and Deng, Xinchi and Dong, Ying and Gong, Kipper and Gu, Tianpeng and Gu, Xiusen and others},
  journal={arXiv preprint arXiv:2509.23951},
  year={2025}
}

@article{xie2025show,
  title={Show-o2: Improved native unified multimodal models},
  author={Xie, Jinheng and Yang, Zhenheng and Shou, Mike Zheng},
  journal={arXiv preprint arXiv:2506.15564},
  year={2025}
}

@article{tian2025unigen,
  title={UniGen-1.5: Enhancing Image Generation and Editing through Reward Unification in Reinforcement Learning},
  author={Tian, Rui and Gao, Mingfei and Gang, Haiming and Lu, Jiasen and Gan, Zhe and Yang, Yinfei and Wu, Zuxuan and Dehghan, Afshin},
  journal={arXiv preprint arXiv:2511.14760},
  year={2025}
}

@article{li2025onecat,
  title={Onecat: Decoder-only auto-regressive model for unified understanding and generation},
  author={Li, Han and Peng, Xinyu and Wang, Yaoming and Peng, Zelin and Chen, Xin and Weng, Rongxiang and Wang, Jingang and Cai, Xunliang and Dai, Wenrui and Xiong, Hongkai},
  journal={arXiv preprint arXiv:2509.03498},
  year={2025}
}

@article{he2025emma,
  title={Emma: Efficient multimodal understanding, generation, and editing with a unified architecture},
  author={He, Xin and Wei, Longhui and Ouyang, Jianbo and Liao, Minghui and Xie, Lingxi and Tian, Qi},
  journal={arXiv preprint arXiv:2512.04810},
  year={2025}
}

@article{liu2025tuna,
  title={Tuna: Taming unified visual representations for native unified multimodal models},
  author={Liu, Zhiheng and Ren, Weiming and Liu, Haozhe and Zhou, Zijian and Chen, Shoufa and Qiu, Haonan and Huang, Xiaoke and An, Zhaochong and Yang, Fanny and Patel, Aditya and others},
  journal={arXiv preprint arXiv:2512.02014},
  year={2025}
}

@article{wang2025ovis,
  title={Ovis-u1 technical report},
  author={Wang, Guo-Hua and Zhao, Shanshan and Zhang, Xinjie and Cao, Liangfu and Zhan, Pengxin and Duan, Lunhao and Lu, Shiyin and Fu, Minghao and Chen, Xiaohao and Zhao, Jianshan and others},
  journal={arXiv preprint arXiv:2506.23044},
  year={2025}
}

@article{pan2025transfer,
  title={Transfer between modalities with metaqueries},
  author={Pan, Xichen and Shukla, Satya Narayan and Singh, Aashu and Zhao, Zhuokai and Mishra, Shlok Kumar and Wang, Jialiang and Xu, Zhiyang and Chen, Jiuhai and Li, Kunpeng and Juefei-Xu, Felix and others},
  journal={arXiv preprint arXiv:2504.06256},
  year={2025}
}

@article{chen2025blip3,
  title={Blip3-o: A family of fully open unified multimodal models-architecture, training and dataset},
  author={Chen, Jiuhai and Xu, Zhiyang and Pan, Xichen and Hu, Yushi and Qin, Can and Goldstein, Tom and Huang, Lifu and Zhou, Tianyi and Xie, Saining and Savarese, Silvio and others},
  journal={arXiv preprint arXiv:2505.09568},
  year={2025}
}

@article{chen2025janus,
  title={Janus-pro: Unified multimodal understanding and generation with data and model scaling},
  author={Chen, Xiaokang and Wu, Zhiyu and Liu, Xingchao and Pan, Zizheng and Liu, Wen and Xie, Zhenda and Yu, Xingkai and Ruan, Chong},
  journal={arXiv preprint arXiv:2501.17811},
  year={2025}
}

@misc{openai2023gpt4v,
  title        = {GPT-4V(ision) System Card},
  author       = {{OpenAI}},
  year         = {2023},
  month        = sep,
  howpublished = {\url{https://openai.com/index/gpt-4v-system-card/}},
  note         = {Accessed: 2026-05-15}
}

@inproceedings{wu2025janus,
  title={Janus: Decoupling visual encoding for unified multimodal understanding and generation},
  author={Wu, Chengyue and Chen, Xiaokang and Wu, Zhiyu and Ma, Yiyang and Liu, Xingchao and Pan, Zizheng and Liu, Wen and Xie, Zhenda and Yu, Xingkai and Ruan, Chong and others},
  booktitle={Proceedings of the Computer Vision and Pattern Recognition Conference},
  pages={12966--12977},
  year={2025}
}

@article{xie2024show,
  title={Show-o: One single transformer to unify multimodal understanding and generation},
  author={Xie, Jinheng and Mao, Weijia and Bai, Zechen and Zhang, David Junhao and Wang, Weihao and Lin, Kevin Qinghong and Gu, Yuchao and Chen, Zhijie and Yang, Zhenheng and Shou, Mike Zheng},
  journal={arXiv preprint arXiv:2408.12528},
  year={2024}
}

@article{liao2025mogao,
  title={Mogao: An omni foundation model for interleaved multi-modal generation},
  author={Liao, Chao and Liu, Liyang and Wang, Xun and Luo, Zhengxiong and Zhang, Xinyu and Zhao, Wenliang and Wu, Jie and Li, Liang and Tian, Zhi and Huang, Weilin},
  journal={arXiv preprint arXiv:2505.05472},
  year={2025}
}

@article{ju2025fulldit,
  title={Fulldit: Multi-task video generative foundation model with full attention},
  author={Ju, Xuan and Ye, Weicai and Liu, Quande and Wang, Qiulin and Wang, Xintao and Wan, Pengfei and Zhang, Di and Gai, Kun and Xu, Qiang},
  journal={arXiv preprint arXiv:2503.19907},
  year={2025}
}

@article{ju2025editverse,
  title={Editverse: Unifying image and video editing and generation with in-context learning},
  author={Ju, Xuan and Wang, Tianyu and Zhou, Yuqian and Zhang, He and Liu, Qing and Zhao, Nanxuan and Zhang, Zhifei and Li, Yijun and Cai, Yuanhao and Liu, Shaoteng and others},
  journal={arXiv preprint arXiv:2509.20360},
  year={2025}
}

@article{ye2025unic,
  title={Unic: Unified in-context video editing},
  author={Ye, Zixuan and He, Xuanhua and Liu, Quande and Wang, Qiulin and Wang, Xintao and Wan, Pengfei and Zhang, Di and Gai, Kun and Chen, Qifeng and Luo, Wenhan},
  journal={arXiv preprint arXiv:2506.04216},
  year={2025}
}

@inproceedings{jiang2025vace,
  title={Vace: All-in-one video creation and editing},
  author={Jiang, Zeyinzi and Han, Zhen and Mao, Chaojie and Zhang, Jingfeng and Pan, Yulin and Liu, Yu},
  booktitle={Proceedings of the IEEE/CVF International Conference on Computer Vision},
  pages={17191--17202},
  year={2025}
}

@article{ku2024anyv2v,
  title={Anyv2v: A tuning-free framework for any video-to-video editing tasks},
  author={Ku, Max and Wei, Cong and Ren, Weiming and Yang, Harry and Chen, Wenhu},
  journal={arXiv preprint arXiv:2403.14468},
  year={2024}
}

@inproceedings{wu2024next,
  title={Next-gpt: Any-to-any multimodal llm},
  author={Wu, Shengqiong and Fei, Hao and Qu, Leigang and Ji, Wei and Chua, Tat-Seng},
  booktitle={Forty-first International Conference on Machine Learning},
  year={2024}
}

@misc{openai2025gptimage1,
    title = {Introducing 4o Image Generation},
    author = {{OpenAI}},
    year = {2025},
    howpublished = {\url{https://openai.com/index/introducing-4o-image-generation/}},
    note = {Accessed: 2026-04-10}
}

@article{xin2025lumina,
  title={Lumina-dimoo: An omni diffusion large language model for multi-modal generation and understanding},
  author={Xin, Yi and Qin, Qi and Luo, Siqi and Zhu, Kaiwen and Yan, Juncheng and Tai, Yan and Lei, Jiayi and Cao, Yuewen and Wang, Keqi and Wang, Yibin and others},
  journal={arXiv preprint arXiv:2510.06308},
  year={2025}
}

@inproceedings{wang2024gpt4video,
  title={Gpt4video: A unified multimodal large language model for Instruction-followed understanding and safety-aware generation},
  author={Wang, Zhanyu and Wang, Longyue and Zhao, Zhen and Wu, Minghao and Lyu, Chenyang and Li, Huayang and Cai, Deng and Zhou, Luping and Shi, Shuming and Tu, Zhaopeng},
  booktitle={Proceedings of the 32nd ACM International Conference on Multimedia},
  pages={3907--3916},
  year={2024}
}

@article{han2025tv2tv,
  title={TV2TV: A Unified Framework for Interleaved Language and Video Generation},
  author={Han, Xiaochuang and Emad, Youssef and Hall, Melissa and Nguyen, John and Padthe, Karthik and Robbins, Liam and Bar, Amir and Chen, Delong and Drozdzal, Michal and Elbayad, Maha and others},
  journal={arXiv preprint arXiv:2512.05103},
  year={2025}
}

@article{wei2025univideo,
  title={Univideo: Unified understanding, generation, and editing for videos},
  author={Wei, Cong and Liu, Quande and Ye, Zixuan and Wang, Qiulin and Wang, Xintao and Wan, Pengfei and Gai, Kun and Chen, Wenhu},
  journal={arXiv preprint arXiv:2510.08377},
  year={2025}
}

@article{tan2025omni,
  title={Omni-video: Democratizing unified video understanding and generation},
  author={Tan, Zhiyu and Yang, Hao and Qin, Luozheng and Gong, Jia and Yang, Mengping and Li, Hao},
  journal={arXiv preprint arXiv:2507.06119},
  year={2025}
}

@article{xiao2025omnibridge,
  title={OmniBridge: Unified Multimodal Understanding, Generation, and Retrieval via Latent Space Alignment},
  author={Xiao, Teng and Li, Zuchao and Zhang, Lefei},
  journal={arXiv preprint arXiv:2509.19018},
  year={2025}
}

@article{zhao2025unified,
  title={Unified multimodal understanding and generation models: Advances, challenges, and opportunities},
  author={Zhao, Shanshan and Zhang, Xinjie and Guo, Jintao and Hu, Jiakui and Duan, Lunhao and Fu, Minghao and Chng, Yong Xien and Wang, Guo-Hua and Chen, Qing-Guo and Xu, Zhao and others},
  journal={arXiv preprint arXiv:2505.02567},
  year={2025}
}

@inproceedings{ma2025janusflow,
  title={Janusflow: Harmonizing autoregression and rectified flow for unified multimodal understanding and generation},
  author={Ma, Yiyang and Liu, Xingchao and Chen, Xiaokang and Liu, Wen and Wu, Chengyue and Wu, Zhiyu and Pan, Zizheng and Xie, Zhenda and Zhang, Haowei and Yu, Xingkai and others},
  booktitle={Proceedings of the IEEE/CVF Conference on Computer Vision and Pattern Recognition},
  pages={7739--7751},
  year={2025}
}

@article{tian2026internvlu,
      title={InternVL-U: Democratizing Unified Multimodal Models for Understanding, Reasoning, Generation and Editing},
      author={Tian, Changyao and Yang, Danni and Chen, Guanzhou and Cui, Erfei and Wang, Zhaokai and Duan, Yuchen and Yin, Penghao and Chen, Sitao and Yang, Ganlin and Liu, Mingxin and Zhu, Zirun and Fan, Ziqian and Gu, Leyao and Wang, Haomin and Wei, Qi and Yin, Jinhui and Yang, Xue and Zhong, Zhihang and Qin, Qi and Xin, Yi and Fu, Bin and Liu, Yihao and Ge, Jiaye and Guo, Qipeng and Luo, Gen and Li, Hongsheng and Qiao, Yu and Chen, Kai and Zhang, Hongjie},
      year={2026},
      eprint={2603.09877},
      archivePrefix={arXiv},
      primaryClass={cs.CV},
      url={https://arxiv.org/abs/2603.09877}
}

@article{zhou2024transfusion,
  title={Transfusion: Predict the next token and diffuse images with one multi-modal model},
  author={Zhou, Chunting and Yu, Lili and Babu, Arun and Tirumala, Kushal and Yasunaga, Michihiro and Shamis, Leonid and Kahn, Jacob and Ma, Xuezhe and Zettlemoyer, Luke and Levy, Omer},
  journal={arXiv preprint arXiv:2408.11039},
  year={2024}
}

@article{zheng2025diffusion,
  title={Diffusion transformers with representation autoencoders},
  author={Zheng, Boyang and Ma, Nanye and Tong, Shengbang and Xie, Saining},
  journal={arXiv preprint arXiv:2510.11690},
  year={2025}
}

@article{yu2024representation,
  title={Representation alignment for generation: Training diffusion transformers is easier than you think},
  author={Yu, Sihyun and Kwak, Sangkyung and Jang, Huiwon and Jeong, Jongheon and Huang, Jonathan and Shin, Jinwoo and Xie, Saining},
  journal={arXiv preprint arXiv:2410.06940},
  year={2024}
}

@article{wu2025less,
  title={Less-to-more generalization: Unlocking more controllability by in-context generation},
  author={Wu, Shaojin and Huang, Mengqi and Wu, Wenxu and Cheng, Yufeng and Ding, Fei and He, Qian},
  journal={arXiv preprint arXiv:2504.02160},
  year={2025}
}

@article{wu2024vmix,
  title={VMix: Improving Text-to-Image Diffusion Model with Cross-Attention Mixing Control},
  author={Wu, Shaojin and Ding, Fei and Huang, Mengqi and Liu, Wei and He, Qian},
  journal={arXiv preprint arXiv:2412.20800},
  year={2024}
}

@inproceedings{dehghani2023scaling,
  title={Scaling vision transformers to 22 billion parameters},
  author={Dehghani, Mostafa and Djolonga, Josip and Mustafa, Basil and Padlewski, Piotr and Heek, Jonathan and Gilmer, Justin and Steiner, Andreas Peter and Caron, Mathilde and Geirhos, Robert and Alabdulmohsin, Ibrahim and others},
  booktitle={International conference on machine learning},
  pages={7480--7512},
  year={2023},
  organization={PMLR}
}

@article{wu2025omnigen2,
  title={OmniGen2: Exploration to Advanced Multimodal Generation},
  author={Wu, Chenyuan and Zheng, Pengfei and Yan, Ruiran and Xiao, Shitao and Luo, Xin and Wang, Yueze and Li, Wanli and Jiang, Xiyan and Liu, Yexin and Zhou, Junjie and others},
  journal={arXiv preprint arXiv:2506.18871},
  year={2025}
}

@article{mou2025dreamo,
  title={Dreamo: A unified framework for image customization},
  author={Mou, Chong and Wu, Yanze and Wu, Wenxu and Guo, Zinan and Zhang, Pengze and Cheng, Yufeng and Luo, Yiming and Ding, Fei and Zhang, Shiwen and Li, Xinghui and others},
  journal={arXiv preprint arXiv:2504.16915},
  year={2025}
}

@article{lin2025uniworld,
  title={Uniworld: High-resolution semantic encoders for unified visual understanding and generation},
  author={Lin, Bin and Li, Zongjian and Cheng, Xinhua and Niu, Yuwei and Ye, Yang and He, Xianyi and Yuan, Shenghai and Yu, Wangbo and Wang, Shaodong and Ge, Yunyang and others},
  journal={arXiv preprint arXiv:2506.03147},
  year={2025}
}

@article{labs2025flux,
  title={FLUX. 1 Kontext: Flow Matching for In-Context Image Generation and Editing in Latent Space},
  author={Labs, Black Forest and Batifol, Stephen and Blattmann, Andreas and Boesel, Frederic and Consul, Saksham and Diagne, Cyril and Dockhorn, Tim and English, Jack and English, Zion and Esser, Patrick and others},
  journal={arXiv preprint arXiv:2506.15742},
  year={2025}
}

@article{wu2025qwen,
  title={Qwen-image technical report},
  author={Wu, Chenfei and Li, Jiahao and Zhou, Jingren and Lin, Junyang and Gao, Kaiyuan and Yan, Kun and Yin, Sheng-ming and Bai, Shuai and Xu, Xiao and Chen, Yilei and others},
  journal={arXiv preprint arXiv:2508.02324},
  year={2025}
}

@article{Qwen2-VL,
  title={Qwen2-VL: Enhancing Vision-Language Model's Perception of the World at Any Resolution},
  author={Wang, Peng and Bai, Shuai and Tan, Sinan and Wang, Shijie and Fan, Zhihao and Bai, Jinze and Chen, Keqin and Liu, Xuejing and Wang, Jialin and Ge, Wenbin and Fan, Yang and Dang, Kai and Du, Mengfei and Ren, Xuancheng and Men, Rui and Liu, Dayiheng and Zhou, Chang and Zhou, Jingren and Lin, Junyang},
  journal={arXiv preprint arXiv:2409.12191},
  year={2024}
}

@article{Qwen-VL,
  title={Qwen-VL: A Versatile Vision-Language Model for Understanding, Localization, Text Reading, and Beyond},
  author={Bai, Jinze and Bai, Shuai and Yang, Shusheng and Wang, Shijie and Tan, Sinan and Wang, Peng and Lin, Junyang and Zhou, Chang and Zhou, Jingren},
  journal={arXiv preprint arXiv:2308.12966},
  year={2023}
}

@article{Qwen3-VL,
      title={Qwen3-VL Technical Report}, 
      author={Shuai Bai and Yuxuan Cai and Ruizhe Chen and Keqin Chen and Xionghui Chen and Zesen Cheng and Lianghao Deng and Wei Ding and Chang Gao and Chunjiang Ge and Wenbin Ge and Zhifang Guo and Qidong Huang and Jie Huang and Fei Huang and Binyuan Hui and Shutong Jiang and Zhaohai Li and Mingsheng Li and Mei Li and Kaixin Li and Zicheng Lin and Junyang Lin and Xuejing Liu and Jiawei Liu and Chenglong Liu and Yang Liu and Dayiheng Liu and Shixuan Liu and Dunjie Lu and Ruilin Luo and Chenxu Lv and Rui Men and Lingchen Meng and Xuancheng Ren and Xingzhang Ren and Sibo Song and Yuchong Sun and Jun Tang and Jianhong Tu and Jianqiang Wan and Peng Wang and Pengfei Wang and Qiuyue Wang and Yuxuan Wang and Tianbao Xie and Yiheng Xu and Haiyang Xu and Jin Xu and Zhibo Yang and Mingkun Yang and Jianxin Yang and An Yang and Bowen Yu and Fei Zhang and Hang Zhang and Xi Zhang and Bo Zheng and Humen Zhong and Jingren Zhou and Fan Zhou and Jing Zhou and Yuanzhi Zhu and Ke Zhu},
	  journal={arXiv preprint arXiv:2511.21631},
      year={2025}
}

@inproceedings{chen2024internvl,
  title={Internvl: Scaling up vision foundation models and aligning for generic visual-linguistic tasks},
  author={Chen, Zhe and Wu, Jiannan and Wang, Wenhai and Su, Weijie and Chen, Guo and Xing, Sen and Zhong, Muyan and Zhang, Qinglong and Zhu, Xizhou and Lu, Lewei and others},
  booktitle={Proceedings of the IEEE/CVF Conference on Computer Vision and Pattern Recognition},
  pages={24185--24198},
  year={2024}
}

@inproceedings{huang2024vbench,
  title={Vbench: Comprehensive benchmark suite for video generative models},
  author={Huang, Ziqi and He, Yinan and Yu, Jiashuo and Zhang, Fan and Si, Chenyang and Jiang, Yuming and Zhang, Yuanhan and Wu, Tianxing and Jin, Qingyang and Chanpaisit, Nattapol and others},
  booktitle={Proceedings of the IEEE/CVF Conference on Computer Vision and Pattern Recognition},
  pages={21807--21818},
  year={2024}
}

@inproceedings{li2024mvbench,
  title={Mvbench: A comprehensive multi-modal video understanding benchmark},
  author={Li, Kunchang and Wang, Yali and He, Yinan and Li, Yizhuo and Wang, Yi and Liu, Yi and Wang, Zun and Xu, Jilan and Chen, Guo and Luo, Ping and others},
  booktitle={Proceedings of the IEEE/CVF Conference on Computer Vision and Pattern Recognition},
  pages={22195--22206},
  year={2024}
}

@article{hu2024ella,
  title={Ella: Equip diffusion models with llm for enhanced semantic alignment},
  author={Hu, Xiwei and Wang, Rui and Fang, Yixiao and Fu, Bin and Cheng, Pei and Yu, Gang},
  journal={arXiv preprint arXiv:2403.05135},
  year={2024}
}

@article{ho2022classifier,
  title={Classifier-free diffusion guidance},
  author={Ho, Jonathan and Salimans, Tim},
  journal={arXiv preprint arXiv:2207.12598},
  year={2022}
}

@article{ghosh2023geneval,
  title={Geneval: An object-focused framework for evaluating text-to-image alignment},
  author={Ghosh, Dhruba and Hajishirzi, Hannaneh and Schmidt, Ludwig},
  journal={Advances in Neural Information Processing Systems},
  volume={36},
  pages={52132--52152},
  year={2023}
}

@article{gao2024mini,
  title={Mini-InternVL: a flexible-transfer pocket multi-modal model with 5\% parameters and 90\% performance},
  author={Gao, Zhangwei and Chen, Zhe and Cui, Erfei and Ren, Yiming and Wang, Weiyun and Zhu, Jinguo and Tian, Hao and Ye, Shenglong and He, Junjun and Zhu, Xizhou and others},
  journal={Visual Intelligence},
  volume={2},
  number={1},
  pages={1--17},
  year={2024},
  publisher={Springer}
}

@article{chen2024far,
  title={How far are we to gpt-4v? closing the gap to commercial multimodal models with open-source suites},
  author={Chen, Zhe and Wang, Weiyun and Tian, Hao and Ye, Shenglong and Gao, Zhangwei and Cui, Erfei and Tong, Wenwen and Hu, Kongzhi and Luo, Jiapeng and Ma, Zheng and others},
  journal={Science China Information Sciences},
  volume={67},
  number={12},
  pages={220101},
  year={2024},
  publisher={Springer}
}

@article{team2023gemini,
  title={Gemini: a family of highly capable multimodal models},
  author={Team, Gemini and Anil, Rohan and Borgeaud, Sebastian and Alayrac, Jean-Baptiste and Yu, Jiahui and Soricut, Radu and Schalkwyk, Johan and Dai, Andrew M and Hauth, Anja and Millican, Katie and others},
  journal={arXiv preprint arXiv:2312.11805},
  year={2023}
}

@article{liu2023visual,
  title={Visual instruction tuning},
  author={Liu, Haotian and Li, Chunyuan and Wu, Qingyang and Lee, Yong Jae},
  journal={Advances in neural information processing systems},
  volume={36},
  pages={34892--34916},
  year={2023}
}

@inproceedings{radford2021learning,
  title={Learning transferable visual models from natural language supervision},
  author={Radford, Alec and Kim, Jong Wook and Hallacy, Chris and Ramesh, Aditya and Goh, Gabriel and Agarwal, Sandhini and Sastry, Girish and Askell, Amanda and Mishkin, Pamela and Clark, Jack and others},
  booktitle={International conference on machine learning},
  pages={8748--8763},
  year={2021},
  organization={PmLR}
}

@article{team2024gemini,
  title={Gemini 1.5: Unlocking multimodal understanding across millions of tokens of context},
  author={Team, Gemini and Georgiev, Petko and Lei, Ving Ian and Burnell, Ryan and Bai, Libin and Gulati, Anmol and Tanzer, Garrett and Vincent, Damien and Pan, Zhufeng and Wang, Shibo and others},
  journal={arXiv preprint arXiv:2403.05530},
  year={2024}
}

@article{achiam2023gpt,
  title={Gpt-4 technical report},
  author={Achiam, Josh and Adler, Steven and Agarwal, Sandhini and Ahmad, Lama and Akkaya, Ilge and Aleman, Florencia Leoni and Almeida, Diogo and Altenschmidt, Janko and Altman, Sam and Anadkat, Shyamal and others},
  journal={arXiv preprint arXiv:2303.08774},
  year={2023}
}

@article{wang2025internvl3_5,
  title={InternVL3.5: Advancing Open-Source Multimodal Models in Versatility, Reasoning, and Efficiency},
  author={Wang, Weiyun and Gao, Zhangwei and Gu, Lixin and Pu, Hengjun and Cui, Long and Wei, Xingguang and Liu, Zhaoyang and Jing, Linglin and Ye, Shenglong and Shao, Jie and others},
  journal={arXiv preprint arXiv:2508.18265},
  year={2025}
}

@article{cui2025emu3,
  title={Emu3. 5: Native multimodal models are world learners},
  author={Cui, Yufeng and Chen, Honghao and Deng, Haoge and Huang, Xu and Li, Xinghang and Liu, Jirong and Liu, Yang and Luo, Zhuoyan and Wang, Jinsheng and Wang, Wenxuan and others},
  journal={arXiv preprint arXiv:2510.26583},
  year={2025}
}

@article{deng2025emerging,
  title={Emerging properties in unified multimodal pretraining},
  author={Deng, Chaorui and Zhu, Deyao and Li, Kunchang and Gou, Chenhui and Li, Feng and Wang, Zeyu and Zhong, Shu and Yu, Weihao and Nie, Xiaonan and Song, Ziang and others},
  journal={arXiv preprint arXiv:2505.14683},
  year={2025}
}

@article{yang2024vision,
  title={Vision model pre-training on interleaved image-text data via latent compression learning},
  author={Yang, Chenyu and Zhu, Xizhou and Zhu, Jinguo and Su, Weijie and Wang, Junjie and Dong, Xuan and Wang, Wenhai and Li, Bin and Zhou, Jie and Qiao, Yu and others},
  journal={Advances in Neural Information Processing Systems},
  volume={37},
  pages={23912--23938},
  year={2024}
}

@inproceedings{zhang-etal-2023-video,
    title = "Video-{LL}a{MA}: An Instruction-tuned Audio-Visual Language Model for Video Understanding",
    author = "Zhang, Hang  and
      Li, Xin  and
      Bing, Lidong",
    editor = "Feng, Yansong  and
      Lefever, Els",
    booktitle = "Proceedings of the 2023 Conference on Empirical Methods in Natural Language Processing: System Demonstrations",
    month = dec,
    year = "2023",
    address = "Singapore",
    publisher = "Association for Computational Linguistics",
    url = "https://aclanthology.org/2023.emnlp-demo.49/",
    doi = "10.18653/v1/2023.emnlp-demo.49",
    pages = "543--553"
}

@article{zhang2023llamaadapter,
  title = {LLaMA-Adapter: Efficient Fine-tuning of Language Models with Zero-init Attention},
  author={Zhang, Renrui and Han, Jiaming and Liu, Chris and Gao, Peng and Zhou, Aojun and Hu, Xiangfei and Yan, Shilin and Lu, Pan and Li, Hongsheng and Qiao, Yu},
  journal={arXiv preprint arXiv:2303.16199},
  year={2023}
}

@inproceedings{Maaz2023VideoChatGPT,
    title={Video-ChatGPT: Towards Detailed Video Understanding via Large Vision and Language Models},
    author={Maaz, Muhammad and Rasheed, Hanoona and Khan, Salman and Khan, Fahad Shahbaz},
    booktitle={Proceedings of the 62nd Annual Meeting of the Association for Computational Linguistics (ACL 2024)},
    year={2024}
}

@inproceedings{liu2024st,
  title={St-llm: Large language models are effective temporal learners},
  author={Liu, Ruyang and Li, Chen and Tang, Haoran and Ge, Yixiao and Shan, Ying and Li, Ge},
  booktitle={European Conference on Computer Vision},
  pages={1--18},
  year={2024},
  organization={Springer}
}

@misc{xu2024pllava,
      title={PLLaVA : Parameter-free LLaVA Extension from Images to Videos for Video Dense Captioning}, 
      author={Lin Xu and Yilin Zhao and Daquan Zhou and Zhijie Lin and See Kiong Ng and Jiashi Feng},
      year={2024},
      eprint={2404.16994},
      archivePrefix={arXiv},
      primaryClass={cs.CV}
}

@article{touvron2023llama,
  title={Llama: Open and efficient foundation language models},
  author={Touvron, Hugo and Lavril, Thibaut and Izacard, Gautier and Martinet, Xavier and Lachaux, Marie-Anne and Lacroix, Timoth{\'e}e and Rozi{\`e}re, Baptiste and Goyal, Naman and Hambro, Eric and Azhar, Faisal and others},
  journal={arXiv preprint arXiv:2302.13971},
  year={2023}
}

@article{liu2024deepseek,
  title={Deepseek-v3 technical report},
  author={Liu, Aixin and Feng, Bei and Xue, Bing and Wang, Bingxuan and Wu, Bochao and Lu, Chengda and Zhao, Chenggang and Deng, Chengqi and Zhang, Chenyu and Ruan, Chong and others},
  journal={arXiv preprint arXiv:2412.19437},
  year={2024}
}

@article{li2025videochat,
  title={Videochat: Chat-centric video understanding},
  author={Li, KunChang and He, Yinan and Wang, Yi and Li, Yizhuo and Wang, Wenhai and Luo, Ping and Wang, Yali and Wang, Limin and Qiao, Yu},
  journal={Science China Information Sciences},
  volume={68},
  number={10},
  pages={200102},
  year={2025},
  publisher={Springer}
}

@inproceedings{qu2025tokenflow,
  title={Tokenflow: Unified image tokenizer for multimodal understanding and generation},
  author={Qu, Liao and Zhang, Huichao and Liu, Yiheng and Wang, Xu and Jiang, Yi and Gao, Yiming and Ye, Hu and Du, Daniel K and Yuan, Zehuan and Wu, Xinglong},
  booktitle={Proceedings of the Computer Vision and Pattern Recognition Conference},
  pages={2545--2555},
  year={2025}
}

@article{wang2024emu3,
  title={Emu3: Next-token prediction is all you need},
  author={Wang, Xinlong and Zhang, Xiaosong and Luo, Zhengxiong and Sun, Quan and Cui, Yufeng and Wang, Jinsheng and Zhang, Fan and Wang, Yueze and Li, Zhen and Yu, Qiying and others},
  journal={arXiv preprint arXiv:2409.18869},
  year={2024}
}

@article{tschannen2025siglip,
  title={Siglip 2: Multilingual vision-language encoders with improved semantic understanding, localization, and dense features},
  author={Tschannen, Michael and Gritsenko, Alexey and Wang, Xiao and Naeem, Muhammad Ferjad and Alabdulmohsin, Ibrahim and Parthasarathy, Nikhil and Evans, Talfan and Beyer, Lucas and Xia, Ye and Mustafa, Basil and others},
  journal={arXiv preprint arXiv:2502.14786},
  year={2025}
}

@article{liu2024world,
  title={World model on million-length video and language with blockwise ringattention},
  author={Liu, Hao and Yan, Wilson and Zaharia, Matei and Abbeel, Pieter},
  journal={arXiv preprint arXiv:2402.08268},
  year={2024}
}

@article{ge2024seed,
  title={Seed-x: Multimodal models with unified multi-granularity comprehension and generation},
  author={Ge, Yuying and Zhao, Sijie and Zhu, Jinguo and Ge, Yixiao and Yi, Kun and Song, Lin and Li, Chen and Ding, Xiaohan and Shan, Ying},
  journal={arXiv preprint arXiv:2404.14396},
  year={2024}
}

@article{team2024chameleon,
  title={Chameleon: Mixed-modal early-fusion foundation models},
  author={Team, Chameleon},
  journal={arXiv preprint arXiv:2405.09818},
  year={2024}
}

@article{wang2023modelscope,
  title={Modelscope text-to-video technical report},
  author={Wang, Jiuniu and Yuan, Hangjie and Chen, Dayou and Zhang, Yingya and Wang, Xiang and Zhang, Shiwei},
  journal={arXiv preprint arXiv:2308.06571},
  year={2023}
}

@article{wang2025lavie,
  title={Lavie: High-quality video generation with cascaded latent diffusion models},
  author={Wang, Yaohui and Chen, Xinyuan and Ma, Xin and Zhou, Shangchen and Huang, Ziqi and Wang, Yi and Yang, Ceyuan and He, Yinan and Yu, Jiashuo and Yang, Peiqing and others},
  journal={International Journal of Computer Vision},
  volume={133},
  number={5},
  pages={3059--3078},
  year={2025},
  publisher={Springer}
}

@inproceedings{chen2024videocrafter2,
  title={Videocrafter2: Overcoming data limitations for high-quality video diffusion models},
  author={Chen, Haoxin and Zhang, Yong and Cun, Xiaodong and Xia, Menghan and Wang, Xintao and Weng, Chao and Shan, Ying},
  booktitle={Proceedings of the IEEE/CVF conference on computer vision and pattern recognition},
  pages={7310--7320},
  year={2024}
}

@article{zhang2025show,
  title={Show-1: Marrying pixel and latent diffusion models for text-to-video generation},
  author={Zhang, David Junhao and Wu, Jay Zhangjie and Liu, Jia-Wei and Zhao, Rui and Ran, Lingmin and Gu, Yuchao and Gao, Difei and Shou, Mike Zheng},
  journal={International Journal of Computer Vision},
  volume={133},
  number={4},
  pages={1879--1893},
  year={2025},
  publisher={Springer}
}

@article{guo2023animatediff,
  title={Animatediff: Animate your personalized text-to-image diffusion models without specific tuning},
  author={Guo, Yuwei and Yang, Ceyuan and Rao, Anyi and Liang, Zhengyang and Wang, Yaohui and Qiao, Yu and Agrawala, Maneesh and Lin, Dahua and Dai, Bo},
  journal={arXiv preprint arXiv:2307.04725},
  year={2023}
}

@article{fei2024video,
  title={Video-ccam: Enhancing video-language understanding with causal cross-attention masks for short and long videos},
  author={Fei, Jiajun and Li, Dian and Deng, Zhidong and Wang, Zekun and Liu, Gang and Wang, Hui},
  journal={arXiv preprint arXiv:2408.14023},
  year={2024}
}

@article{tuna2,
  title={Tuna-2: Pixel Embeddings Beat Vision Encoders
         for Multimodal Understanding and Generation},
  author={Liu, Zhiheng and Ren, Weiming and Huang, Xiaoke
          and Chen, Shoufa and Li, Tianhong and Chen, Mengzhao
          and Ji, Yatai and He, Sen and Schult, Jonas
          and Zeng, Belinda and Xiang, Tao and Chen, Wenhu
          and Luo, Ping and Zettlemoyer, Luke and Cong, Yuren},
  journal={arXiv preprint arXiv:2604.24763},
  year={2026}
}

@article{chen2024timemarker,
  title={TimeMarker: A Versatile Video-LLM for Long and Short Video Understanding with Superior Temporal Localization Ability},
  author={Shimin Chen and Xiaohan Lan and Yitian Yuan and Zequn Jie and Lin Ma},
  journal={arXiv preprint arXiv:2411.18211},
  year={2024}
}

@inproceedings{wang2024internvideo2,
  title={Internvideo2: Scaling foundation models for multimodal video understanding},
  author={Wang, Yi and Li, Kunchang and Li, Xinhao and Yu, Jiashuo and He, Yinan and Chen, Guo and Pei, Baoqi and Zheng, Rongkun and Wang, Zun and Shi, Yansong and others},
  booktitle={European conference on computer vision},
  pages={396--416},
  year={2024},
  organization={Springer}
}

@article{Qwen2.5-VL,
  title={Qwen2.5-VL Technical Report},
  author={Bai, Shuai and Chen, Keqin and Liu, Xuejing and Wang, Jialin and Ge, Wenbin and Song, Sibo and Dang, Kai and Wang, Peng and Wang, Shijie and Tang, Jun and Zhong, Humen and Zhu, Yuanzhi and Yang, Mingkun and Li, Zhaohai and Wan, Jianqiang and Wang, Pengfei and Ding, Wei and Fu, Zheren and Xu, Yiheng and Ye, Jiabo and Zhang, Xi and Xie, Tianbao and Cheng, Zesen and Zhang, Hang and Yang, Zhibo and Xu, Haiyang and Lin, Junyang},
  journal={arXiv preprint arXiv:2502.13923},
  year={2025}
}

@article{fan2025unified,
  title={Unified autoregressive visual generation and understanding with continuous tokens},
  author={Fan, Lijie and Tang, Luming and Qin, Siyang and Li, Tianhong and Yang, Xuan and Qiao, Siyuan and Steiner, Andreas and Sun, Chen and Li, Yuanzhen and Zhu, Tao and others},
  journal={arXiv preprint arXiv:2503.13436},
  year={2025}
}

@article{li2024autoregressive,
  title={Autoregressive image generation without vector quantization},
  author={Li, Tianhong and Tian, Yonglong and Li, He and Deng, Mingyang and He, Kaiming},
  journal={Advances in Neural Information Processing Systems},
  volume={37},
  pages={56424--56445},
  year={2024}
}

@article{liu2024mardini,
  title={Mardini: Masked autoregressive diffusion for video generation at scale},
  author={Liu, Haozhe and Liu, Shikun and Zhou, Zijian and Xu, Mengmeng and Xie, Yanping and Han, Xiao and P{\'e}rez, Juan C and Liu, Ding and Kahatapitiya, Kumara and Jia, Menglin and others},
  journal={arXiv preprint arXiv:2410.20280},
  year={2024}
}

@article{wan2025wan,
      title={Wan: Open and Advanced Large-Scale Video Generative Models}, 
      author={Team Wan and Ang Wang and Baole Ai and Bin Wen and Chaojie Mao and Chen-Wei Xie and Di Chen and Feiwu Yu and Haiming Zhao and Jianxiao Yang and Jianyuan Zeng and Jiayu Wang and Jingfeng Zhang and Jingren Zhou and Jinkai Wang and Jixuan Chen and Kai Zhu and Kang Zhao and Keyu Yan and Lianghua Huang and Mengyang Feng and Ningyi Zhang and Pandeng Li and Pingyu Wu and Ruihang Chu and Ruili Feng and Shiwei Zhang and Siyang Sun and Tao Fang and Tianxing Wang and Tianyi Gui and Tingyu Weng and Tong Shen and Wei Lin and Wei Wang and Wei Wang and Wenmeng Zhou and Wente Wang and Wenting Shen and Wenyuan Yu and Xianzhong Shi and Xiaoming Huang and Xin Xu and Yan Kou and Yangyu Lv and Yifei Li and Yijing Liu and Yiming Wang and Yingya Zhang and Yitong Huang and Yong Li and You Wu and Yu Liu and Yulin Pan and Yun Zheng and Yuntao Hong and Yupeng Shi and Yutong Feng and Zeyinzi Jiang and Zhen Han and Zhi-Fan Wu and Ziyu Liu},
      journal = {arXiv preprint arXiv:2503.20314},
      year={2025}
}

@inproceedings{lin2024video,
  title={Video-llava: Learning united visual representation by alignment before projection},
  author={Lin, Bin and Ye, Yang and Zhu, Bin and Cui, Jiaxi and Ning, Munan and Jin, Peng and Yuan, Li},
  booktitle={Proceedings of the 2024 conference on empirical methods in natural language processing},
  pages={5971--5984},
  year={2024}
}

@article{yang2025cambrian,
  title={Cambrian-s: Towards spatial supersensing in video},
  author={Yang, Shusheng and Yang, Jihan and Huang, Pinzhi and Brown, Ellis and Yang, Zihao and Yu, Yue and Tong, Shengbang and Zheng, Zihan and Xu, Yifan and Wang, Muhan and others},
  journal={arXiv preprint arXiv:2511.04670},
  year={2025}
}

@article{li2024llava,
  title={Llava-onevision: Easy visual task transfer},
  author={Li, Bo and Zhang, Yuanhan and Guo, Dong and Zhang, Renrui and Li, Feng and Zhang, Hao and Zhang, Kaichen and Zhang, Peiyuan and Li, Yanwei and Liu, Ziwei and others},
  journal={arXiv preprint arXiv:2408.03326},
  year={2024}
}

@misc{liu2024llavanext,
  title={Llavanext: Improved reasoning, ocr, and world knowledge},
  author={Liu, Haotian and Li, Chunyuan and Li, Yuheng and Li, Bo and Zhang, Yuanhan and Shen, Sheng and Lee, Yong Jae},
  year={2024}
}

@inproceedings{liu2024improved,
  title={Improved baselines with visual instruction tuning},
  author={Liu, Haotian and Li, Chunyuan and Li, Yuheng and Lee, Yong Jae},
  booktitle={Proceedings of the IEEE/CVF conference on computer vision and pattern recognition},
  pages={26296--26306},
  year={2024}
}

@article{dai2023instructblip,
  title={Instructblip: Towards general-purpose vision-language models with instruction tuning},
  author={Dai, Wenliang and Li, Junnan and Li, Dongxu and Tiong, Anthony and Zhao, Junqi and Wang, Weisheng and Li, Boyang and Fung, Pascale N and Hoi, Steven},
  journal={Advances in neural information processing systems},
  volume={36},
  pages={49250--49267},
  year={2023}
}

@article{laurenccon2023obelics,
  title={Obelics: An open web-scale filtered dataset of interleaved image-text documents},
  author={Lauren{\c{c}}on, Hugo and Saulnier, Lucile and Tronchon, L{\'e}o and Bekman, Stas and Singh, Amanpreet and Lozhkov, Anton and Wang, Thomas and Karamcheti, Siddharth and Rush, Alexander and Kiela, Douwe and others},
  journal={Advances in Neural Information Processing Systems},
  volume={36},
  pages={71683--71702},
  year={2023}
}

@article{alayrac2022flamingo,
  title={Flamingo: a visual language model for few-shot learning},
  author={Alayrac, Jean-Baptiste and Donahue, Jeff and Luc, Pauline and Miech, Antoine and Barr, Iain and Hasson, Yana and Lenc, Karel and Mensch, Arthur and Millican, Katherine and Reynolds, Malcolm and others},
  journal={Advances in neural information processing systems},
  volume={35},
  pages={23716--23736},
  year={2022}
}

@article{dai2026chatumm,
  title={ChatUMM: Robust Context Tracking for Conversational Interleaved Generation},
  author={Dai, Wenxun and Zhao, Zhiyuan and Zhong, Yule and Cheng, Yiji and Zhang, Jianwei and Wang, Linqing and Zhang, Shiyi and Lin, Yunlong and He, Runze and Song, Fellix and others},
  journal={arXiv preprint arXiv:2602.06442},
  year={2026}
}

@article{cui2025paddleocr,
  title={Paddleocr 3.0 technical report},
  author={Cui, Cheng and Sun, Ting and Lin, Manhui and Gao, Tingquan and Zhang, Yubo and Liu, Jiaxuan and Wang, Xueqing and Zhang, Zelun and Zhou, Changda and Liu, Hongen and others},
  journal={arXiv preprint arXiv:2507.05595},
  year={2025}
}

@article{seedream2025seedream,
  title={Seedream 4.0: Toward next-generation multimodal image generation},
  author={Seedream, Team and Chen, Yunpeng and Gao, Yu and Gong, Lixue and Guo, Meng and Guo, Qiushan and Guo, Zhiyao and Hou, Xiaoxia and Huang, Weilin and Huang, Yixuan and others},
  journal={arXiv preprint arXiv:2509.20427},
  year={2025}
}

@article{seedance2026seedance,
  title={Seedance 2.0: Advancing video generation for world complexity},
  author={Seedance, Team and Chen, De and Chen, Liyang and Chen, Xin and Chen, Ying and Chen, Zhuo and Chen, Zhuowei and Cheng, Feng and Cheng, Tianheng and Cheng, Yufeng and others},
  journal={arXiv preprint arXiv:2604.14148},
  year={2026}
}

@article{mao2026toward,
  title={Toward Accurate Image Generation via Dynamic Generative Image Transformer},
  author={Mao, Zhendong and Huang, Mengqi and Lin, Yijing and Wang, Quan and Zhang, Lei and Zhang, Yongdong},
  journal={IEEE Transactions on Pattern Analysis and Machine Intelligence},
  year={2026},
  publisher={IEEE}
}

@inproceedings{huang2023towards,
  title={Towards accurate image coding: Improved autoregressive image generation with dynamic vector quantization},
  author={Huang, Mengqi and Mao, Zhendong and Chen, Zhuowei and Zhang, Yongdong},
  booktitle={Proceedings of the IEEE/CVF Conference on Computer Vision and Pattern Recognition},
  pages={22596--22605},
  year={2023}
}

@inproceedings{huang2024realcustom,
  title={Realcustom: Narrowing real text word for real-time open-domain text-to-image customization},
  author={Huang, Mengqi and Mao, Zhendong and Liu, Mingcong and He, Qian and Zhang, Yongdong},
  booktitle={Proceedings of the IEEE/CVF Conference on Computer Vision and Pattern Recognition},
  pages={7476--7485},
  year={2024}
}

@article{mao2025realcustom++,
  title={RealCustom++: Representing Images as Real Textual Word for Real-Time Customization},
  author={Mao, Zhendong and Huang, Mengqi and Ding, Fei and Liu, Mingcong and He, Qian and Zhang, Yongdong},
  journal={IEEE Transactions on Pattern Analysis and Machine Intelligence},
  year={2025},
  publisher={IEEE}
}

@article{wu2025uso,
  title={Uso: Unified style and subject-driven generation via disentangled and reward learning},
  author={Wu, Shaojin and Huang, Mengqi and Cheng, Yufeng and Wu, Wenxu and Tian, Jiahe and Luo, Yiming and Ding, Fei and He, Qian},
  journal={arXiv preprint arXiv:2508.18966},
  year={2025}
}

@article{cheng2025umo,
  title={UMO: Scaling Multi-Identity Consistency for Image Customization via Matching Reward},
  author={Cheng, Yufeng and Wu, Wenxu and Wu, Shaojin and Huang, Mengqi and Ding, Fei and He, Qian},
  journal={arXiv preprint arXiv:2509.06818},
  year={2025}
}

@inproceedings{lin2025realgeneral,
  title={Realgeneral: Unifying visual generation via temporal in-context learning with video models},
  author={Lin, Yijing and Huang, Mengqi and Zhuang, Shuhan and Mao, Zhendong},
  booktitle={Proceedings of the IEEE/CVF International Conference on Computer Vision},
  pages={14994--15004},
  year={2025}
}

@inproceedings{huang2022dse,
  title={Dse-gan: Dynamic semantic evolution generative adversarial network for text-to-image generation},
  author={Huang, Mengqi and Mao, Zhendong and Wang, Penghui and Wang, Quan and Zhang, Yongdong},
  booktitle={Proceedings of the 30th ACM International Conference on Multimedia},
  pages={4345--4354},
  year={2022}
}

@article{liu2026flow,
  title={Flow-grpo: Training flow matching models via online rl},
  author={Liu, Jie and Liu, Gongye and Liang, Jiajun and Li, Yangguang and Liu, Jiaheng and Wang, Xintao and Wan, Pengfei and Zhang, Di and Ouyang, Wanli},
  journal={Advances in neural information processing systems},
  volume={38},
  pages={40783--40818},
  year={2026}
}

@article{xue2025dancegrpo,
  title={Dancegrpo: Unleashing grpo on visual generation},
  author={Xue, Zeyue and Wu, Jie and Gao, Yu and Kong, Fangyuan and Zhu, Lingting and Chen, Mengzhao and Liu, Zhiheng and Liu, Wei and Guo, Qiushan and Huang, Weilin and others},
  journal={arXiv preprint arXiv:2505.07818},
  year={2025}
}

@article{zheng2025diffusionnft,
  title={Diffusionnft: Online diffusion reinforcement with forward process},
  author={Zheng, Kaiwen and Chen, Huayu and Ye, Haotian and Wang, Haoxiang and Zhang, Qinsheng and Jiang, Kai and Su, Hang and Ermon, Stefano and Zhu, Jun and Liu, Ming-Yu},
  journal={arXiv preprint arXiv:2509.16117},
  year={2025}
}

@article{huang2026self,
  title={Self forcing: Bridging the train-test gap in autoregressive video diffusion},
  author={Huang, Xun and Li, Zhengqi and He, Guande and Zhou, Mingyuan and Shechtman, Eli},
  journal={Advances in Neural Information Processing Systems},
  volume={38},
  pages={167283--167308},
  year={2026}
}

@article{wu2026stream,
  title={Stream-R1: Reliability-Perplexity Aware Reward Distillation for Streaming Video Generation},
  author={Wu, Bin and Huang, Mengqi and Wu, Shaojin and Jia, Weinan and Wang, Yuxin and Mao, Zhendong and Zhang, Yongdong},
  journal={arXiv preprint arXiv:2605.03849},
  year={2026}
}

@article{tu2026stream,
  title={Stream-T1: Test-Time Scaling for Streaming Video Generation},
  author={Tu, Yijing and Wu, Shaojin and Huang, Mengqi and Wang, Wenchuan and Wang, Yuxin and Liu, Chunxiao and Mao, Zhendong},
  journal={arXiv preprint arXiv:2605.04461},
  year={2026}
}

@article{feng2026dreamlite,
  title={DreamLite: A Lightweight On-Device Unified Model for Image Generation and Editing},
  author={Feng, Kailai and Wei, Yuxiang and Chen, Bo and Pan, Yang and Ye, Hu and Liu, Songwei and Yan, Chenqian and Gao, Yuan},
  journal={arXiv preprint arXiv:2603.28713},
  year={2026}
}

\end{document}